\documentclass[lettersize,journal]{IEEEtran}
\usepackage{amsmath,amsfonts}
\usepackage{algorithmic}
\usepackage{algorithm}
\usepackage{array}
\usepackage[caption=false,font=normalsize,labelfont=sf,textfont=sf]{subfig}
\usepackage{textcomp}
\usepackage{stfloats}
\usepackage{url}
\usepackage{verbatim}
\usepackage{graphicx}
\usepackage{cite}
\usepackage{multirow}
\usepackage{multicol}
\usepackage{makecell}
\usepackage{booktabs}%
\usepackage{pifont} 
\usepackage{colortbl}
\usepackage[table]{xcolor} 
\usepackage{hyperref}
\usepackage{nicefrac}
\usepackage{tabularx}
\usepackage{ragged2e}

\newcommand{\mathsmallish}[1]{%
  \mathchoice
    {\scalebox{0.75}{$\displaystyle #1$}}
    {\scalebox{0.75}{$\textstyle #1$}}
    {\scalebox{0.75}{$\scriptstyle #1$}}
    {\scalebox{0.75}{$\scriptscriptstyle #1$}}
}
\newcommand{\mathtiny}[1]{%
  \mathchoice
    {\scalebox{0.65}{$\displaystyle #1$}}
    {\scalebox{0.65}{$\textstyle #1$}}
    {\scalebox{0.65}{$\scriptstyle #1$}}
    {\scalebox{0.65}{$\scriptscriptstyle #1$}}
}

\DeclareMathOperator*{\argmin}{argmin}
\newcommand{\xmark}{{\color[HTML]{000000}\ding{55}}}   
\newcommand{\cmark}{{\color[HTML]{000000}\ding{51}}} 

\begin{document}

\title{Towards High-Resolution Industrial Image Anomaly Detection}

\author{Ximiao Zhang, Min Xu, and Xiuzhuang Zhou, \IEEEmembership{Senior Member, IEEE}

\thanks{Ximiao Zhang and Xiuzhuang Zhou are with the School of Intelligent Engineering and Automation, Beijing University of Posts and Telecommunications, Beijing 100876, China. (E-mail: 2024010482@bupt.cn; xiuzhuang.zhou@bupt.edu.cn). 

Min Xu is with the College of Information and Engineering, Capital Normal University, Beijing 100048, China. (E-mail: xumin@cnu.edu.cn).

Corresponding author: Xiuzhuang Zhou.
}}
\markboth{IEEE TRANSACTIONS ON IMAGE PROCESSING}%
{Shell \MakeLowercase{\textit{et al.}}: Towards High-Resolution Industrial Image Anomaly Detection}


\maketitle

\begin{abstract}
Current anomaly detection methods primarily focus on low-resolution scenarios. For high-resolution images, conventional downsampling often misses subtle anomalies due to the loss of fine-grained discriminative information. Although recent studies have explored lightweight networks or image tiling with ensemble strategies, they still fall short of industrial requirements for both accuracy and efficiency. To address the above issues, we propose HiAD, a general framework for high-resolution anomaly detection that identifies anomalies of varying sizes under constrained computational resources. Specifically, HiAD employs a dual-branch architecture that integrates anomaly cues across different scales to comprehensively capture both subtle and large-scale anomalies. Furthermore, it incorporates a multi-resolution feature fusion module that injects multi-scale semantic context into patch-level representations, thereby suppressing ultrafine texture variations while retaining anomaly-relevant structural details. To improve adaptability and efficiency, HiAD constructs a detector pool with an adaptive strategy selection mechanism that automatically identifies the optimal detector assignment strategy for varying input data, enabling robust anomaly detection while effectively controlling computational costs. HiAD requires neither additional supervision nor auxiliary learnable parameters and can be seamlessly integrated with a wide range of anomaly detection methods. Extensive experiments are conducted on multiple high-resolution anomaly detection benchmarks, including the specifically constructed MVTec-HD, VisA-HD, and RealIAD-HD benchmarks, as well as the existing MVTec 2 benchmark. The results consistently demonstrate the effectiveness and superior performance of HiAD across diverse experimental settings. The code is available at \url{https://github.com/cnulab/HiAD}.
\end{abstract}


\section{Introduction}
\IEEEPARstart{U}{nsupervised} image anomaly detection relies solely on normal samples to detect and locate anomalous regions. It has many practical applications and has attracted extensive research in recent years \cite{tao2022deep, liu2024deep}. However, the detection resolution of existing methods is far lower than the imaging resolution of industrial cameras, which hinders the detection of subtle anomalous regions. Subtle anomalies are common in modern industrial production, especially in precision manufacturing, which may manifest as minor scratches or tiny structural defects, significantly impacting product quality. Subtle anomalies cannot be detected at low resolutions, as the information loss introduced by downsampling inevitably results in missed detections by modern neural networks. Although some recent works have employed lightweight networks \cite{cao2023high, cao2025varad} or simple tiled ensemble strategies \cite{rolih2024divide, meininger2024unsupervised} to improve the resolution of anomaly detection, they still suffer from poor scalability, limited accuracy, and high computational cost, rendering them impractical for real-world deployment. Therefore, developing a scalable high-resolution anomaly detection framework that offers both high precision and computational efficiency has become an urgent requirement for the industrial sector and the anomaly detection community.

We conduct a comprehensive study on high-resolution (1K$-$4K) industrial image anomaly detection and summarize its key challenges into three main aspects. First, high-resolution anomaly detection methods must meet industrial demands for precision and efficiency. However, due to limited GPU memory, high-resolution images cannot be processed in a single forward pass, which may lead to a decline in detection performance and efficiency. Second, existing pre-trained backbones designed for low-resolution images struggle to detect large-scale anomalies at high resolutions. This is due to the limited receptive field and scalability of these pre-trained networks on high-resolution images, which make it difficult to capture anomalous semantics spanning hundreds or even thousands of pixels. Third, fine-grained texture information in high-resolution images complicates the distribution of samples, which increases the intra-class distance of normal samples. This may cause the anomaly detection model to incorrectly classify texture variations as anomalies. 

\begin{figure}[t]
  \centering
   \includegraphics[width=\linewidth]{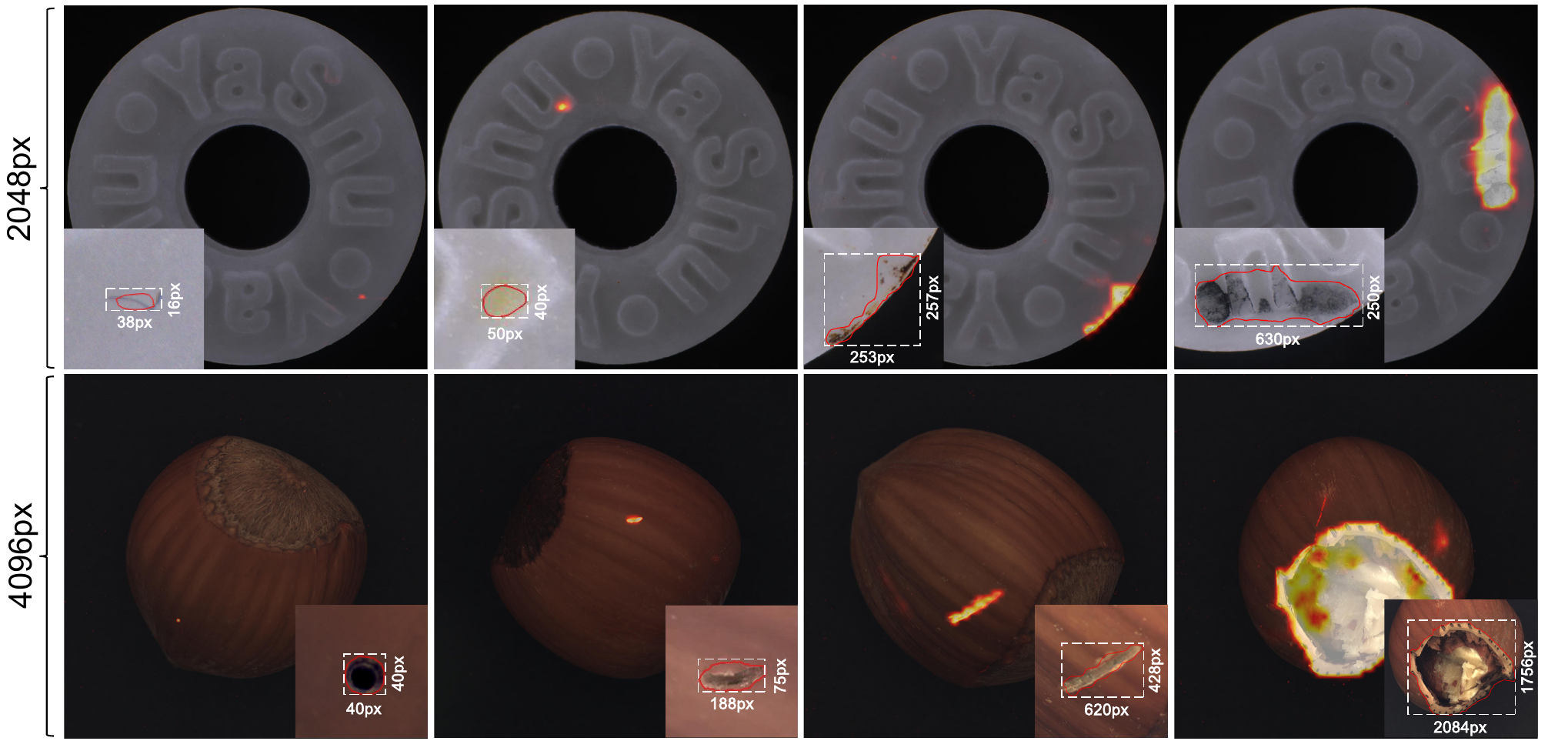}
   \caption{HiAD can effectively detect anomalous regions of various scales in high-resolution images. Brightness represents anomaly scores, and red boundaries delineate detected anomalous regions.}
   \label{fig:fig1}
   \vspace{0.1cm}
\end{figure}

To address the aforementioned challenges, we propose \textbf{HiAD}, an unsupervised \textbf{Hi}gh-resolution \textbf{A}nomaly \textbf{D}etection framework designed to extend existing low-resolution methods to arbitrary high-resolution scenarios. As illustrated in Figure \ref{fig:fig1}, HiAD accurately localizes anomalous regions ranging from tens to thousands of pixels in 4K imagery. Specifically, HiAD leverages a dual-branch architecture that integrates a high-resolution branch for preserving subtle local cues with a low-resolution branch for capturing global context and large-scale anomalies. In the high-resolution branch, images are divided into patches for localized detection. A detector pool, combined with multiple detector assignment strategies, ensures broad adaptability of HiAD across varied data distributions. Moreover, we introduce an adaptive strategy selection mechanism that autonomously identifies the optimal assignment scheme by evaluating feature assignment fidelity and spatial coherence across different strategies. This enables HiAD to tailor its detection schemes to specific industrial products under a given computational budget, achieving a favorable balance between detection accuracy and efficiency. To further address the challenges posed by fine-grained texture variations in high-resolution images, HiAD incorporates a multi-resolution feature fusion module that leverages multi-scale semantic cues to calibrate local patch features and reduce false-positive detections. Unlike existing high-resolution frameworks \cite{chen2019collaborative,guo2022isdnet,qin2025boosting} that rely on task-specific model optimization, HiAD adopts a detector-agnostic architecture without introducing supervision-dependent auxiliary parameters. This design enables existing unsupervised anomaly detectors to be seamlessly integrated with HiAD without altering their original learning objectives. Furthermore, its parallelized computational pipeline enables patch-level training and inference across multiple GPUs, offering flexible resource allocation and strong computational scalability.

To comprehensively validate the effectiveness of the proposed HiAD framework, we conduct extensive experiments on the synthetic benchmarks MVTec-HD and VisA-HD, as well as the real-world benchmarks RealIAD-HD and MVTec 2 \cite{heckler2026mvtec}. The first three are specifically constructed for the high-resolution anomaly detection task, building upon the widely studied MVTec-AD \cite{bergmann2019mvtec}, VisA \cite{zou2022spot}, and Real-IAD \cite{wang2024real} datasets. These benchmarks feature high-resolution images with anomalous regions of varying scales, posing a significant challenge for anomaly detection. HiAD achieves satisfactory performance across all benchmarks and multiple experimental settings, substantially outperforming alternative methods and demonstrating its superiority and practicality for high-resolution anomaly detection. The main contributions of this work are summarized as follows:

\begin{itemize}
\item We identify the key challenges in high-resolution industrial image anomaly detection and propose HiAD, a general and scalable framework. To the best of our knowledge, HiAD is the first approach to scale practical detection resolutions up to the 1K$-$4K range, while accurately detecting anomalous regions across varying scales with high computational efficiency.  
\item HiAD supports adaptive strategy selection for detector assignment and can be seamlessly integrated with a wide range of anomaly detection methods, demonstrating broad applicability across diverse high-resolution scenarios.
\item We conduct comprehensive experiments across multiple high-resolution anomaly detection benchmarks. The results consistently demonstrate the effectiveness and reliability of the proposed HiAD framework, paving the way for future research.
\end{itemize}

\section{Related Work}

\subsection{Unsupervised Image Anomaly Detection}

Image anomaly detection has gained widespread attention in recent years, leading to a series of studies \cite{roth2022towards, defard2021padim, deng2022anomaly, zavrtanik2021draem, liang2023omni}. Due to the scarcity of anomaly samples in real-world scenarios, most anomaly detection methods follow an unsupervised paradigm, training solely on normal data and identifying anomaly samples that deviate from normal patterns during inference. These methods can be roughly categorized into three main types: reconstruction-based methods, deep feature embedding-based methods, and anomaly synthesis-based methods. Reconstruction-based methods \cite{zhang2023exploring, zhang2025diffusionad, guo2025dinomaly} train a reconstruction network using only normal images. When an anomaly image is input, these networks fail to effectively reconstruct the anomalous regions, enabling anomaly detection and localization through reconstruction residuals. Deep feature embedding-based methods \cite{yi2020patch, defard2021padim, roth2022towards, gudovskiy2022cflow} learn the feature distribution of normal images and identify anomalies by measuring the distance between query image features and the learned feature distribution. Constructing memory banks \cite{roth2022towards, bae2023pni} and learning normalizing flow models \cite{yu2021fastflow, gudovskiy2022cflow, liu2024cross} are common approaches for feature modeling. Anomaly synthesis-based methods \cite{li2021cutpaste, schluter2022natural, zhang2023prototypical, zhang2024realnet, wang2026stones} train anomaly detection models using synthetic anomaly samples. These approaches emphasize that synthetic anomalies can effectively simulate real-world anomalies. Despite their advancements, these methods are designed for low-resolution images, assuming that the detection can be completed in a single forward pass on modern GPUs. For high-resolution images containing subtle anomalies, these methods often struggle with detection.

\subsection{High-Resolution Anomaly Detection}

High-resolution images are widely used in fields such as industrial production \cite{wang2024real, horwath2020understanding, fu2024towards}, autonomous driving \cite{cordts2016cityscapes, zheng2021deep, ding2025hilm}, remote sensing \cite{jiang2022survey, li2023anomaly,zheng2024single}, and histopathology \cite{van2021deep, cheng2021robust,li2023task}. For deep learning models, using high-resolution images as input poses numerous challenges, as they involve a large number of parameters, leading to significant increases in computational cost and inference latency. There are two baseline approaches to applying deep learning models to high-resolution anomaly detection tasks. The first approach is to downsample high-resolution images to the desired resolution for detection. This method is called Uniform Downsampling because the quality is uniformly reduced across the entire image, which will inevitably lead anomaly detection models to miss subtle anomalies, making it unacceptable for tasks with high precision requirements. The second approach divides high-resolution images into smaller patches with maximum resolution, detects each patch independently, and finally aggregates the results. This approach preserves detailed image information and enables high-precision detection, but it also faces challenges in terms of computational cost. 

Some recent works have attempted to improve the resolution of anomaly detection using lightweight networks. For instance, STCIKD \cite{cao2023high} converts high-resolution images into video sequences for anomaly detection, while VarAD \cite{cao2025varad} reconstructs image patches using a lightweight Mamba model \cite{gu2024mamba}, enabling the detection of industrial images at a resolution of $1024 \times 1024$. Despite these efforts, the above methods still require high-resolution images to be processed in a single forward pass, which imposes inherent limitations on scalability. A more closely related work is that of Rolih et al. \cite{rolih2024divide}, who propose a simple tiled ensemble method for high-resolution anomaly detection. They first divide high-resolution images into smaller patches and then train a separate anomaly detection model for each position, ultimately aggregating the detection results. However, the number of models required grows quadratically with resolution, making it difficult to scale to higher resolutions. Beyond the challenges of computational scalability, the inherently open-set and unsupervised nature of the task makes it difficult to reliably optimize additional trainable parameters, thereby limiting model expressiveness. Moreover, the lack of publicly available benchmarks for high-resolution anomaly detection hinders the systematic validation and real-world deployment of existing methods.

\section{Approach}

\begin{figure}[t]
  \centering
   \includegraphics[width=\linewidth]{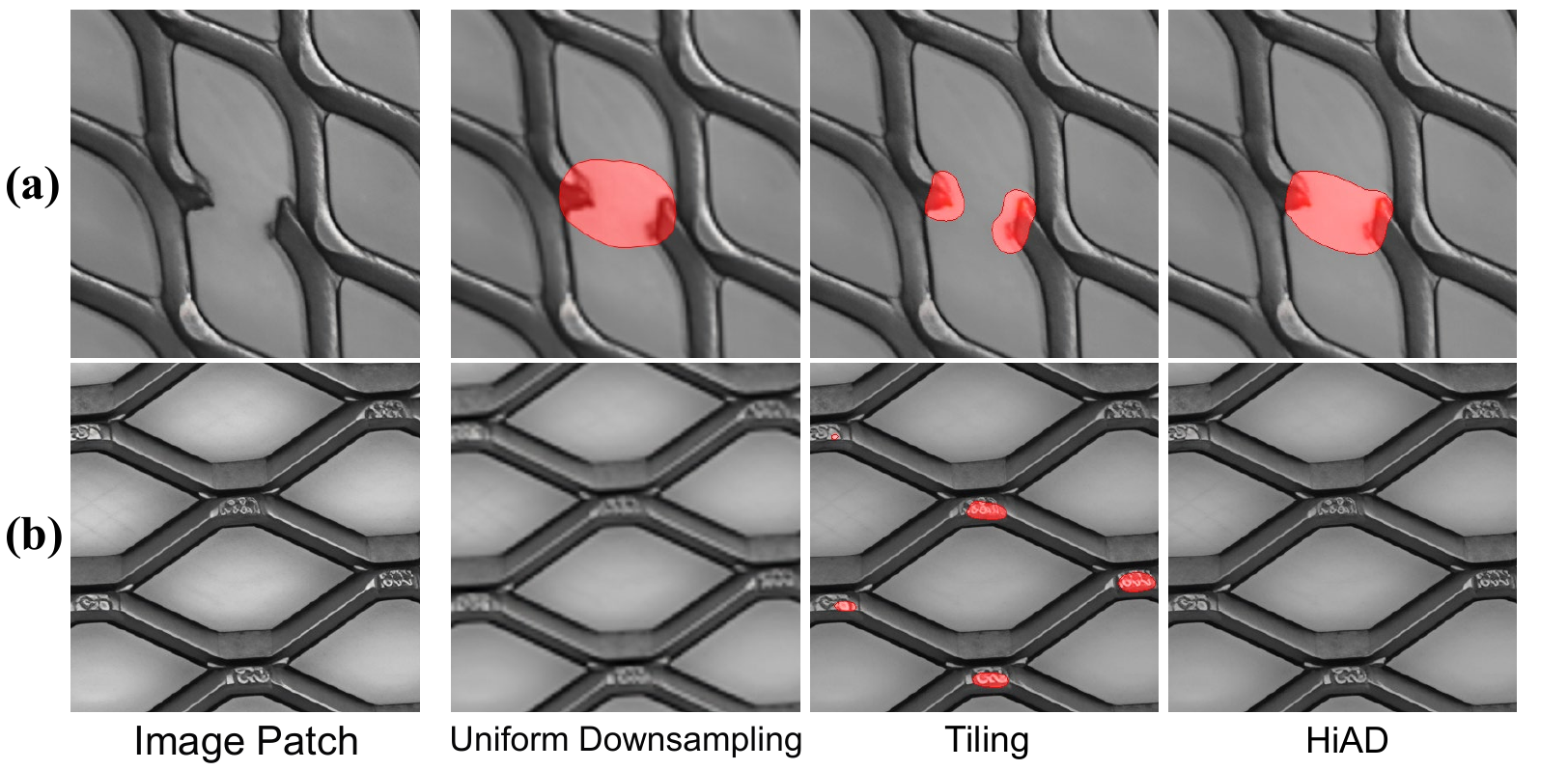}
   \caption{(a) Tiling results in discontinuous detection for large-scale anomalies. (b) Tiling misidentifies image texture variations as anomalies. In this experiment, all patches are cropped from grid images with a resolution of $2048 \times 2048$. Tiling and HiAD perform anomaly detection at the original resolution, while Uniform Downsampling downsamples the images to $512 \times 512$.}
   \label{fig:fig2}
\end{figure}

Our goal is to detect anomalous regions at varying scales in high-resolution images while maintaining computational efficiency. \textbf{Tiling} is a common solution for processing high-resolution images, in which the image is first divided into multiple patches for individual detection followed by aggregation of results. However, this straightforward approach still suffers from three key limitations that hinder its applicability in real-world scenarios:
\begin{itemize}
\item Accurate identification of large-scale anomalous regions in high-resolution images requires the model to capture global anomaly semantics across large spatial extents. However, the limited receptive field and scalability of existing pre-trained networks, coupled with the global information loss caused by Tiling, hinder the effectiveness of simple tiling-based solutions in detecting large-scale anomalies. This may result in discontinuous or even missed detections. For instance, as shown in Figure \ref{fig:fig2}(a), a broken grid anomaly is fully identified at low resolution, but when scaled to a higher resolution, the same anomalous region spans more pixels, causing Tiling to produce discontinuous detection results. In some industrial scenarios, it is crucial to fully segment the anomalous region, such as when assessing the severity based on the area of the anomalous region.
\item Due to the limited generalization ability of existing anomaly detection methods, fine-grained texture variations in high-resolution images may be incorrectly identified as anomalies. As shown in Figure \ref{fig:fig2}(b), the unique metal texture in the high-resolution grid image causes Tiling to produce over-detection. This unique metal texture may never have appeared in the training data, but it represents intra-class variation of normal samples. We refer to the fine-grained texture that exceeds the model capacity of existing anomaly detection methods as \textbf{Ultrafine Texture}. Ultrafine textures are pervasive in high-resolution images and lead to widespread false-positive detections.
\item Tiling divides high-resolution images into multiple patches. The natural question is how we can assign these patches to anomaly detection models to balance detection performance and computational cost. A straightforward method is to train a single model for all patches. However, this effectively transforms the task into a multi-class anomaly detection problem \cite{you2022unified, zhang2023exploring, guo2025dinomaly} where inter-patch differences increase the complexity of modeling the normal feature distribution, often resulting in degraded anomaly detection performance. To mitigate this, Rolih et al. \cite{rolih2024divide} train a model for each spatial position, resulting in uncontrollable computational and storage costs.
\end{itemize}

In this section, we introduce HiAD, our high-resolution anomaly detection framework designed to address the challenges outlined above. We first describe the overall architecture of HiAD, and then detail its key components, including the multi-resolution feature fusion module, the detector pool along with various detector assignment strategies, as well as the proposed adaptive strategy selection method.

\subsection{Overview of HiAD}

\begin{figure*}[t]
  \centering
   \includegraphics[width=0.95\linewidth]{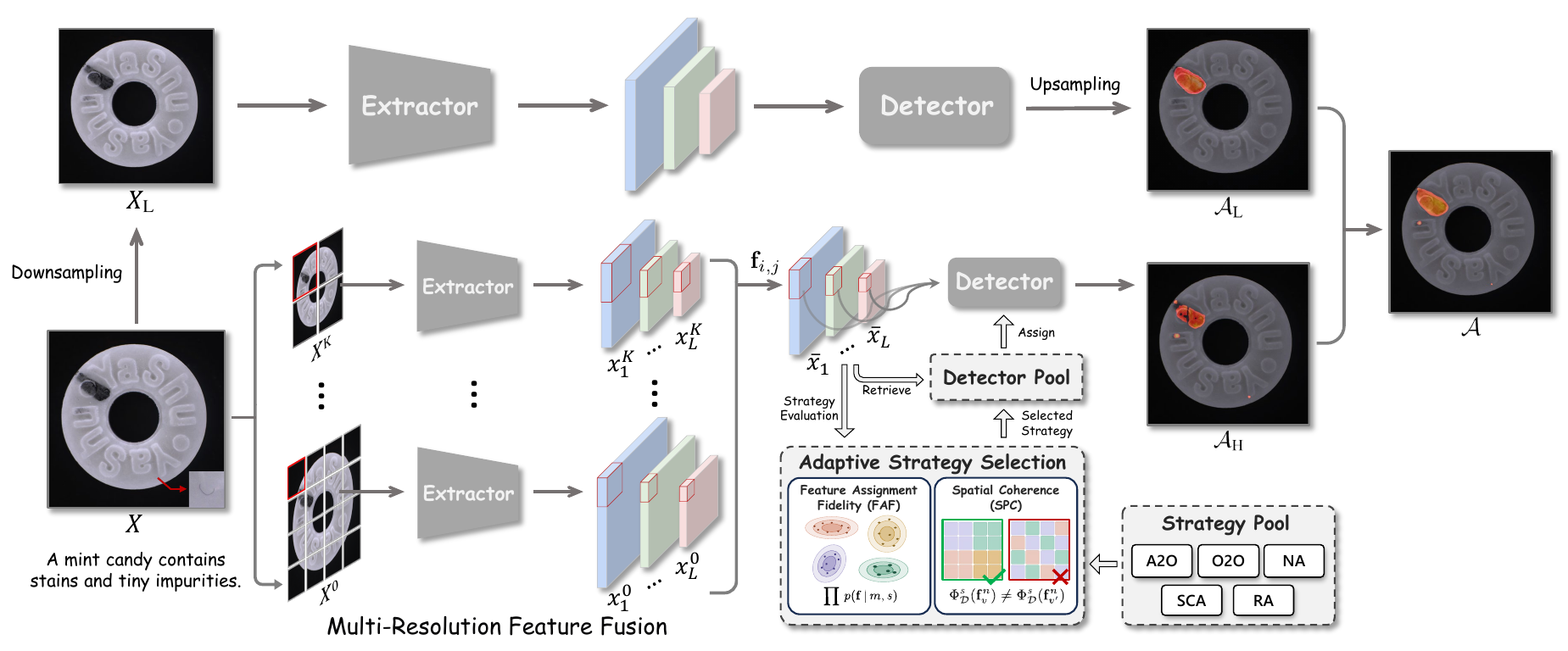}
   \caption{Overview of the proposed HiAD framework, which integrates low-resolution (top) and high-resolution (bottom) branches to detect anomalous regions at different scales. The high-resolution branch employs multi-resolution feature fusion to mitigate false-positive detections induced by ultrafine textures. In addition, a detector pool coupled with adaptive strategy selection enables HiAD to choose the most appropriate detector assignment strategy based on image characteristics, thereby handling structural diversity in high-resolution images under practical computational constraints.}
   \label{fig:fig3}
\end{figure*}

The overall pipeline of HiAD is illustrated in Figure \ref{fig:fig3}, which consists of a high-resolution branch and a low-resolution branch, performing anomaly detection at different scales. For the high-resolution branch, given an input image $X \in \mathbb{R}^{H \times W \times 3}$, we divide $X$ into a set of image patches $\mathcal{P}$ using a patch size $H_{\text{p}} \times W_\text{p}$ and a stride $H_\text{s} \times W_\text{s}$, where $P_{i,j} \in \mathbb{R}^{H_\text{p} \times W_\text{p} \times 3} $ denotes the patch at location $(i, j)$. We define the inverse operation of patch division as $X = G(\mathcal{P})$, where the aggregation function $G(\cdot)$ reassembles the image patches in $\mathcal{P}$ into the image $X$. For overlapping regions in multiple patches, the aggregation function takes the average of the overlapping pixels. HiAD extracts multi-resolution fusion features from each patch, represented as $\mathbf{f}_{i,j} = \boldsymbol{f}(P_{i,j})$, where $\boldsymbol{f}(\cdot)$ denotes the multi-resolution feature extraction function. For the image patch set $\mathcal{P}$, we define its feature set as $\boldsymbol{F}=\{\mathbf{f}_{i,j}\mid \mathbf{f}_{i,j} = \boldsymbol{f}(P_{i,j}), P_{i,j} \in \mathcal{P} \}$. In the high-resolution branch, we train multiple detectors, which form a detector pool $\mathcal{D}=\{ D_1, D_2, \dots, D_M\}$, where each detector is responsible for detecting a subset of patches, such as detecting patches from a specific position. The detectors are model instances trained on different image patches using any feature-based anomaly detection method, such as PatchCore \cite{roth2022towards}. To handle high-resolution images with diverse spatial structures, we devise multiple detector assignment strategies, forming a strategy pool $\mathcal{S}$. The adaptive strategy selection function selects the optimal assignment strategy $s^*= \Psi_{\mathcal{S}}(\mathcal{F})$ according to the distribution of the feature sets $\mathcal{F}=\{\boldsymbol{F}_1,\boldsymbol{F}_2,\dots,\boldsymbol{F}_N\}$. The detector assignment function $\Phi(\cdot)$ assigns each patch feature to the corresponding detector according to the selected assignment strategy, denoted as $D=\Phi_{\mathcal{D}}^{s^*}(\mathbf{f}_{i,j})$, where $D \in \mathcal{D}$. The assigned detector outputs the anomaly score $A_{i,j}=D(\mathbf{f}_{i,j}) \in \mathbb{R}^{H_{\text{p}} \times W_{\text{p}}} $ based on the patch feature. Then, the anomaly map from the high-resolution branch is:
\begin{equation}
\label{eq:equ1}
\mathcal{A}_{\text{H}} = G(\{ D(\mathbf{f}_{i,j}) \mid D=\Phi_{\mathcal{D}}^{s^*}(\mathbf{f}_{i,j}), \mathbf{f}_{i,j} \in \boldsymbol{F}\}).
\end{equation}
For the low-resolution branch, we first downsample the high-resolution image $X$ to a desired low resolution $X_{\text{L}} \in \mathbb{R}^{H_{\text{L}} \times W_\text{L} \times 3}$. Then, we use a pre-trained network $\phi(\cdot)$ and a low-resolution detector $D_{\text{L}}(\cdot)$ for feature extraction and anomaly detection, to obtain the anomaly map $\mathcal{A}_{\text{L}}=Up(D_{\text{L}}(\phi(X_{\text{L}})))$, where the $Up(\cdot)$ function upsamples the low-resolution anomaly map to match the original resolution, with $\mathcal{A}_\text{H},\mathcal{A}_\text{L} \in \mathbb{R}^{H \times W}$. The subsequent challenge lies in aggregating detection results from different branches to fully leverage the complementary strengths of both branches. We find that the parameter-free pixel-wise maximum is a straightforward and effective scheme; it effectively integrates the high-resolution branch's sensitivity to subtle anomalies with the low-resolution branch's coverage of global, large-scale anomalies, thereby ensuring comprehensive recall of anomalies across various scales. Formally, for each pixel position $(a, b)$, the final anomaly map $\mathcal{A}$ is defined as:
\begin{equation}
\label{eq:equ2}
\mathcal{A}(a,b) = max(\mathcal{A}_\text{H}(a,b),\mathcal{A}_\text{L}(a,b)).
\end{equation}

\subsection{Multi-Resolution Feature Fusion}

Multi-Resolution Feature Fusion (MRFF) is designed to enhance detector robustness against fine-grained texture variations in high-resolution images, thereby reducing false positives. These ultrafine textures are unique to high-resolution images and tend to diminish progressively with successive downsampling. We first define a set of downsampling rates $\mathcal{T}=\{0,1,\dots,K \}$. For a given high-resolution image $X$ and a downsampling rate $k \in \mathcal{T}$, the downsampled image is defined as $X^k = Down(X,2^k) \in \mathbb{R}^{H/2^k \times W/2^k \times 3}$, where the $Down$ function downsamples the image $X$ by a factor of $2^k$. Through this process, we can obtain a set of progressively downsampled images $\{X^0,X^1,\dots,X^K \}$, where $X^0 = X$. Subsequently, we extract features from these multi-resolution images in a patch-by-patch manner and spatially aggregate them. The resulting feature map of image $X^k$ at the $l$-th intermediate layer is defined as $x^k_l$. Then, the multi-resolution fusion feature $\bar{x}_{l}$ of image $X$ at the $l$-th layer is computed as a weighted combination of features from multiple resolutions:
\begin{equation}
\label{eq:equ3}
\bar{x}_{l}=\sum_{k=0}^{K}Up(x^k_l,2^k)\mathbf{w}_k,
\end{equation}
where the $Up$ function upsamples the multi-resolution features to match the original feature resolution of image $X$, and $\mathbf{w}_k \in [0,1] $ represents the fusion weight, satisfying $\sum_{k=0}^{K}\mathbf{w}_k=1$. We fuse the features from each layer and then divide them into patch features based on the feature-level patch size and stride. The multi-resolution fusion feature of the image patch at position $(i, j)$ is represented as
$\mathbf{f}_{i,j}=\{\bar{x}_{1}(i,j),\bar{x}_{2}(i,j),\dots,\bar{x}_{L}(i,j)\}$, where $L$ is the number of extracted intermediate layers.

\begin{figure}[t]
  \centering
   \includegraphics[width=\linewidth]{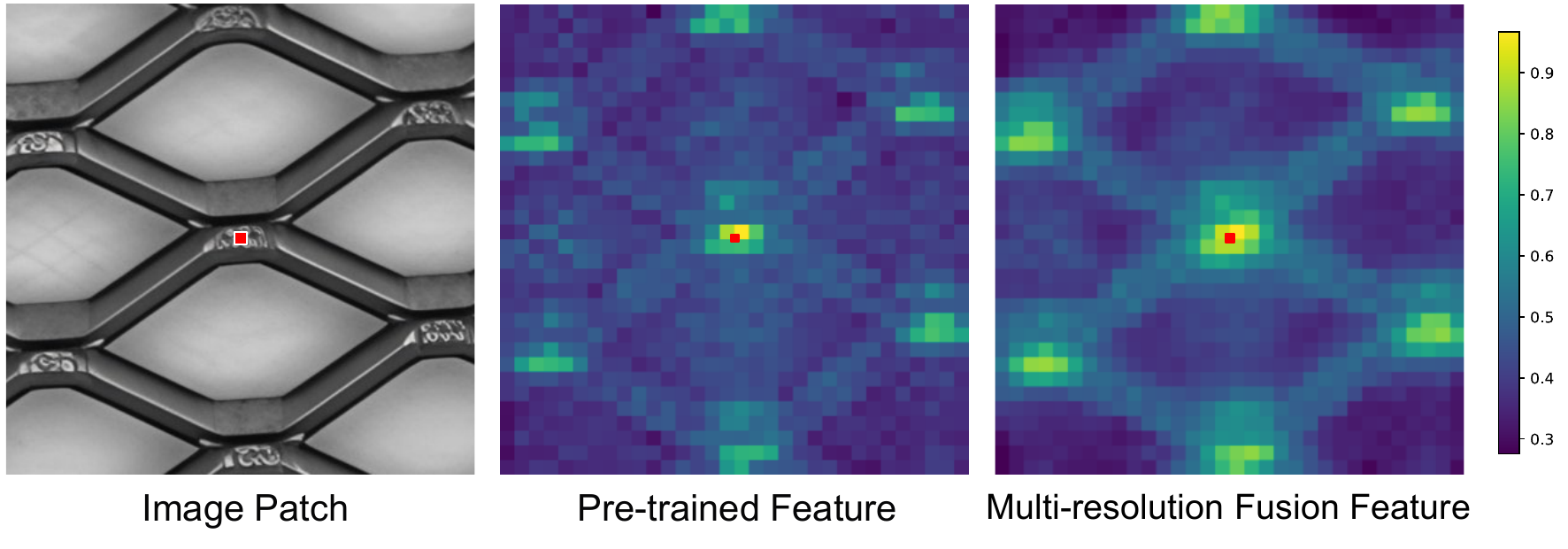}
   \caption{Comparison of cosine similarity maps between the pre-trained feature and the multi-resolution fusion feature. The similarity maps are computed between the red-marked patch feature and all other features.}
   \label{fig:fig4}
\end{figure}

The MRFF module integrates high-resolution features with their lower-resolution counterparts to attenuate ultrafine textures while preserving critical structural variations through the high-level semantic knowledge of pre-trained backbones. As illustrated in Figure \ref{fig:fig4}, this strategy effectively suppresses high-frequency noise and enhances feature consistency across similar normal regions, thereby improving the robustness of HiAD to complex texture variations while maintaining high precision in detecting diverse anomalous patterns.

\subsection{Detector Pool and Detector Assignment}

Inspired by the Mixture of Experts (MoE) architecture \cite{jacobs1991adaptive, shazeer2017outrageously}, we incorporate a detector pool into HiAD along with multiple detector assignment strategies to achieve both broad applicability and high detection efficiency. Employing multiple detectors increases the overall model capacity of HiAD, thereby improving its capability to detect a wide range of patches. Different detector assignment strategies establish distinct mappings between patches and detectors, increasing HiAD's adaptability to diverse high-resolution images. Moreover, the detector pool can be flexibly scaled to match available resources, allowing HiAD to accommodate varying computational budgets while keeping computation and storage overheads manageable. Specifically, for a high-resolution image set $\mathcal{X}=\{X_1,X_2,\dots,X_N\}$, we divide each image into $\boldsymbol{I} \times \boldsymbol{J}$ patches and extract multi-resolution fusion feature sets $\mathcal{F}=\{\boldsymbol{F}_1,\boldsymbol{F}_2,\dots,\boldsymbol{F}_N\}$. We represent the patch feature of the $n$-th image at position $(i,j)$ as $\mathbf{f}_{i,j}^n \in \boldsymbol{F}_n$. In this section, $\mathbf{f}_{i,j}^n$ represents an integrated feature obtained by concatenating features from multiple intermediate layers. HiAD trains $M$ detectors for the image set $\mathcal{X}$, forming a detector pool $\mathcal{D}=\{ D_1,D_2,\dots,D_M\}$. Each patch is assigned to a specific detector based on its feature representation, while ensuring consistent assignment during the training and inference phases. We propose five distinct detector assignment strategies for HiAD, as detailed below:

\textit{1) All-to-One Assignment (A2O).} A2O assigns all image patches to a single detector, denoted as $\mathcal{D}=\{D\}$, and $D = \Phi_{\mathcal{D}}^{ \text{\tiny A2O}}(\mathbf{f}^n_{i,j})$ for all patches. Since A2O requires training only a single model, it achieves the highest computational and storage efficiency. For homogeneous high-resolution images, such as the texture categories in the MVTec-AD dataset \cite{bergmann2019mvtec}, it exhibits comparable performance to other assignment methods. However, when images have complex structures, a single detector has to jointly model the feature distributions of multiple patches, which may lead to performance degradation.

\textit{2) One-to-One Assignment (O2O).} O2O trains a dedicated detector for each spatial position, defined as $\mathcal{D}=\{D_{i,j}\mid i \in [0,\boldsymbol{I}), j \in [0,\boldsymbol{J}) \}$ and $D_{i,j} = \Phi_{\mathcal{D}}^{\text{\tiny O2O}}(\mathbf{f}^n_{i,j})$. This approach binds detectors directly to spatial positions, making it particularly suitable for anomaly detection methods that require position calibration, such as PaDiM \cite{defard2021padim}. However, when the number of positions is excessively large, this method results in significant computational and storage costs.

\textit{3) Neighborhood Assignment (NA).} This method represents a trade-off between A2O and O2O, which assigns each spatial neighborhood to a detector. Specifically, we divide all positions into $M$ non-overlapping neighborhoods $\{\mathbf{N}_1,\mathbf{N}_2,\dots,\mathbf{N}_M\} $. Image patches within the same neighborhood are assigned to the same detector, expressed as $D_m = \Phi_{\mathcal{D}}^{\text{\tiny NA}}(\mathbf{f}^n_{i,j})$, where $(i,j) \in \mathbf{N}_m$. NA decouples the number of detectors from the number of spatial positions, effectively managing the computational cost.

\textit{4) Spatial Clustering Assignment (SCA).} SCA performs assignment based on both the position and appearance of patches. For each spatial position, the average feature is computed as: $ \bar{\mathbf{f}}_{i,j} = \frac{1}{N}\sum_{n=1}^{N}\mathbf{f}^{n}_{i,j}$, where $\bar{\mathbf{f}}_{i,j}$ represents the general appearance of the patches at position $(i,j)$. These average features are then clustered into $M$ position clusters based on feature similarity, denoted as: $\{ \bar{\mathbf{f}}_{i,j}\mid i \in [0,\boldsymbol{I}), j \in [0,\boldsymbol{J}) \} \rightarrow \{ \mathbf{C}_1,\mathbf{C}_2,\dots,\mathbf{C}_M \}$, where each cluster contains multiple spatial positions with similar appearances. The assignment process can be expressed as $D_m = \Phi_{\mathcal{D}}^{\text{\tiny SCA}}(\mathbf{f}^n_{i,j})$, where $ (i,j) \in \mathbf{C}_m$. This method adaptively assigns multiple spatial positions with similar appearances to a detector, allowing each detector to learn similar feature distributions while avoiding multi-class modeling. It is suitable for industrial products with basic positional calibration. 

\textit{5) Retrieval Assignment (RA).} RA assigns detectors solely based on the appearances of image patches. First, all patch features are clustered into $M$ feature clusters, denoted as: $\{ \mathbf{f}^n_{i,j} \mid n\in[1,N], i \in [0,\boldsymbol{I}), j \in [0,\boldsymbol{J}) \} \rightarrow \{\mathbf{R}_1, \mathbf{R}_2, \dots, \mathbf{R}_M \}$, where $ \mathbf{R}_m $ represents the centroid of the $ m $-th cluster. Each cluster is assigned to a detector, and retrieval is performed based on the distance between patch features and the cluster centroids: $D_m = \Phi_{\mathcal{D}}^{\text{\tiny RA}}(\mathbf{f}^n_{i,j})$, where $m = \argmin_{m \in \{1,2,\dots,M \}} || \mathbf{f}^n_{i,j}-\mathbf{R}_m ||^2_2$. RA ignores spatial positions, allowing patches from the same location to be assigned to different detectors based on visual features. Therefore, it is suitable for detecting industrial products without positional calibration, such as the rotated screw category in the MVTec-AD dataset \cite{bergmann2019mvtec}. 

These assignment strategies provide a diverse set of candidate partitions that accommodate the structural and semantic diversity of industrial images, covering homogeneous, spatially aligned, and unaligned product layouts. By fully exploiting the regularity and consistency of the intrinsic object structures in industrial images, they simplify the feature distributions handled by individual detectors, thereby reducing their multi-class modeling burden. Since the assignments are derived solely from spatial priors or normal-feature distributions and are independent of the underlying detector architecture, they require neither anomaly supervision nor learnable routing parameters, preserving the unsupervised and detector-agnostic nature of HiAD.

\subsection{Adaptive Strategy Selection}

Given the structural and semantic diversity of industrial images, manually selecting the optimal detector assignment strategy remains challenging. The effectiveness of a specific assignment scheme is intrinsically coupled with the underlying feature distribution and the quality of feature representation. To address this issue, we propose an Adaptive Strategy Selection (ASS) mechanism, which automatically evaluates various assignment schemes based on the inherent properties of the target data and selects the most suitable detector assignment strategy accordingly. Specifically, ASS comprises two key metrics, namely Feature Assignment Fidelity and Spatial Coherence.

\textit{Feature Assignment Fidelity (FAF).} This metric evaluates the degree of match between the assigned patch features and the corresponding detector-wise feature distributions under different assignment strategies. Intuitively, an effective assignment strategy should consistently group similar patch features into the same detector, thereby alleviating the complexity of multi-class modeling and reducing the risk of false positives. Given a well-designed assignment strategy $s$, a query feature $\mathbf{f}$ assigned to the $m$-th detector is expected to follow a multivariate Gaussian distribution:
\begin{equation}
\label{eq:equ4}
\mathbf{f}\,|\,m,s \sim \mathcal{N}(\boldsymbol{\mu}_{m}^{s}, \boldsymbol{\Sigma}_{m}^{s}),
\end{equation}
where $\boldsymbol{\mu}_{m}^{s}$ and $\boldsymbol{\Sigma}_{m}^{s}$ represent the mean vector and covariance matrix of the features assigned to the $m$-th detector, which are typically estimated from the training set. To simplify covariance estimation and facilitate efficient computation, we assume that individual feature dimensions are mutually independent, leading to the following diagonal covariance matrix: $\boldsymbol{\Sigma}_{m}^{s}=\text{diag}\big((\sigma_{\mathsmallish{m,1}}^s)^2,(\sigma_{\mathsmallish{m,2}}^s)^2,\cdots,(\sigma_{\mathsmallish{m,C}}^s)^2\big)$, where $C$ denotes the feature dimension. Then, the probability density function of query feature $\mathbf{f}$ can be expressed as:
\begin{equation}
\label{eq:equ5}
p(\mathbf{f}\,|\,m,s) = \prod_{c=1}^C \frac{1}{\sqrt{2\pi \big((\sigma_{\mathsmallish{m,c}}^{s})^2 + \epsilon \big)}} \exp \left( -\frac{(\mathbf{f}_{\mathsmallish{c}} - \boldsymbol{\mu}_{\mathsmallish{m,c}}^s)^2}{2\big( (\sigma_{\mathsmallish{m,c}}^s)^2 + \epsilon \big)} \right),
\end{equation}
where $\epsilon = 10^{-4}$ is a small constant to ensure numerical stability. To assess the quality of the assignment strategy $s$, we calculate the likelihood of the query feature set $\mathcal{F}_m$ assigned to the $m$-th detector:
\begin{equation}
\label{eq:equ6}
\mathcal{L}(\boldsymbol{\mu}_m^s, \boldsymbol{\Sigma}_m^s \, | \, \mathcal{F}_m) = \prod_{\mathbf{f} \in \mathcal{F}_m} p(\mathbf{f}\,|\,m,s).
\end{equation}
Subsequently, we take the negative log-likelihood of Equation (\ref{eq:equ6}) and evaluate the feature distribution of each detector to derive the FAF score of strategy $s$:
\begin{equation}
\label{eq:equ7}
\mathcal{L}_{\text{FAF}}(s)=-\frac{1}{C \, |\mathcal{F}|}\sum_{m=1}^{M}\ln \mathcal{L}(\boldsymbol{\mu}_m^s, \boldsymbol{\Sigma}_m^s \,|\, \mathcal{F}_m).
\end{equation}
To ensure that the metric is invariant to feature dimensionality, the FAF score is normalized by the dimensionality $C$. By omitting constant terms and irrelevant coefficients, the final FAF score is simplified as follows:
\begin{equation}
\label{eq:equ8}
\begin{aligned}
\mathcal{L}_{\text{FAF}}(s) \approx \frac{1}{C} \sum_{m=1}^{M} \omega_m \Big[ & \underbrace{\sum_{c=1}^{C}\ln \big((\sigma_{\mathsmallish{m,c}}^s)^2 + \epsilon \big)}_{\text{ \tiny Distribution Compactness}} \\
 + & \underbrace{\frac{1}{|\mathcal{F}_m|}  \sum_{\mathbf{f} \in \mathcal{F}_m} \sum_{c=1}^C \frac{(\mathbf{f}_{\mathsmallish{c}} - \boldsymbol{\mu}_{\mathsmallish{m,c}}^s)^2}{(\sigma_{\mathsmallish{m,c}}^s)^2 + \epsilon}}_{\text{ \tiny Mahalanobis-like Distance}} \Big],
\end{aligned}
\end{equation}
where $\omega_m= |\mathcal{F}_m| / |\mathcal{F}|$ represents the weight of the feature set assigned to the $m$-th detector relative to the entire query set. The FAF score comprises two components. The first term characterizes the compactness of local feature distributions in terms of distribution entropy, while the second term measures the semantic alignment between query features and their assigned detectors using a Mahalanobis-like distance. Therefore, a lower FAF score indicates a more effective assignment strategy with higher fidelity.

\textit{Spatial Coherence (SPC).} Another vital metric is the spatial coherence of the assignment. In particular, spatially neighboring image patches should be assigned to the same detector, as this preserves the structural coherence of the anomaly map and thereby enhances detection performance. To quantify this, we define the set of spatial indices for the $\boldsymbol{I} \times \boldsymbol{J}$ image patches as: $\mathcal{V} = \{ (i, j) \mid i \in [0,\boldsymbol{I}), j \in [0,\boldsymbol{J}) \}$. Then, we construct the set $\mathcal{E}$ consisting of all spatially adjacent patch pairs:
\begin{equation}
\label{eq:equ9}
\mathcal{E} = \Big\{ (v, v') \in \mathcal{V} \times \mathcal{V} \,\big|\, \|v - v'\|_{1} = 1 \Big\}.
\end{equation}
For a given strategy $s$, the SPC score is defined as the ratio of heterogeneous edges:
\begin{equation}
\label{eq:equ10}
\mathcal{L}_{\text{SPC}}(s) = \frac{1}{N |\mathcal{E}|} \sum_{n=1}^N \sum_{(v, v') \in \mathcal{E}} \mathbf{1} \big( \Phi_{\mathcal{D}}^s(\mathbf{f}_{v}^n) \neq \Phi_{\mathcal{D}}^s(\mathbf{f}_{v'}^n)\big),
\end{equation}
where $\mathbf{1}(\cdot)$ denotes the indicator function. A lower $\mathcal{L}_{\text{SPC}}$ score indicates a more spatially coherent assignment strategy.

\textit{Adaptive Strategy Selection (ASS).} The adaptive strategy selection function is ultimately defined as a weighted sum of the FAF and SPC terms to ensure robust strategy evaluation:
\begin{equation}
\label{eq:equ11}
s^* = \argmin_{s \in \mathcal{S}} \big( \mathcal{L}_{\text{FAF}}(s) + \lambda \mathcal{L}_{\text{SPC}}(s) \big).
\end{equation}
In our implementation, we set $\lambda=1$ to assign equal importance to the two terms. Finally, the strategy that achieves the optimal trade-off between feature fidelity and spatial coherence is selected for subsequent detector deployment, with its detailed workflow provided in the \textit{Supplementary Material}.

\section{High-Resolution Anomaly Detection Benchmarks}

In this section, we provide a detailed introduction to the high-resolution anomaly detection benchmarks adopted in our study, including the synthetic benchmarks MVTec-HD and VisA-HD, and the real-world benchmarks RealIAD-HD and MVTec 2. These benchmarks are constructed to reflect practical high-resolution anomaly detection tasks where anomalous regions are often subtle and may be missed under low-resolution settings, thereby presenting significant challenges. 

\begin{table}[t]
  \centering
  \renewcommand\arraystretch{1.0}
  \large
  \caption{Statistical overview of the synthetic high-resolution anomaly detection benchmarks. \textbf{``Large"} refers to anomalous images from the original dataset, while \textbf{``Tiny"} denotes images with synthetic subtle anomalies.}
  \resizebox{0.92\linewidth}{!}{
    \begin{tabular}{c|c|c|c|ccc|c}
    \bottomrule
    \multicolumn{1}{c|}{\multirow{3}[1]{*}{Benchmark}} & \multicolumn{1}{c|}{\multirow{3}[1]{*}{Dataset}} & \multirow{3}[1]{*}{Category} & \multicolumn{1}{c|}{\multirow{3}[1]{*}{\# Train}} & \multicolumn{3}{c|}{\# Test} & \multicolumn{1}{c}{\multirow{3}[1]{*}{\# Total}} \\
\cline{5-7}          &       & \multicolumn{1}{c|}{} &       & \multicolumn{1}{c}{\multirow{2}[1]{*}{\# Normal}} & \multicolumn{2}{c|}{\# Anomaly} &  \\
          &       & \multicolumn{1}{c|}{} &       &       & \multicolumn{1}{c}{\# Large} & \multicolumn{1}{c|}{\# Tiny} &  \\
    \hline
    \multicolumn{1}{c|}{\multirow{16}[1]{*}{MVTec-HD}} & \multicolumn{1}{c|}{\multirow{8}[1]{*}{MVTec-2K}} & Bottle & 146   & 50    & 47    & 69    & 312 \\
          &       & Capsule & 179   & 63    & 109   & 95    & 446 \\
          &       & Grid  & 230   & 55    & 57    & 42    & 384 \\
          &       & Hazelnut & 351   & 80    & 70    & 75    & 576 \\
          &       & Screw & 260   & 84    & 48    & 52    & 444 \\
          &       & Transistor & 183   & 87    & 40    & 39    & 349 \\
          &       & Wood  & 217   & 48    & 55    & 54    & 374 \\
          &       & \textbf{Total} & 1566  & 467   & 426   & 426   & 2885 \\
\cline{2-8}          & \multicolumn{1}{c|}{\multirow{8}[2]{*}{MVTec-4K}} & Bottle & 146   & 50    & 47    & 68    & 311 \\
          &       & Capsule & 179   & 63    & 109   & 88    & 439 \\
          &       & Grid  & 230   & 55    & 57    & 37    & 379 \\
          &       & Hazelnut & 351   & 80    & 70    & 73    & 574 \\
          &       & Screw & 260   & 84   & 48    & 55    & 447 \\
          &       & Transistor & 183   & 87    & 40    & 34    & 344 \\
          &       & Wood  & 217   & 48    & 55    & 41    & 361 \\
          &       & \textbf{Total} & 1566  & 467   & 426   & 396   & 2855 \\
    \hline
    \multicolumn{1}{c|}{\multirow{6}[1]{*}{VisA-HD}} & \multicolumn{1}{c|}{\multirow{6}[1]{*}{VisA-2K}} & Capsules & 497   & 101   & 100   & 96    & 794 \\
          &       & Fryum & 393   & 107   & 66    & 58    & 624 \\
          &       & Macaroni & 870   & 130   & 100   & 114   & 1214 \\
          &       & PCB   & 874   & 129   & 100   & 109   & 1212 \\
          &       & Pipe Fryum & 394   & 106   & 100   & 115   & 715 \\
          &       & \textbf{Total} & 3028  & 573   & 466   & 492   & 4559 \\
    \toprule
    \end{tabular}%
}
  \label{tab:table1}%
  \vspace{-0.2cm}
\end{table}%

\textbf{Synthetic Benchmark.} We develop a pipeline to extend the existing anomaly detection benchmarks, MVTec-AD \cite{bergmann2019mvtec} and VisA \cite{zou2022spot}, making them suitable for high-resolution tasks. Specifically, we first employ StableSR \cite{wang2024exploiting} to perform super-resolution on the images from the MVTec-AD \cite{bergmann2019mvtec} and VisA \cite{zou2022spot} datasets, thereby enhancing the resolution and image details. Then, we synthesize subtle anomalies by extracting ground-truth anomalous regions from the original datasets, applying downsampling, and seamlessly pasting them onto normal images using Poisson image editing \cite{perez2003poisson}. This approach produces visually coherent and contextually consistent synthetic anomalies. The resulting images form the ``Tiny" anomaly category, characterized by subtle anomalous regions that are difficult to detect at low resolutions. All synthetic anomalies are constrained to have a minimum bounding box side length of 16 pixels, ensuring that they represent physically plausible defects rather than pixel-level noise. This threshold is determined based on a statistical analysis of existing real-world benchmarks. Most images contain one anomaly region, while a small number include up to five. Each image undergoes manual inspection. We discard any samples exhibiting color bleeding, blurred boundaries, or any artifacts generated in the synthesis process, ensuring that the detected anomaly cues are grounded in the semantic features of the defects. The datasets are strictly partitioned to prevent any overlap between training and testing samples. Due to the inherently high computational complexity of high-resolution anomaly detection, we remove certain redundant categories from the original datasets to reduce the experimental cost. For example, in the MVTec-AD dataset \cite{bergmann2019mvtec}, the categories Wood, Tile, Carpet, and Leather exhibit highly similar characteristics. Therefore, we retain only the representative category, Wood, for our study. The final MVTec-HD benchmark includes seven representative categories: Bottle, Capsule, Grid, Hazelnut, Screw, Transistor, and Wood. We then construct two datasets with resolutions of $2048 \times 2048$ and $4096 \times 4096$, referred to as MVTec-2K and MVTec-4K, respectively. For the VisA-HD benchmark, we select five representative categories: Capsules, Fryum, Macaroni, PCB, and Pipe Fryum. These are processed to create a dataset at $2048 \times 2048$ resolution, referred to as VisA-2K. Table \ref{tab:table1} presents a statistical overview of these synthetic benchmarks. The number of synthetic subtle anomaly samples is controlled to approximately match that of the original anomaly samples, based on the assumption that subtle anomalies and larger anomalies occur with similar frequencies. 

\textbf{Real-world Benchmark.} The Real-IAD dataset \cite{wang2024real} comprises multi-view high-resolution images captured from real industrial scenarios, with resolutions ranging from 2000 to 5000 pixels. For the RealIAD-HD benchmark, we select three representative product categories characterized by particularly subtle anomalies: Bottle Cap, Mint, and USB Adaptor. These categories exhibit the smallest average defect areas in the Real-IAD dataset \cite{wang2024real}, making them particularly challenging for anomaly detection under low-resolution settings. In contrast, other categories contain few or no subtle anomalies and are therefore excluded from this study. To standardize the evaluation process, we utilize top-view images exclusively and uniformly resize them to $2048 \times 2048$ pixels, forming the RealIAD-2K dataset. Additionally, we incorporate the recently released MVTec 2 dataset \cite{heckler2026mvtec} into our evaluation. The MVTec 2 dataset comprises eight categories of high-resolution images from real-world industrial scenarios, with an average resolution approaching 2K. The dataset encompasses several critical challenges, including subtle anomalies, low-contrast anomalies, illumination variations, and high intra-class variability of normal data, all of which pose significant hurdles for unsupervised anomaly detection methods.

\begin{figure}[t]
  \centering
   \includegraphics[width=\linewidth]{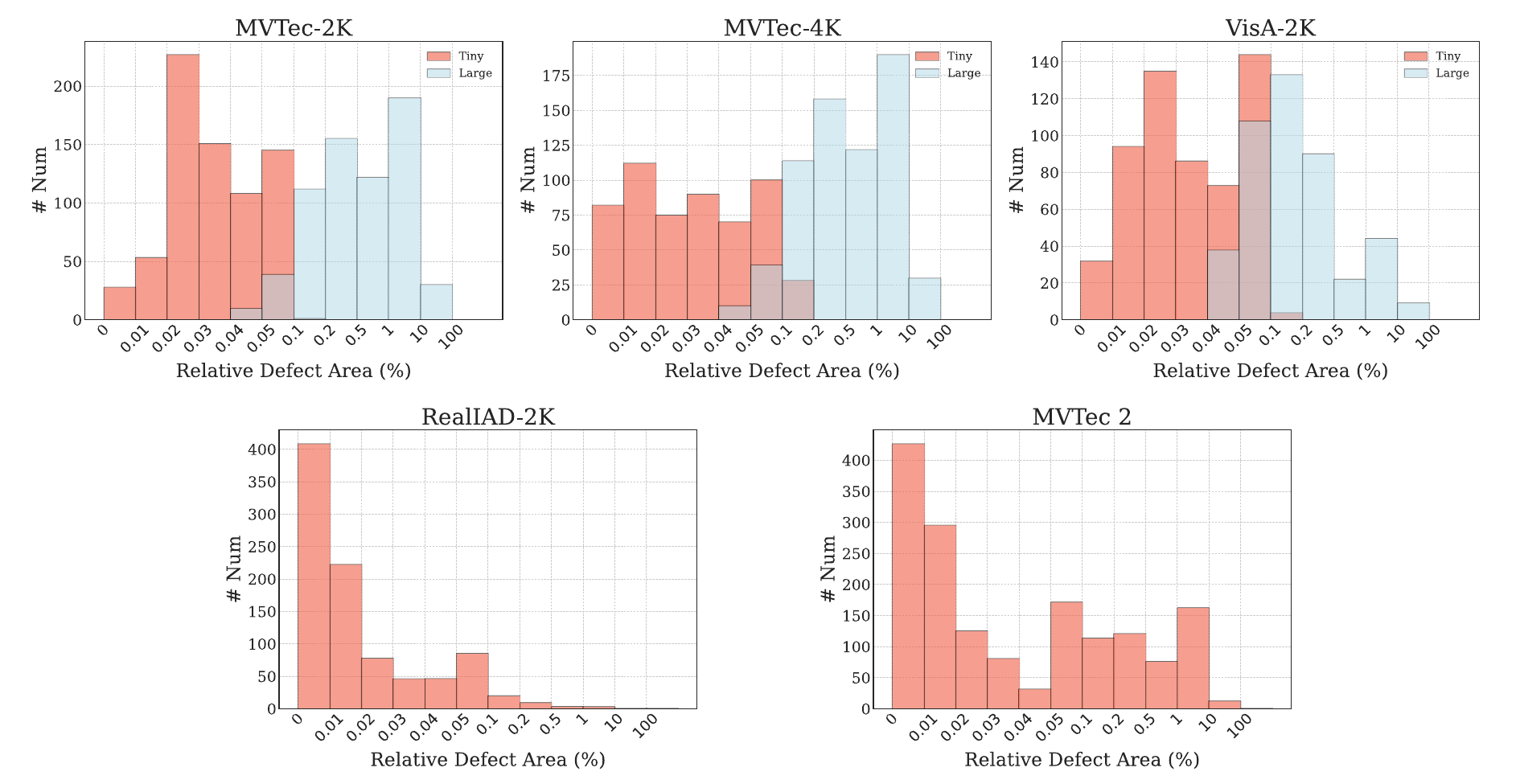}
   \caption{Statistics of the Relative Defect Area for different datasets.}
   \label{fig:fig5}
   \vspace{-0.2cm}
\end{figure}

Figure \ref{fig:fig5} presents the statistical results of the Relative Defect Area \cite{lehr2024ad3} for each dataset, defined as the proportion of anomalous pixels to the total number of pixels. The statistical results indicate that the MVTec-HD and VisA-HD benchmarks focus on detecting anomalous regions across various scales, whereas RealIAD-HD and MVTec 2 contain more subtle anomalies, making them particularly suitable for evaluating small-defect detection capability.

\section{Experiment}

To evaluate the effectiveness of the proposed HiAD framework and the reliability of the constructed high-resolution benchmarks, we conduct comprehensive experiments. This section presents the implementation details of HiAD, the evaluation metrics, quantitative and qualitative results under different task settings, and systematic ablation studies.

\begin{table*}[t]
  \centering
  \renewcommand\arraystretch{1.0}
  \setlength{\tabcolsep}{3mm}
  \caption{Summary of hyperparameters for different anomaly detection methods integrated into HiAD.}
    \small
    \begin{tabular}{l|l|c|l|c|c}
    \bottomrule
    Method & Method Type & Patch Size & Backbone & Intermediate Layers & \multicolumn{1}{c}{\makecell{Maximum \\ Iterations}} \\
    \hline
    PatchCore \cite{roth2022towards} & Deep Feature Embedding &$512 \times 512$& WideResNet-50‌ & Layers \{2, 3\}      & \multicolumn{1}{c}{--} \\
    PaDiM \cite{defard2021padim} & Deep Feature Embedding &$512 \times 512$& WideResNet-50 & Layers \{2, 3\}     & \multicolumn{1}{c}{--} \\
    FastFlow \cite{yu2021fastflow} & Normalizing Flow &$512 \times 512$& WideResNet-50 & Layers \{2, 3\}   & 30,000 \\
    ViTAD \cite{zhang2023exploring} & Reconstruction &$384 \times 384$& ViT-B/16 & Layers \{3, 6, 9\}     & 20,000 \\
    Dinomaly \cite{guo2025dinomaly} & Reconstruction &$512 \times 512$& DINOv3 ViT-B/16 & Layers \{3--10\}     & 5,000 \\
    RealNet \cite{zhang2024realnet}& Anomaly Synthesis \& Reconstruction &$512 \times 512$& WideResNet-50  & Layers \{2, 3\}   & 30,000 \\
    RD++ \cite{tien2023revisiting} & Anomaly Synthesis \& Distillation  &$512 \times 512$& WideResNet-50 & Layers \{2, 3\}    & 5,000 \\
    DeSTSeg \cite{zhang2023destseg}& Anomaly Synthesis \& Distillation  &$512 \times 512$& ResNet-18  & Layers \{2, 3\}   & 3,000 \\
    \toprule
    \end{tabular}%
  \label{tab:table2}%
\end{table*}%

\subsection{Implementation Details}

We incorporate eight representative anomaly detection methods into HiAD, including PatchCore \cite{roth2022towards}, ViTAD \cite{zhang2023exploring}, Dinomaly \cite{guo2025dinomaly}, DeSTSeg \cite{zhang2023destseg}, RealNet \cite{zhang2024realnet}, FastFlow \cite{yu2021fastflow}, RD++ \cite{tien2023revisiting}, and PaDiM \cite{defard2021padim}. Table \ref{tab:table2} presents the detailed hyperparameter configurations for these methods. For each method, we adapt the intermediate layers of the backbones to better accommodate high-resolution images. The image-level anomaly score is computed from the overall anomaly map $\mathcal{A}$, typically using its maximum value, following the settings used in different methods under low-resolution scenarios. We always use the same anomaly detection method for both high-resolution and low-resolution branches. For each method, the largest feasible patch size is selected to minimize the number of image patches. Specifically, for ViTAD \cite{zhang2023exploring}, we set both the patch size and the resolution of the low-resolution branch to $384 \times 384$, aligning with the pre-training resolution of its backbone. For other methods, they are set to $512 \times 512$. We set the stride size to be equal to the patch size, indicating that there is no overlap between image patches. The original image is progressively downsampled by a factor of $2^k$ until it reaches a resolution of $1024 \times 1024$ for multi-resolution feature extraction, and a set of equal weights is used for feature fusion. For deep feature embedding-based methods such as PatchCore \cite{roth2022towards} and PaDiM \cite{defard2021padim}, we employ the O2O assignment strategy, which reduces the size of each memory bank to enhance inference speed. For other methods, we employ the proposed ASS for strategy selection. We use KMeans \cite{macqueen1967some} as the clustering algorithm in both SCA and RA. For the MVTec-4K dataset, the detector pool size $M$ is set to 8; for all other datasets, $M$ is set to 4. It should be noted that when $1< M <\boldsymbol{I} \times \boldsymbol{J}$, A2O and O2O are no longer applicable. In this case, NA provides a more general assignment formulation, with A2O and O2O being its special cases at the two extremes. Accordingly, ASS selects the strategy from $\{\text{NA},\text{SCA},\text{RA}\}$. In addition, we randomly partition 20\% of the training images as a validation set, which serves as a query set for assignment strategy selection and is also utilized for anomaly map normalization. Specifically, each detector in both branches independently applies min-max normalization using validation-set statistics and clips the resulting anomaly maps to $[0,1]$. The ASS and assignment clustering are always conducted before training, and the results are saved for subsequent training and inference.

\subsection{Evaluation Metrics}

\begin{figure}[t]
  \centering
   \includegraphics[width=0.95\linewidth]{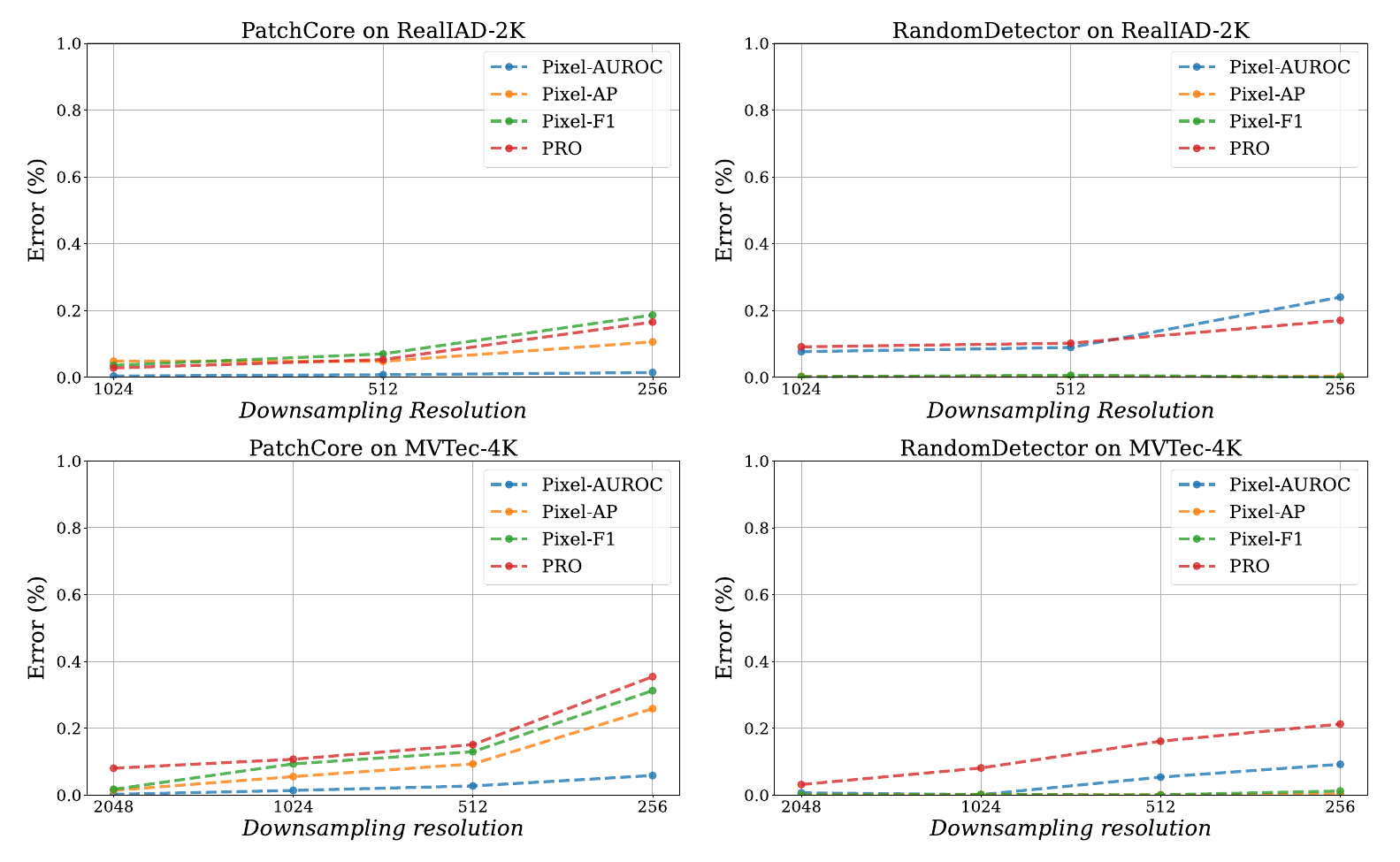}
   \caption{The error curves of pixel-level metrics computed at different downsampling resolutions compared to the full-resolution evaluation. Specifically, the \textit{RandomDetector} outputs anomaly scores following a uniform distribution within the range of 0 to 1, simulating an extreme scenario of random prediction.}
   \label{fig:fig6}
\end{figure}

We evaluate the performance of HiAD at both the image and pixel levels. Following previous work \cite{bergmann2019mvtec, zou2022spot}, we adopt the Area Under the Receiver Operating Characteristic Curve (I-AUC) to measure image-level anomaly detection accuracy, and use Pixel-AUROC (P-AUC), Average Precision (P-AP), F1-score-max (P-F1), and Per-Region Overlap (PRO) for evaluating pixel-level anomaly localization performance. However, directly computing pixel-level metrics on high-resolution images is computationally prohibitive, as it requires sorting over billions of anomaly scores. To address this, we follow the common practice in the community by computing pixel-level metrics on appropriately downsampled anomaly maps and ground-truth masks, which serves as a highly efficient and robust approximation of full-resolution evaluation. Unlike neural network-based computational processes, non-parametric pixel-wise computations (such as AUROC and AP) rely on rank-order statistics. Proper downsampling has minimal impact on the proportion of correctly and incorrectly predicted pixels, making these metrics highly robust to spatial scaling. To explicitly prevent the loss of microscopic defects during this approximation, we resize the predictions and masks to $512 \times 512$. Geometrically, even in the highest-resolution MVTec-4K dataset, synthetic anomalies maintain a minimum bounding box side length of 16 pixels, ensuring they are not lost during downsampling to this resolution. Figure \ref{fig:fig6} further confirms this structural preservation: the deviations of all pixel-level metrics remain rigorously constrained below 0.1\% and 0.2\% for evaluations on 2K and 4K images, respectively. Notably, for several benchmarks containing subtle anomalies, such as Real-IAD \cite{wang2024real} and MVTec 2 \cite{heckler2026mvtec}, the evaluation resolution typically adopted by the community is $256 \times 256$ \cite{wang2024real, heckler2026mvtec, guo2025dinomaly}. In contrast, our evaluation resolution is significantly higher than existing standards.

\subsection{Experimental Results}

\begin{table*}[t]
  \centering
  \renewcommand\arraystretch{1.0}
  \caption{Comparison of anomaly detection results at different resolutions across various methods on the 2K datasets. Best results are highlighted in \textbf{bold}, while second-best results are \underline{underlined}.}
  \resizebox{\linewidth}{!}{
  \scriptsize
\begin{tabular}{c|c|ccccc|ccccc|ccccc}
    \bottomrule
    \multicolumn{1}{c|}{\multirow{2}[3]{*}{\textbf{Resolution}}} & \multirow{2}[3]{*}{\textbf{Method}} & \multicolumn{15}{c}{\textbf{High-Resolution (With HiAD)}} \\
\cline{3-17}          & \multicolumn{1}{c|}{} & \multicolumn{5}{c|}{\textbf{MVTec-2K}} & \multicolumn{5}{c|}{\textbf{VisA-2K}} & \multicolumn{5}{c}{\textbf{RealIAD-2K}}  \\
\cline{3-17}          & \multicolumn{1}{c|}{} & \multicolumn{1}{c}{\textbf{I-AUC}} & \multicolumn{1}{c}{\textbf{P-AUC}} & \multicolumn{1}{c}{\textbf{P-AP}} & \multicolumn{1}{c}{\textbf{P-F1}} & \multicolumn{1}{c|}{\textbf{PRO}} & \multicolumn{1}{c}{\textbf{I-AUC}} & \multicolumn{1}{c}{\textbf{P-AUC}} & \multicolumn{1}{c}{\textbf{P-AP}} & \multicolumn{1}{c}{\textbf{P-F1}} & \multicolumn{1}{c|}{\textbf{PRO}} & \multicolumn{1}{c}{\textbf{I-AUC}} & \multicolumn{1}{c}{\textbf{P-AUC}} & \multicolumn{1}{c}{\textbf{P-AP}} & \multicolumn{1}{c}{\textbf{P-F1}} & \multicolumn{1}{c}{\textbf{PRO}}  \\
    \hline
    \multicolumn{1}{c|}{\multirow{6}[5]{*}{2048 $\times$ 2048}}  
          & PatchCore \cite{roth2022towards} & \textbf{98.44} & 97.79 & 65.84 & 64.07 & \underline{97.39} & \textbf{98.82} & 99.66 & \textbf{64.07} & \textbf{63.98} & \underline{98.74} & \textbf{98.37} & \textbf{99.89} & 45.16 & 47.34 & \underline{99.10}  \\
          & Dinomaly \cite{guo2025dinomaly} & 97.17	 & \textbf{99.07} & \textbf{72.98} & \textbf{69.10} & \textbf{97.58} & \underline{98.57} & \textbf{99.74} & 56.72 & 57.89 & \textbf{98.86} & 94.26 & \underline{99.86} & 39.62 & 40.88 &  \textbf{99.23} \\
          & ViTAD \cite{zhang2023exploring} & 96.45 & \underline{99.06} & 64.78 & 64.09 & 97.30 & 95.19 & 99.56 & 52.47 & 54.71 & 96.99 & 94.47 & 99.79 & 36.83 & 41.88 & 98.07  \\
          & DeSTSeg \cite{zhang2023destseg} & \underline{97.93} & 97.33 & 70.70 & 68.33  & 96.51 & 94.22 & 98.70 & \underline{63.79} & \underline{62.52} & 91.59 & 93.50  & 99.46 & \textbf{54.67} & \textbf{53.96} & 95.55  \\
          & RealNet \cite{zhang2024realnet} & 96.18 & 96.85 & 63.48 & 62.21 & 91.59  & 92.51 & 98.87  & 54.42 & 56.44 & 88.87 & 93.02 & 99.63 & 46.84 & 48.86 & 92.71  \\
          & FastFlow \cite{yu2021fastflow} & 91.71 & 98.52  & 54.22 & 55.44 &   94.48 & 94.85 & 99.58 & 51.11 & 54.40 & 95.81  & 93.66 & 99.32 & 25.97 & 32.82 & 96.79  \\                                                      
          & RD++ \cite{tien2023revisiting}  & 91.85  & 96.74 & 59.41  & 59.87 & 95.86 & 93.98 & 99.34  & 51.68  & 55.79  & 96.34  & 88.03 & 98.87 & 29.03 & 36.27 & 94.02  \\
          & PaDiM \cite{defard2021padim} & 86.22 & 98.13 & 44.43 & 48.50  & 95.95 & 84.56 & 98.79 & 26.25 & 33.33 & 94.36 & 89.67 & 99.60  & 20.74 & 28.22 & 97.98  \\
    \hline
    \multicolumn{1}{c|}{\multirow{6}[5]{*}{1024 $\times$ 1024}} 
          & PatchCore \cite{roth2022towards} & 95.68 & 97.58 & 65.73 & 64.17 & 96.28 & 95.95 & 99.65 & 59.91 & 61.15 & 98.00    & \underline{95.50}  & 99.73 & 35.64 & 40.24 & 97.21\\
          & Dinomaly \cite{guo2025dinomaly} & 94.80 & 98.90 & 69.95 & 67.16 & 96.07 & 96.61	&\underline{99.69} & 54.69 & 55.83 & 96.37 & 92.01 & 99.55 & 23.60	& 33.22 &  96.34 \\
          & ViTAD \cite{zhang2023exploring} & 93.94 & 98.97 & 63.95 & 64.37 & 94.56 & 91.37  & 99.31 & 45.68 & 50.01 & 94.73  & 92.49 & 99.61 & 20.66 & 29.77 & 96.46  \\
          & DeSTSeg \cite{zhang2023destseg} & 92.90 & 98.05    & \underline{71.14}   & \underline{68.93} & 95.43 &  91.36 & 99.07 & 59.86 & 58.11 & 90.68 & 88.19 & 98.86 & \underline{46.91} & \underline{50.04} & 90.44  \\
          & RealNet \cite{zhang2024realnet} & 92.64 & 95.92 & 61.47 & 61.25 & 87.28 & 91.73 & 99.16 & 55.34 & 56.92 & 89.65 & 89.24 & 99.47 & 41.25 & 45.58  & 84.43  \\
          & FastFlow \cite{yu2021fastflow} & 90.08 & 98.11 & 52.41 & 54.34 & 93.84 & 90.62 & 99.57 & 51.52 & 53.62 & 95.05 & 92.12 & 99.33 & 21.53 & 29.83 & 94.77  \\
          & RD++ \cite{tien2023revisiting}  & 90.24  & 96.63 & 59.32 & 59.84 & 95.23 & 91.51 & 98.94 & 48.10 & 52.97 & 95.32 & 82.11 & 97.94 & 24.82 & 32.46 & 88.96 \\
          & PaDiM \cite{defard2021padim} & 86.01 & 98.22 & 45.55 & 49.39 & 95.14 & 83.92 & 98.79 & 26.20  & 33.46 & 94.15 & 88.12 & 99.35 & 19.99 & 27.43 & 95.33 \\
    \hline
    \multicolumn{1}{c|}{\multirow{2}[3]{*}{\textbf{Resolution}}} & \multirow{2}[3]{*}{\textbf{Method}} & \multicolumn{15}{c}{\textbf{Low-Resolution (Without HiAD)}} \\
\cline{3-17}          & \multicolumn{1}{c|}{} & \multicolumn{5}{c|}{\textbf{MVTec-2K}} & \multicolumn{5}{c|}{\textbf{VisA-2K}} & \multicolumn{5}{c}{\textbf{RealIAD-2K}}  \\
\cline{3-17}          & \multicolumn{1}{c|}{} & \multicolumn{1}{c}{\textbf{I-AUC}} & \multicolumn{1}{c}{\textbf{P-AUC}} & \multicolumn{1}{c}{\textbf{P-AP}} & \multicolumn{1}{c}{\textbf{P-F1}} & \multicolumn{1}{c|}{\textbf{PRO}} & \multicolumn{1}{c}{\textbf{I-AUC}} & \multicolumn{1}{c}{\textbf{P-AUC}} & \multicolumn{1}{c}{\textbf{P-AP}} & \multicolumn{1}{c}{\textbf{P-F1}} & \multicolumn{1}{c|}{\textbf{PRO}} & \multicolumn{1}{c}{\textbf{I-AUC}} & \multicolumn{1}{c}{\textbf{P-AUC}} & \multicolumn{1}{c}{\textbf{P-AP}} & \multicolumn{1}{c}{\textbf{P-F1}} & \multicolumn{1}{c}{\textbf{PRO}} \\
    \hline
    \multicolumn{1}{c|}{\multirow{5}[5]{*}{512 $\times$ 512}} 
          & PatchCore \cite{roth2022towards} & \textbf{91.26} & 97.47 & 64.15 & 62.90  & \underline{94.25} & \underline{90.79} & \textbf{99.55} & \underline{56.32} & \textbf{58.78} & \textbf{96.87} & \textbf{91.44} & \textbf{99.51} & 25.53 & 33.43 & \textbf{95.09}  \\
          & Dinomaly \cite{guo2025dinomaly} & \underline{90.96} & \underline{98.40}	 & \underline{66.63} & \underline{64.56} & 93.76 & \textbf{92.32} & 98.95 & 50.87 & 50.94 & 94.50 & 88.95 & 98.91 & 19.37 & 29.12 & 93.10  \\
          & DeSTSeg \cite{zhang2023destseg} & 85.04 & 97.00    & \textbf{67.68} & \textbf{66.64} & 85.76 & 82.80  & 98.43 & \textbf{56.71} & \underline{56.78} & 80.90  & 86.21 & 98.69 & \textbf{33.09} & \textbf{40.43} & 81.64 \\
          & RealNet \cite{zhang2024realnet} & 86.20  & 95.84 & 57.52 & 57.90  & 76.63 & 86.66 & 98.75 & 50.76 & 53.39 & 75.18 & 84.21 & 99.21 & \underline{30.55} & \underline{37.65} & 69.38 \\
          & FastFlow \cite{yu2021fastflow} & 88.47 & 98.33 & 54.85 & 56.81 & 91.00    & 83.35 & 98.32 & 45.75 & 49.59 & 86.19 & \underline{91.23} & 98.81 & 20.50  & 31.27 & 90.82  \\
          & RD++ \cite{tien2023revisiting}  & 86.14 & 96.52 & 56.12 & 57.59 & 90.95 & 89.97 & \underline{99.28} & 49.80  & 53.83 & \underline{94.62} & 79.03 & 97.58 & 20.58 & 26.78 & 84.55  \\
          & PaDiM \cite{defard2021padim} & 85.32 & 98.12 & 45.65 & 49.07 & \textbf{94.89} & 82.27 & 98.83 & 26.54 & 33.68 & 93.97 & 87.23 & \underline{99.25} & 20.13 & 27.16 & \underline{94.33} \\
    \hline
    \multicolumn{1}{c|}{384 $\times$ 384} & ViTAD \cite{zhang2023exploring}  & 84.13 & \textbf{98.42} & 58.59 & 60.18 & 80.58 & 83.06 & 97.57 & 42.12 & 46.72 & 82.91 & 87.43 & 97.44 & 11.87 & 21.13 & 84.58  \\
    \hline
    \multicolumn{1}{c|}{\multirow{5}[5]{*}{256 $\times$ 256}} 
          & PatchCore \cite{roth2022towards} & 83.96 & 98.30  & 54.55 & 56.48 & 84.11 & 78.37 & 98.11 & 49.95 & 52.09 & 85.43 & 84.58 & 97.71 & 10.10  & 18.22 & 83.98  \\
          & Dinomaly \cite{guo2025dinomaly} & 87.95 & 98.28 & 55.38 & 57.15 & 83.35 & 83.75 & 97.05 & 41.74 & 45.77 & 83.51 & 85.63 & 98.30 & 12.71 & 17.87 &  86.26 \\
          & DeSTSeg \cite{zhang2023destseg} & 81.02 & 97.92 & 64.72 & 63.62 & 72.33 & 76.49 & 96.49 & 44.26 & 47.82 & 64.37 & 74.32 & 94.70  & 14.52 & 25.13 & 60.92  \\
          & RealNet \cite{zhang2024realnet} & 77.75 & 97.11 & 54.25 & 55.11 & 59.07 & 75.48 & 97.65 & 46.54 & 49.31 & 64.01 & 67.77 & 95.11 & 10.55 & 21.58 & 45.94  \\
          & FastFlow \cite{yu2021fastflow} & 81.19 & 97.71 & 48.38 & 50.66 & 76.53 & 76.42 & 96.14 & 44.03 & 46.63 & 76.87 & 85.04 & 96.16 & 10.04 & 18.45 & 79.69  \\
          & RD++ \cite{tien2023revisiting}  & 81.96 & 98.20  & 50.99 & 53.48 & 84.61 & 74.61 & 96.87 & 41.74 & 44.72 & 77.97 & 68.21 & 93.12 & 8.78  & 16.09 & 64.75 \\
          & PaDiM \cite{defard2021padim} & 83.40  & 98.38 & 47.32 & 51.82 & 88.11 & 76.56 & 97.71 & 28.88 & 34.76 & 83.17 & 81.80  & 97.53 & 8.00  & 15.33 & 83.38  \\
    \toprule   
    \end{tabular}%
    }
  \label{tab:table3}%
\end{table*}%

\begin{table}[t]
  \centering
  \renewcommand\arraystretch{1.25}
   \caption{Comparison of anomaly detection results at different resolutions across various methods on the MVTec-4K and MVTec 2 datasets.}
   \resizebox{0.97\linewidth}{!}{
    \begin{tabular}{c|c|ccccc}
    \bottomrule
    \multicolumn{7}{c}{MVTec-4K} \\
    \hline
    Resolution & \multicolumn{1}{c|}{Method} & \multicolumn{1}{c}{\textbf{I-AUC}} & \multicolumn{1}{c}{\textbf{P-AUC}} & \multicolumn{1}{c}{\textbf{P-AP}} & \multicolumn{1}{c}{\textbf{P-F1}} & \multicolumn{1}{c}{\textbf{PRO}} \\
    \hline
    \multicolumn{1}{c|}{\multirow{3}[3]{*}{\makecell{4096 $\times$ 4096 \\ (With HiAD)}}} & \multicolumn{1}{c|}{Dinomaly \cite{guo2025dinomaly}} & \underline{96.33} & \textbf{98.99} & \underline{69.84} & \underline{67.28} & \textbf{97.71} \\
    \multicolumn{1}{c|}{} & \multicolumn{1}{c|}{DeSTSeg \cite{zhang2023destseg}} & \textbf{98.81} & 98.14 & \textbf{75.53} & \textbf{71.60} & 96.34\\
    \multicolumn{1}{c|}{} & \multicolumn{1}{c|}{PatchCore \cite{roth2022towards}} & 95.32 & 97.85 & 63.23 & 62.09 & \underline{97.30} \\
    \multicolumn{1}{c|}{} & \multicolumn{1}{c|}{ViTAD \cite{zhang2023exploring}} & 95.44 & \underline{98.94} & 63.41 & 63.44 & 97.11 \\
    \hline
    \multicolumn{1}{c|}{\multirow{3}[1]{*}{\makecell{512 $\times$ 512 \\ (Without HiAD)}}} & \multicolumn{1}{c|}{Dinomaly \cite{guo2025dinomaly}} & 88.40  & 97.01 & 62.53 & 60.59 & 94.44 \\
    \multicolumn{1}{c|}{} & \multicolumn{1}{c|}{DeSTSeg \cite{zhang2023destseg}} & 84.60  & 96.95 & 66.82 & 64.84 & 85.20 \\
    \multicolumn{1}{c|}{} & \multicolumn{1}{c|}{PatchCore \cite{roth2022towards}} & 85.62 & 96.55 & 57.91 & 58.43 & 94.17 \\
     \hline
    \multicolumn{1}{c|}{\multirow{2}[1]{*}{\makecell{384 $\times$ 384 \\ (Without HiAD)}}} & \multicolumn{1}{c|}{\multirow{2}[1]{*}{ViTAD \cite{zhang2023exploring}}} & \multirow{2}[1]{*}{82.32} & \multirow{2}[1]{*}{96.69} & \multirow{2}[1]{*}{56.60} & \multirow{2}[1]{*}{58.95} & \multirow{2}[1]{*}{80.20} \\
    \multicolumn{1}{c|}{} &       &       &       &       &       &  \\
    \hline
    \multicolumn{7}{c}{MVTec 2} \\
    \hline
    Resolution & \multicolumn{1}{c|}{Method} & \multicolumn{1}{c}{\textbf{I-AUC}} & \multicolumn{1}{c}{\textbf{P-AUC}} & \multicolumn{1}{c}{\textbf{P-AP}} & \multicolumn{1}{c}{\textbf{P-F1}} & \multicolumn{1}{c}{\textbf{PRO}} \\
    \hline
    \multicolumn{1}{c|}{\multirow{3}[3]{*}{\makecell{Original \\ (With HiAD)}}} & \multicolumn{1}{c|}{Dinomaly \cite{guo2025dinomaly}} & \textbf{80.10}  & \textbf{94.92} & \textbf{34.52} & \textbf{39.41} & \textbf{79.32} \\
    \multicolumn{1}{c|}{} & \multicolumn{1}{c|}{DeSTSeg \cite{zhang2023destseg}} & \underline{77.78} & 89.37 & 29.18 & 32.84 & 73.15 \\
    \multicolumn{1}{c|}{} & \multicolumn{1}{c|}{PatchCore \cite{roth2022towards}} & 76.69 & 90.88 & 21.83 & 28.10  & \underline{76.64} \\
    \multicolumn{1}{c|}{} & \multicolumn{1}{c|}{ViTAD \cite{zhang2023exploring}} & 75.72 & 90.06 & \underline{29.53} & \underline{35.25} & 75.58 \\
    \hline
    \multicolumn{1}{c|}{\multirow{3}[1]{*}{\makecell{512 $\times$ 512 \\ (Without HiAD)}}} & \multicolumn{1}{c|}{Dinomaly \cite{guo2025dinomaly}} & 69.47 & \underline{91.96} & 28.30  & 34.69 & 61.56 \\
    \multicolumn{1}{c|}{} & \multicolumn{1}{c|}{DeSTSeg \cite{zhang2023destseg}} & 68.30  & 84.64 & 22.61 & 25.52 & 53.31 \\
    \multicolumn{1}{c|}{} & \multicolumn{1}{c|}{PatchCore \cite{roth2022towards}} & 70.83 & 88.02 & 18.97 & 23.77 & 56.07 \\
    \hline
    \multicolumn{1}{c|}{\multirow{2}[1]{*}{\makecell{384 $\times$ 384 \\ (Without HiAD)}}} & \multicolumn{1}{c|}{\multirow{2}[1]{*}{ViTAD \cite{zhang2023exploring}}} & \multirow{2}[1]{*}{65.03} & \multirow{2}[1]{*}{86.79} & \multirow{2}[1]{*}{20.32} & \multirow{2}[1]{*}{26.67} & \multirow{2}[1]{*}{51.62} \\
    \multicolumn{1}{c|}{} &       &       &       &       &       &  \\
    \toprule
    \end{tabular}%
   }
  \label{tab:table4}%
\end{table}%

\begin{figure*}[t]
  \centering
   \includegraphics[width=\linewidth]{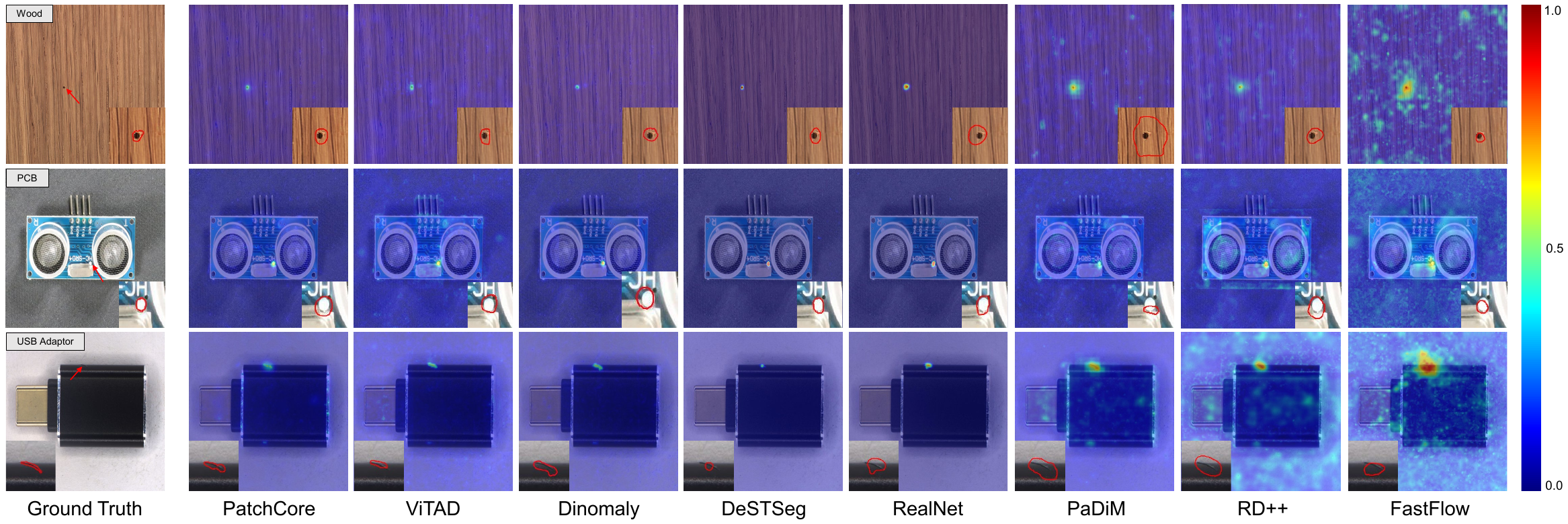}
   \caption{Qualitative results of HiAD using different anomaly detection methods for subtle anomaly detection at a detection resolution of $2048 \times 2048$.}
   \label{fig:fig7}
   \vspace{-0.2cm}
\end{figure*}

\textit{1) High-resolution Anomaly Detection Results.} Table \ref{tab:table3} presents the performance on the 2K datasets under different detection resolutions. Due to the pervasive presence of subtle anomalies in the datasets, all baseline methods exhibit missed detections at low resolutions, leading to suboptimal performance. When employing HiAD for high-resolution detection, these methods exhibit significant performance improvements on both real and synthetic benchmarks, highlighting the necessity of high-resolution anomaly detection. Among all methods, PatchCore \cite{roth2022towards}, Dinomaly \cite{guo2025dinomaly}, ViTAD \cite{zhang2023exploring}, and DeSTSeg \cite{zhang2023destseg} achieve superior anomaly detection performance, demonstrating consistent effectiveness across all benchmarks. Table \ref{tab:table4} presents the anomaly detection results of HiAD on the MVTec-4K and MVTec 2 datasets. Compared with 2K images, 4K images feature richer texture details and larger spans of anomalous regions, making anomaly detection more challenging. HiAD achieves satisfactory performance even at a resolution of $4096 \times 4096$, demonstrating its excellent scalability and potential for real-world ultra-high-resolution anomaly detection tasks. Furthermore, HiAD achieves significant performance gains on the more challenging MVTec 2 dataset, with the image-level AUROC increasing by approximately 10\% compared to its low-resolution counterparts. This underscores the feasibility of HiAD for complex industrial scenarios. Figure \ref{fig:fig7} presents the qualitative results of HiAD in localizing subtle anomalous regions. HiAD can effectively identify imperceptible anomalies with a relative defect area of less than 0.01\%, significantly enhancing the detection capabilities of existing anomaly detection methods.

\begin{table}[t]
  \centering
  \renewcommand\arraystretch{1.2}
  \caption{Comparison of HiAD with alternative anomaly detection methods on the MVTec-4K and MVTec 2 datasets.}
  \resizebox{\linewidth}{!}{
    \begin{tabular}{c|c|ccccc}
     \bottomrule
    \multicolumn{7}{c}{MVTec-4K} \\
    \hline
    Method & Resolution & \multicolumn{1}{c}{\textbf{I-AUC}} & \multicolumn{1}{c}{\textbf{P-AUC}} & \multicolumn{1}{c}{\textbf{P-AP}} & \multicolumn{1}{c}{\textbf{P-F1}} & \multicolumn{1}{c}{\textbf{PRO}} \\
    \hline
    EfficientAD-S \cite{batzner2024efficientad} & 1792 $\times$ 1792 & 78.06 & 83.96 & 25.33 & 31.52 & 78.17 \\
    EfficientAD-M \cite{batzner2024efficientad} & 1280 $\times$ 1280 & 85.36 & 91.01 & 35.28 & 38.70  & 85.88 \\
    VarAD \cite{cao2025varad} & 1280 $\times$ 1280 & 85.19 & 97.56 & 58.18 & 60.13 & 91.32 \\
    \hline
    Tiling{\fontsize{6}{3}\selectfont (DeSTSeg)} \cite{rolih2024divide} & 4096 $\times$ 4096 & 95.38 & 94.65 & 45.76 & 50.26 & 86.31 \\
    Tiling{\fontsize{6}{3}\selectfont (Dinomaly)} \cite{rolih2024divide} & 4096 $\times$ 4096 & 93.23 & 97.12 & 61.22 & 61.01 & \underline{96.49} \\
    \hline
    \multirow{2}[4]{*}{HiAD{\fontsize{6}{3}\selectfont (DeSTSeg)}} & 1024 $\times$ 1024 & 92.63 & 97.70 & 72.59 & 70.22 & 93.84 \\
    \multicolumn{1}{c|}{} & 2048 $\times$ 2048 & \underline{97.57}  & 98.27 & \underline{73.65} & \underline{70.49} & 95.65 \\
    \multicolumn{1}{c|}{} & 4096 $\times$ 4096 & \textbf{98.81} & 98.14 & \textbf{75.53} & \textbf{71.60} &  96.34 \\
    \hline
    \multirow{2}[4]{*}{HiAD{\fontsize{6}{3}\selectfont (Dinomaly)}} & 1024 $\times$ 1024 & 91.91 & 97.93 & 63.68 & 60.46 & 96.12 \\
    \multicolumn{1}{c|}{} & 2048 $\times$ 2048 & 93.89 & \underline{98.42} & 64.20  & 62.55 & 96.07 \\
    \multicolumn{1}{c|}{} & 4096 $\times$ 4096 & 96.33 & \textbf{98.99} & 69.84 & 67.28 & \textbf{97.71} \\
    \hline 
    \multicolumn{7}{c}{MVTec 2} \\
    \hline
    Method & Resolution & \multicolumn{1}{c}{\textbf{I-AUC}} & \multicolumn{1}{c}{\textbf{P-AUC}} & \multicolumn{1}{c}{\textbf{P-AP}} & \multicolumn{1}{c}{\textbf{P-F1}} & \multicolumn{1}{c}{\textbf{PRO}} \\
    \hline
    EfficientAD-S \cite{batzner2024efficientad} & 1792 $\times$ 1792 & 71.14 & 81.74 & 10.97 & 16.43 & 60.75 \\
    EfficientAD-M \cite{batzner2024efficientad} & 1280 $\times$ 1280 & 70.93 & 80.56 & 13.69 & 18.07 & 64.37 \\
    VarAD \cite{cao2025varad} & 1280 $\times$ 1280 & 69.41 & 87.99 & 22.60  & 28.49 & 64.48 \\
    \hline
    Tiling{\fontsize{6}{3}\selectfont (DeSTSeg)} \cite{rolih2024divide} & Original & 68.91 & 87.96 & 22.82 & 27.41 & 63.59 \\
    Tiling{\fontsize{6}{3}\selectfont (Dinomaly)} \cite{rolih2024divide} & Original & 73.60  & \underline{92.60}  & \underline{31.93} & \underline{35.07} & \underline{78.01} \\
    \hline
    HiAD{\fontsize{6}{3}\selectfont (DeSTSeg)} & Original & \underline{77.78} & 89.37 & 29.18 & 32.84 & 73.15 \\
    HiAD{\fontsize{6}{3}\selectfont (Dinomaly)} & Original & \textbf{80.10}  & \textbf{94.92} & \textbf{34.52} & \textbf{39.41} & \textbf{79.32} \\
    \toprule
    \end{tabular}%
    }
  \label{tab:table5}%
  \vspace{-0.1cm}
\end{table}%

\textit{2) Comparison with Alternative Methods.} Table \ref{tab:table5} compares the anomaly detection performance of HiAD with several alternative methods. Specifically, EfficientAD \cite{batzner2024efficientad} and VarAD \cite{cao2025varad} adopt lightweight architectures to enable high-resolution anomaly detection. For a fair comparison, we configure these methods to operate at their maximum feasible resolutions under the same GPU memory constraints. The results indicate that such lightweight designs often struggle to effectively model the complex distributions inherent in high-resolution images, leading to suboptimal performance. In addition, HiAD outperforms the Tiling method \cite{rolih2024divide} while using fewer detectors and consistently maintains a performance advantage across different integration settings. These results highlight the effectiveness of HiAD in handling ultra-high-resolution images and addressing challenging anomaly detection scenarios.

\begin{table}[t]
  \centering
  \renewcommand\arraystretch{1.2}
  \caption{Computational efficiency comparison of HiAD integrated with different anomaly detection methods. Peak tensor memory is summed across all GPUs.}
    \resizebox{\linewidth}{!}{
    \Large
    \begin{tabular}{c|c|cc|c|c|c}
    \bottomrule
    \multicolumn{1}{c|}{\multirow{2}[1]{*}{Resolution}} & \multirow{2}[1]{*}{Method} & \multicolumn{2}{c|}{Inference Time (s) $\downarrow$} & \multirow{2}[1]{*}{\makecell{Training Time \\ (GPU h) $\downarrow$}} & \multirow{2}[1]{*}{\makecell{Peak Tensor \\ Memory (GB) $\downarrow$}} & \multirow{2}[1]{*}{\makecell{\# Params (M) $\downarrow$}}\\
\cline{3-4}          & \multicolumn{1}{c|}{} & \multicolumn{1}{c}{1 GPU} & \multicolumn{1}{c|}{4 GPUs} & \multicolumn{1}{c|}{} & \multicolumn{1}{c|}{} & \multicolumn{1}{c}{} \\
    \hline
    \multicolumn{1}{c|}{\multirow{8}[2]{*}{\makecell{$\text{2048} \times \text{2048}$ \\ $(M=4)$}}} & DeSTSeg \cite{zhang2023destseg} & \textbf{0.84}  & \textbf{0.49}  & \multicolumn{1}{c|}{7.5} & \multicolumn{1}{c|}{\textbf{0.85}} &  \multicolumn{1}{c}{\textbf{163.0}}  \\
          & Dinomaly \cite{guo2025dinomaly} & 0.97  & 0.64  & \multicolumn{1}{c|}{4.0} & \multicolumn{1}{c|}{2.93} &  \multicolumn{1}{c}{393.1} \\
          & FastFlow \cite{yu2021fastflow} & 0.98  & 0.54  & \multicolumn{1}{c|}{\textbf{3.5}} & \multicolumn{1}{c|}{1.80} &  \multicolumn{1}{c}{358.0}\\
          & RD++  \cite{tien2023revisiting} & 1.19  & 0.57  & \multicolumn{1}{c|}{4.5} & \multicolumn{1}{c|}{3.21} &  \multicolumn{1}{c}{519.9}\\
          & RealNet \cite{zhang2024realnet} & 1.20   & 0.60   & \multicolumn{1}{c|}{8.0} & \multicolumn{1}{c|}{5.49} &  \multicolumn{1}{c}{896.6}\\
          & ViTAD \cite{zhang2023exploring} & 2.09  & 0.78  & \multicolumn{1}{c|}{4.0} & \multicolumn{1}{c|}{2.92} &  \multicolumn{1}{c}{417.8}\\
          & PatchCore \cite{roth2022towards} & 2.51  & 0.91  & --     & -- & --\\
          & PaDiM \cite{defard2021padim} & 11.52 & 3.21  & --     & -- & --\\
    \hline
    \multicolumn{1}{c|}{\multirow{8}[2]{*}{\makecell{$\text{4096} \times \text{4096}$ \\ $(M=8)$}}} & DeSTSeg \cite{zhang2023destseg} & \textbf{3.23}  & \textbf{1.79}  & \multicolumn{1}{c|}{16.0} & \multicolumn{1}{c|}{\textbf{1.95}} & \multicolumn{1}{c}{\textbf{291.1}}\\
          & Dinomaly \cite{guo2025dinomaly} & 3.38  & 1.81  & \multicolumn{1}{c|}{9.0} & \multicolumn{1}{c|}{5.69} & \multicolumn{1}{c}{638.8}\\
          & FastFlow \cite{yu2021fastflow} & 3.66  & 1.90   & \multicolumn{1}{c|}{\textbf{7.5}} & \multicolumn{1}{c|}{3.37} & \multicolumn{1}{c}{624.6}\\
          & RD++  \cite{tien2023revisiting} & 4.34  & 2.12  & \multicolumn{1}{c|}{9.0} & \multicolumn{1}{c|}{5.90}  & \multicolumn{1}{c}{915.8}  \\
          & RealNet \cite{zhang2024realnet} & 4.77  & 2.21  & \multicolumn{1}{c|}{15.0} & \multicolumn{1}{c|}{9.84}  &  \multicolumn{1}{c}{1,593.9}\\
          & ViTAD \cite{zhang2023exploring} & 5.14  & 2.55  & \multicolumn{1}{c|}{\textbf{7.5}} & \multicolumn{1}{c|}{5.39} & \multicolumn{1}{c}{682.5}\\
          & PatchCore \cite{roth2022towards}& 9.96  & 3.54  & --     & -- & --\\
          & PaDiM \cite{defard2021padim} & 47.18 & 12.61 & --    & -- & --\\
    \toprule 
    \end{tabular}%
    }
    \vspace{-0.1cm}
  \label{tab:table6}%
\end{table}%

\textit{3) Computational Efficiency Evaluation.} Table \ref{tab:table6} compares the computational efficiency of HiAD combined with different anomaly detection methods. All experiments are implemented in PyTorch and conducted on NVIDIA RTX 4090 GPUs with 24GB of VRAM and 64GB of system memory, without any specialized acceleration techniques. All inference measurements are conducted with a batch size of 1. HiAD supports assigning image patches to different GPUs for parallel inference. Under the setting of 4 GPUs, HiAD can process a 2K image in 0.49 seconds and a 4K image in 1.79 seconds with DeSTSeg \cite{zhang2023destseg}. By employing efficient detector assignment strategies, HiAD allows flexible control over the parameter count to accommodate different detection requirements. With a detector pool size of 4, its overall parameter count after integration with most methods remains roughly comparable to that of widely adopted high-capacity foundation models, such as DINO-ViT-L/16 (300M) \cite{simeoni2025dinov3} and CLIP-ViT-L/14 (428M) \cite{radford2021learning}, demonstrating its practical feasibility.

\textit{4) Multi-class High-Resolution Anomaly Detection Results.} In the multi-class setting \cite{you2022unified}, anomaly detection is performed over multiple target categories without using class labels during either training or inference. For high-resolution images, this task becomes more challenging due to the vast variety of image patches. Under the multi-class setting, ASS selects the RA assignment strategy, which groups visually similar image patches from different classes into the same detector, thereby simplifying the feature distribution handled by each detector and improving anomaly detection performance. Table \ref{tab:table7} reports the multi-class high-resolution results of HiAD, evaluated on the MVTec-2K, VisA-2K, and RealIAD-2K datasets, which contain 7, 5, and 3 categories, respectively. HiAD achieves promising anomaly detection results across all datasets, with only slight performance degradation compared to the one-class-one-model setting. These results highlight HiAD's broad applicability to diverse anomaly detection scenarios, underlining its suitability for multi-class, high-resolution tasks.

\begin{table*}[t!] 
    \centering
    \caption{Multi-class anomaly detection results of HiAD on the 2K datasets at a resolution of $2048 \times 2048$, with $M=4$.}
    \renewcommand{\arraystretch}{1.1}
    \resizebox{\linewidth}{!}{
    \large
    \begin{tabular}{c|c|ccccc|ccccc|ccccc}
    \bottomrule
    \multicolumn{1}{c|}{\multirow{2}[1]{*}{Dataset}} & \multirow{2}[1]{*}{Category} & \multicolumn{5}{c|}{Dinomaly \cite{guo2025dinomaly}} & \multicolumn{5}{c|}{ViTAD \cite{zhang2023exploring}}  & \multicolumn{5}{c}{DeSTSeg \cite{zhang2023destseg}} \\
    \cline{3-17} 
    & \multicolumn{1}{c|}{} & \textbf{I-AUC} & \textbf{P-AUC} & \textbf{P-AP} & \textbf{P-F1} & \multicolumn{1}{c|}{\textbf{PRO}} & \textbf{I-AUC} & \textbf{P-AUC} & \textbf{P-AP} & \textbf{P-F1} & \multicolumn{1}{c|}{\textbf{PRO}} & \textbf{I-AUC} & \textbf{P-AUC} & \textbf{P-AP} & \textbf{P-F1} & \textbf{PRO} \\
    \hline
    \multicolumn{1}{c|}{\multirow{8}[1]{*}{MVTec-2K}} 
          & Bottle & \textbf{98.55} & \textbf{99.78} & \textbf{88.36}  & 80.28 & \textbf{97.88} & 97.78 & 99.65 & 79.90  & 77.33 & 97.74 & 95.12 & 97.99 & 88.03 & \textbf{83.10} & 96.43 \\
          & Capsule & \textbf{95.22} & \textbf{99.27} & \textbf{54.97} & 54.60 & 98.30 &  91.28 & 98.95 & 37.75 & 44.52 & \textbf{98.69} & 91.99 & 96.38 & 54.88 & \textbf{55.80} & 93.47 \\
          & Grid  & \textbf{99.85} & \textbf{99.85} & 74.30 & 67.94 & \textbf{99.26} & 96.91 & 99.10  & 31.64 & 36.73 & 95.84 & 98.70 & 99.71 & \textbf{74.89} & \textbf{69.04} & 98.63 \\
          & Hazelnut & 92.94 & \textbf{99.80} & \textbf{82.63} & \textbf{79.19} & 96.81 & \textbf{95.88} & 99.62 & 69.62 & 71.03 & \textbf{97.55} & 91.20  & 99.49 & 79.73 & 73.30 & 92.82 \\
          & Screw & \textbf{95.24} & \textbf{99.90} & 55.62 & \textbf{66.42} & \textbf{99.58} &  91.30  & 99.45 & 34.91 & 40.69 & 99.06 & 94.23 & 98.27 & \textbf{58.78} & 57.92 & 97.03 \\
          & Transistor & 96.80 & \textbf{93.65} & 52.44 & 51.52 & \textbf{92.51} &  95.58 & 93.43 & \textbf{53.05} & \textbf{51.68} & 86.99 & \textbf{97.52} & 77.63 & 41.30  & 45.41 & 82.80 \\
          & Wood  & 93.00 & \textbf{99.46} & 83.24 & 75.55 & 95.45 &  98.59 & 98.79 & 69.40  & 64.98 & 91.61 & \textbf{99.18} & 98.99 & \textbf{84.61} & \textbf{77.43} & \textbf{95.50} \\
          & \cellcolor{gray!15}\textbf{AVG} & \cellcolor{gray!15}\textbf{95.94} & \cellcolor{gray!15}\textbf{98.82} & \cellcolor{gray!15}\textbf{70.22} & \cellcolor{gray!15}\textbf{67.93} & \cellcolor{gray!15}\textbf{97.11} &  \cellcolor{gray!15}95.33 & \cellcolor{gray!15}98.43 & \cellcolor{gray!15}53.75 & \cellcolor{gray!15}55.28 & \cellcolor{gray!15}95.36 & \cellcolor{gray!15}95.42 & \cellcolor{gray!15}95.49 & \cellcolor{gray!15}68.89 & \cellcolor{gray!15}66.00 & \cellcolor{gray!15}93.81 \\
    \hline
    \multicolumn{1}{c|}{\multirow{6}[1]{*}{VisA-2K}} 
          & Capsules & \textbf{96.22} & \textbf{99.58} & 42.18 & 50.13 & \textbf{97.96} &  91.85 & 98.71 & 31.01 & 41.46 & 97.92 & 82.54 & 98.11 & \textbf{69.23} & \textbf{70.27} & 94.16 \\
          & Fryum & 97.51 & 99.54 & 45.83 & 46.77 & \textbf{98.69} & \textbf{98.54} & \textbf{99.59} & 42.45 & 47.45 & 97.65 & 93.44 & 99.44 & \textbf{64.25} & \textbf{62.13} & 90.04 \\
          & Macaroni & \textbf{95.67} & \textbf{99.91} & 36.78 & 41.22	& \textbf{99.28} &  92.98 & 99.63 & \textbf{41.06} & \textbf{47.68} & 96.43 & 89.74 & 99.41 & 39.88 & 44.61 & 90.56 \\
          & PCB   & \textbf{98.94} & \textbf{99.58} & \textbf{88.45} & \textbf{82.99} & \textbf{97.30} &  93.59 & 99.53 & 76.94 & 71.58 & 96.78 & 98.25 & 99.50  & 82.73 & 75.29 & 96.35 \\
          & Pipe Fryum & 94.31 & 99.68 & 66.52 & 62.02 & 97.81 &  \textbf{97.74} & \textbf{99.75} & \textbf{68.41} & \textbf{67.67} & \textbf{98.34} & 96.08 & 97.69 & 65.60  & 61.57 & 83.80 \\
          & \cellcolor{gray!15}\textbf{AVG} & \cellcolor{gray!15}\textbf{96.53} & \cellcolor{gray!15}\textbf{99.66} & \cellcolor{gray!15}55.95 & \cellcolor{gray!15}56.63 & \cellcolor{gray!15}\textbf{98.21} & \cellcolor{gray!15}94.94 & \cellcolor{gray!15}99.44 & \cellcolor{gray!15}51.97 & \cellcolor{gray!15}55.17 & \cellcolor{gray!15}97.42 & \cellcolor{gray!15}92.01 & \cellcolor{gray!15}98.83 & \cellcolor{gray!15}\textbf{64.34} & \cellcolor{gray!15}\textbf{62.77} & \cellcolor{gray!15}90.98 \\
    \hline
    \multicolumn{1}{c|}{\multirow{4}[1]{*}{RealIAD-2K}} 
          & Bottle Cap & \textbf{95.79} & \textbf{99.96} & 36.96 & 36.06 & 99.24 &  95.52 & 99.95 & 35.78 & 42.68 & \textbf{99.71} & 94.12 & 99.84 & \textbf{71.84} & \textbf{66.87} & 94.87 \\
          & Mint  & 85.72 & \textbf{99.69} & 26.59 & 34.14 & 97.60 &  80.17 & 99.58 & 26.77 & 37.32 & \textbf{97.99} & \textbf{87.52} & 99.65 & \textbf{46.03} & \textbf{46.34} & 96.24 \\
          & USB Adaptor & 95.96 & \textbf{99.94} & 37.20 & 39.72 & \textbf{99.66} &  \textbf{96.19} & 99.60  & 22.00    & 29.37 & 95.82 & 93.22 & 99.84 & \textbf{37.39} & \textbf{43.04} & 91.18 \\
          & \cellcolor{gray!15}\textbf{AVG} & \cellcolor{gray!15}\textbf{92.49} & \cellcolor{gray!15}\textbf{99.86} & \cellcolor{gray!15}33.58 & \cellcolor{gray!15}36.64 & \cellcolor{gray!15}\textbf{98.83} &  \cellcolor{gray!15}90.62 & \cellcolor{gray!15}99.71 & \cellcolor{gray!15}28.19 & \cellcolor{gray!15}36.46 & \cellcolor{gray!15}97.84 & \cellcolor{gray!15}91.62 & \cellcolor{gray!15}99.78 & \cellcolor{gray!15}\textbf{51.75} & \cellcolor{gray!15}\textbf{52.08} & \cellcolor{gray!15}94.10 \\
    \toprule
    \end{tabular}%
    }
  \label{tab:table7}%
   \vspace{-0.2cm}
\end{table*}%

\begin{table}[t]
  \centering
  \caption{Zero-shot anomaly detection results of HiAD integrated with AnomalyCLIP on the MVTec-2K dataset.}
  \renewcommand{\arraystretch}{1.6}
  \resizebox{0.97\linewidth}{!}{
    \begin{tabular}{c|ccccc}
    \bottomrule
    Method & \multicolumn{1}{c}{\textbf{I-AUC}} & \multicolumn{1}{c}{\textbf{P-AUC}} & \multicolumn{1}{c}{\textbf{P-AP}} & \multicolumn{1}{c}{\textbf{P-F1}} & \multicolumn{1}{c}{\textbf{PRO}} \\
    \hline
    AnomalyCLIP \cite{zhou2024anomalyclip} & \multirow{1}[1]{*}{83.8} & \multirow{1}[1]{*}{91.8} & \multirow{1}[1]{*}{28.0} & \multirow{1}[1]{*}{34.4} & \multirow{1}[1]{*}{87.5} \\
    \cellcolor{gray!15}HiAD{\fontsize{6}{3}\selectfont (AnomalyCLIP)}  & \multicolumn{1}{c}{\multirow{1}[1]{*}{\cellcolor{gray!15}87.9$\mathtiny{(4.1\uparrow)}$}} & \multicolumn{1}{c}{\multirow{1}[1]{*}{\cellcolor{gray!15}93.2$\mathtiny{(1.4\uparrow)}$}} & \multicolumn{1}{c}{\multirow{1}[1]{*}{\cellcolor{gray!15}33.1$\mathtiny{(5.1\uparrow)}$}} & \multicolumn{1}{c}{\multirow{1}[1]{*}{\cellcolor{gray!15}40.3$\mathtiny{(5.9\uparrow)}$}} & \multicolumn{1}{c}{\multirow{1}[1]{*}{\cellcolor{gray!15}91.5$\mathtiny{(4.0\uparrow)}$}} \\
   \toprule
    \end{tabular}%
    }
  \label{tab:table8}%
 \vspace{-0.2cm}
\end{table}%

\textit{5) Zero-shot Anomaly Detection Results.} We further evaluate the zero-shot performance of HiAD, with the results reported in Table \ref{tab:table8}. Specifically, we integrate HiAD with AnomalyCLIP \cite{zhou2024anomalyclip}, using CLIP-ViT-L/14 \cite{radford2021learning} as the backbone, and set the patch size to 518 $\times$ 518. HiAD performs anomaly detection at an input resolution of 2048 $\times$ 2048, while the low-resolution AnomalyCLIP operating at 518$\times$518 serves as the baseline. Both methods are trained on the VisA-2K dataset and evaluated on MVTec-2K. The experimental results demonstrate that HiAD still delivers substantial performance gains under the zero-shot setting, improving I-AUC and PRO by 4.1\% and 4.0\%, respectively. Nevertheless, owing to the inherent challenges of high-resolution anomaly detection, this cross-domain setting still struggles to identify certain challenging subtle anomalies, leaving a performance gap relative to the full-shot setting. In future work, we will further investigate strategies to improve the zero-shot anomaly detection performance of HiAD.

\subsection{Ablation Study}

\textit{1) Ablation Study on Low-Resolution Branch and Multi-Resolution Feature Fusion.} Table \ref{tab:table9} validates the effectiveness of the low-resolution branch and multi-resolution feature fusion in HiAD. When both are removed, HiAD degenerates into a tiling-based scheme with detector assignment. Due to its inability to effectively detect large-scale anomalies and its susceptibility to low-level texture variations, it yields suboptimal performance. The incorporation of the low-resolution branch generally improves anomaly detection performance. However, it is worth noting that for the RealIAD-2K dataset, which contains few large-scale anomalous regions, incorporating the low-resolution branch may result in degraded anomaly localization performance. Apart from the aforementioned case, we recommend using the low-resolution branch to enhance the overall anomaly detection performance of HiAD. By replacing the original pre-trained features with multi-resolution fusion features, HiAD consistently improves performance across all benchmarks. This demonstrates that MRFF effectively mitigates the adverse impact of ultrafine textures while ensuring accurate identification of subtle anomalies. 

\begin{table*}[t]
  \centering
  \renewcommand\arraystretch{1.1}
  \caption{Ablation study results on the low-resolution branch and multi-resolution feature fusion at a resolution of $2048 \times 2048$. }
  \resizebox{0.95\linewidth}{!}{
  \scriptsize
  \begin{tabular}{c|c|c|ccccc|ccccc}
     \bottomrule
    \multicolumn{1}{c|}{\multirow{2}[1]{*}{Method}} & \multirow{2}[1]{*}{\makecell{ Low-Resolution\\Branch}} & \multirow{2}[1]{*}{\makecell{Multi-Resolution \\ Feature Fusion}} & \multicolumn{5}{c|}{MVTec-2K} & \multicolumn{5}{c}{RealIAD-2K} \\
\cline{4-13}          & \multicolumn{1}{c|}{} & \multicolumn{1}{c|}{} & \multicolumn{1}{c}{\textbf{I-AUC}} & \multicolumn{1}{c}{\textbf{P-AUC}} & \multicolumn{1}{c}{\textbf{P-AP}} & \multicolumn{1}{c}{\textbf{P-F1}} & \multicolumn{1}{c|}{\textbf{PRO}} & \multicolumn{1}{c}{\textbf{I-AUC}} & \multicolumn{1}{c}{\textbf{P-AUC}} & \multicolumn{1}{c}{\textbf{P-AP}} & \multicolumn{1}{c}{\textbf{P-F1}} & \multicolumn{1}{c}{\textbf{PRO}} \\
    \hline
    \multicolumn{1}{c|}{\multirow{4}[1]{*}{Dinomaly \cite{guo2025dinomaly}}} & \xmark     & \xmark     & 94.56 & 97.79 & 66.01  & 64.78 & 95.66 & 92.78 & 99.77  & 40.13 & 44.57 & 98.83\\
          & \cmark     & \xmark     & 94.98 & 98.69 & 70.02 & 67.54 & 96.24 & 93.74 & 99.72 & 36.21 &  38.35 & 98.10 \\
          & \xmark    & \cmark     & 96.41 & 98.59 & 71.52 & 68.35 & 97.52 & 93.33 & \textbf{99.89} & \textbf{40.54} & \textbf{45.14} &  \textbf{99.31} \\
          & \cellcolor{gray!15}\cmark     & \cellcolor{gray!15}\cmark    & \cellcolor{gray!15}\textbf{97.17} & \cellcolor{gray!15}\textbf{99.07} & \cellcolor{gray!15}\textbf{72.98} & \cellcolor{gray!15}\textbf{69.10} & \cellcolor{gray!15}\textbf{97.58}  & \cellcolor{gray!15}\textbf{94.26} & \cellcolor{gray!15}99.86 & \cellcolor{gray!15}39.62 & \cellcolor{gray!15}40.88 &  \cellcolor{gray!15}99.23\\    
    \hline
    \multicolumn{1}{c|}{\multirow{4}[1]{*}{ViTAD \cite{zhang2023exploring}}} & \xmark     & \xmark     & 95.07 & 96.78  & 60.96 & 61.04 & 96.11 & 93.05 & 99.73 & \textbf{41.02} & 44.76 & 98.12 \\
          & \cmark     & \xmark     & 95.45 & 98.74 & 63.91 & 62.50 & 97.17 & 93.59 & 99.74 & 34.64 & 41.77 & 97.96 \\
          & \xmark     & \cmark     & \textbf{96.82} & 98.24 & \textbf{65.74} & \textbf{64.13} & \textbf{97.84} & 93.85 & 99.78 & \textbf{41.02} & \textbf{45.31} & \textbf{98.14} \\
          & \cellcolor{gray!15}\cmark     & \cellcolor{gray!15}\cmark     & \cellcolor{gray!15}96.45 & \cellcolor{gray!15}\textbf{99.06} & \cellcolor{gray!15}64.78 & \cellcolor{gray!15}64.09 & \cellcolor{gray!15}97.30 & \cellcolor{gray!15}\textbf{94.47} & \cellcolor{gray!15}\textbf{99.79} & \cellcolor{gray!15}36.83 & \cellcolor{gray!15}41.88 & \cellcolor{gray!15}98.07 \\
    \hline
  \multicolumn{1}{c|}{\multirow{4}[1]{*}{PatchCore \cite{roth2022towards}}} & \xmark     & \xmark     & 97.75 & 95.55 & 52.53 & 52.55 & 95.88 & 97.27 & 99.78 & 44.70  & 47.94 & 98.40 \\
          & \cmark     & \xmark     & 98.37 & 97.69 & 64.74 & 63.05 & 97.26 & 97.50  & 99.85 & 43.03 & 46.95 & 99.03 \\
          & \xmark     & \cmark    & 98.25 & 96.54 & 57.95 & 57.53 & 96.74 & 98.20  & 99.86 & \textbf{45.26} & \textbf{47.98} & 98.72 \\
          & \cellcolor{gray!15}\cmark     & \cellcolor{gray!15}\cmark    & \cellcolor{gray!15}\textbf{98.44} & \cellcolor{gray!15}\textbf{97.79} & \cellcolor{gray!15}\textbf{65.84} & \cellcolor{gray!15}\textbf{64.07} & \cellcolor{gray!15}\textbf{97.39} & \cellcolor{gray!15}\textbf{98.37} & \cellcolor{gray!15}\textbf{99.89} & \cellcolor{gray!15}45.16 & \cellcolor{gray!15}47.34 & \cellcolor{gray!15}\textbf{99.10} \\
  \hline
    \multicolumn{1}{c|}{\multirow{4}[1]{*}{PaDiM \cite{defard2021padim}}} & \xmark     & \xmark     & 77.75 & 92.22 & 26.60  & 30.91 & 90.24 & 83.67 & 99.50  & 13.52 & 23.87 & 99.07 \\
          & \cmark     & \xmark     & 81.03 & 96.74 & 37.82 & 41.79 & 94.53 & 86.66 & 99.58 & 18.99 & 27.64 & 98.06 \\
          & \xmark     & \cmark     & 82.76 & 95.17 & 35.25 & 38.72 & 93.25 & 89.56 & \textbf{99.71} & \textbf{24.60} & \textbf{32.16} & \textbf{99.18} \\
          & \cellcolor{gray!15}\cmark     & \cellcolor{gray!15}\cmark    & \cellcolor{gray!15}\textbf{86.22} & \cellcolor{gray!15}\textbf{98.13} & \cellcolor{gray!15}\textbf{44.43} & \cellcolor{gray!15}\textbf{48.50} & \cellcolor{gray!15}\textbf{95.95} & \cellcolor{gray!15}\textbf{89.67} & \cellcolor{gray!15}99.60  & \cellcolor{gray!15}20.74 & \cellcolor{gray!15}28.22 & \cellcolor{gray!15}97.98 \\
    \toprule
    \end{tabular}
    }
  \label{tab:table9}
  \vspace{-0.1cm}
\end{table*}%

\begin{table*}[t]
  \centering
  \renewcommand\arraystretch{1.4}
  \caption{Comparison of different detector assignment strategies with Dinomaly as the detector. The detector pool size is indicated in parentheses, with smaller numbers indicating lower computational cost.}
    \resizebox{0.95\linewidth}{!}{
    \Large
    \begin{tabular}{c|ccccc|ccccc|ccccc|ccccc}
    \bottomrule
    \multirow{2}[1]{*}{\makecell{Assignment \\ Strategy}} & \multicolumn{5}{c|}{MVTec-2K} & \multicolumn{5}{c|}{VisA-2K} & \multicolumn{5}{c|}{RealIAD-2K} & \multicolumn{5}{c}{MVTec 2} \\
\cline{2-21}    \multicolumn{1}{c|}{} & \multicolumn{1}{c}{\textbf{I-AUC}} & \multicolumn{1}{c}{\textbf{P-AUC}} & \multicolumn{1}{c}{\textbf{P-AP}} & \multicolumn{1}{c}{\textbf{P-F1}} & \multicolumn{1}{c|}{\textbf{PRO}} & \multicolumn{1}{c}{\textbf{I-AUC}} & \multicolumn{1}{c}{\textbf{P-AUC}} & \multicolumn{1}{c}{\textbf{P-AP}} & \multicolumn{1}{c}{\textbf{P-F1}} & \multicolumn{1}{c|}{\textbf{PRO}} & \multicolumn{1}{c}{\textbf{I-AUC}} & \multicolumn{1}{c}{\textbf{P-AUC}} & \multicolumn{1}{c}{\textbf{P-AP}} & \multicolumn{1}{c}{\textbf{P-F1}} & \multicolumn{1}{c|}{\textbf{PRO}} & \multicolumn{1}{c}{\textbf{I-AUC}} & \multicolumn{1}{c}{\textbf{P-AUC}} & \multicolumn{1}{c}{\textbf{P-AP}} & \multicolumn{1}{c}{\textbf{P-F1}} & \multicolumn{1}{c}{\textbf{PRO}} \\
    \hline
    A2O${\mathsmallish{(1)}}$   & 95.29 & 99.06 & 70.50  & 67.83 & 96.47 & 95.12 & 99.63 & 52.60  & 54.00    & 97.23 & 90.33 & 99.84 & 22.87 & 33.21 & 97.07 & 76.17 & 93.75 & 31.42 & 35.97 & 76.70 \\
    NA${\mathsmallish{(4)}}$    & 96.08 & \textbf{99.14} & 71.00   & 67.67 & 97.17 & 96.19 & \underline{99.71} & 54.24 & 56.87 & 98.39 & \underline{93.35} & \underline{99.88} & 25.27 & 36.35 & 98.96 & 76.27 & 93.95 & 32.02 & 36.52 & 77.04 \\
    SCA${\mathsmallish{(4)}}$   & 97.04 & 99.07 & 72.25 & 68.62 & \underline{97.44} & 98.23 & \textbf{99.74} & 56.10  & 57.33 & \underline{98.84} & \textbf{94.26} & 99.86 & \textbf{39.62} & \textbf{40.88} & \textbf{99.23}  & 76.68 & 94.61 & 33.07 & 37.81 & 77.00 \\
    RA${\mathsmallish{(4)}}$    & 96.78 & 99.07 & 72.52 & 68.56 & 97.05 & 97.17 & \textbf{99.74} & 55.64 & 57.24 & 98.69 & 92.50  & \textbf{99.89} & 27.55 & 34.37 & \underline{99.10}  & 78.00    & 94.65 & 33.09 & 37.81 & 78.60 \\

    O2O${\mathsmallish{(16)}}$   &    96.63   &   \underline{99.11}    &  72.30	   &  68.13  &   97.26   &    96.48   &    \textbf{99.74}   &   55.15   &   56.35  &  98.81   &    93.05	&    99.86	&  \underline{30.23} &   \underline{38.32}	&  99.05    &   77.24    &   94.11    &    32.93   &  37.37   & 77.71 \\
    \hline
    FAF${\mathsmallish{(4)}}$   & \underline{97.16} & 99.07 & \underline{72.91} & \underline{69.06} & 97.43 & \underline{98.41} & \textbf{99.74} & \textbf{56.80}  & \underline{57.59} & \textbf{98.86} & \textbf{94.26} & 99.86 & \textbf{39.62} & \textbf{40.88} & \textbf{99.23}  & \underline{78.90}  & \underline{94.90}  & \underline{33.94} & \underline{38.92} & \underline{78.76} \\
    SPC${\mathsmallish{(4)}}$   & 96.24 & \textbf{99.14} & 71.64 & 68.08 & 97.17 & 96.19 & \underline{99.71} & 54.24 & 56.87 & 98.39 & \underline{93.35} & \underline{99.88} & 25.27 & 36.35 & 98.96 & 76.27 & 93.95 & 32.02 & 36.52 & 77.04 \\
    \cellcolor{gray!15}ASS${\mathsmallish{(4)}}$   & \cellcolor{gray!15}\textbf{97.17} & \cellcolor{gray!15}99.07 & \cellcolor{gray!15}\textbf{72.98} & \cellcolor{gray!15}\textbf{69.10}  & \cellcolor{gray!15}\textbf{97.58} & \cellcolor{gray!15}\textbf{98.57} & \cellcolor{gray!15}\textbf{99.74} & \cellcolor{gray!15}\underline{56.72} & \cellcolor{gray!15}\textbf{57.89} & \cellcolor{gray!15}\textbf{98.86} & \cellcolor{gray!15}\textbf{94.26} & \cellcolor{gray!15}99.86 & \cellcolor{gray!15}\textbf{39.62} & \cellcolor{gray!15}\textbf{40.88} & \cellcolor{gray!15}\textbf{99.23} & \cellcolor{gray!15}\textbf{80.10}  & \cellcolor{gray!15}\textbf{94.92} & \cellcolor{gray!15}\textbf{34.52} & \cellcolor{gray!15}\textbf{39.41} & \cellcolor{gray!15}\textbf{79.32} \\
    \toprule
    \end{tabular}%
    }
  \label{tab:table10}%
  \vspace{-0.1cm}
\end{table*}%

\begin{table}[t]
  \centering
  \renewcommand\arraystretch{2.0}
  \caption{Selection results of ASS across different benchmarks using DINOv3 as the backbone, with $M=4$.}
   \resizebox{0.95\linewidth}{!}{
  \begin{tabular}{l | l}
  \bottomrule
    \makecell[l]{Selected \\ Strategy} & Category \\
    \hline
    NA  &  \makecell[l]{MVTec-2K: \{Wood\} \\ MVTec 2: \{Rice\}} \\
    \hline
    SCA & \makecell[l]{MVTec-2K: \{Bottle, Capsule, Hazelnut, Transistor\} \\ 
                    VisA-2K: \{Fryum, Macaroni, PCB, Pipe Fryum\} \\ 
                    RealIAD-2K: \{Bottle Cap, Mint, USB Adaptor\} \\ 
                    MVTec 2: \{Sheet Metal, Vial\}} \\
    \hline
    RA  & \makecell[l]{MVTec-2K: \{Screw, Grid\} \\ 
                    VisA-2K: \{Capsules\} \\ 
                    MVTec 2: \{Can, Fabric, Fruit Jelly, Wallplugs, Walnuts\}} \\
    \toprule
  \end{tabular}
  }
  \label{tab:table11}%
\end{table}

\begin{figure}[t]
  \centering
   \includegraphics[width=\linewidth]{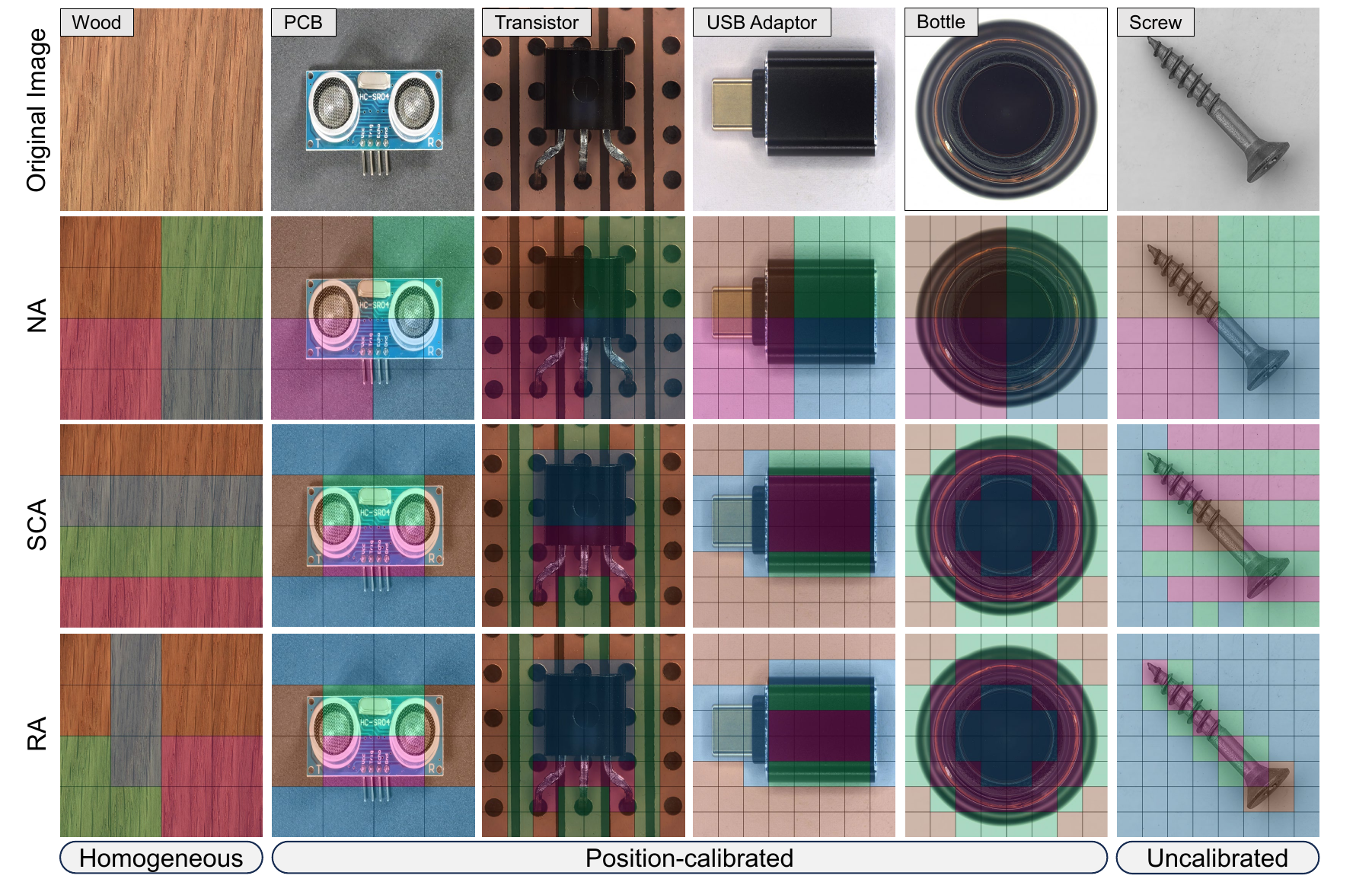}
   \caption{Qualitative comparison of NA, SCA, and RA. The detector pool size is set to 4, and image patches assigned to the same detector are marked with the same color.}
   \label{fig:fig8}
   \vspace{-0.2cm}
\end{figure}

\textit{2) Ablation Study on Detector Assignment.} Table \ref{tab:table10} compares the anomaly detection performance of different detector assignment strategies, using Dinomaly \cite{guo2025dinomaly} as the detector. The proposed ASS yields performance gains across multiple benchmarks, demonstrating its robustness and adaptability under diverse data distributions. To investigate the selection process, we examine each metric independently. When FAF is used alone, HiAD shows a propensity for $\{\text{SCA, RA} \}$, which may yield fragmented anomaly maps and compromise detection performance. Conversely, relying exclusively on SPC biases the selection toward NA; however, this strategy disregards the semantic information of patches, thereby limiting the detector's capacity to accommodate structural diversity. By contrast, ASS effectively integrates these two metrics to achieve a favorable trade-off. Table \ref{tab:table11} reports the selection results of ASS across different benchmarks. For homogeneous high-resolution images, such as the Wood category in the MVTec-2K dataset, the FAF scores across different strategies are nearly identical. In this case, ASS selects NA to preserve the spatial continuity. For images with well-defined spatial structures and positional calibration, such as the PCB category, SCA emerges as the optimal choice due to its robust spatial clustering quality and assignment consistency. Conversely, for products without positional calibration, such as the Screw category, patches at the same spatial position across different images exhibit substantial appearance variations, making SCA ineffective for clustering, as shown in Figure \ref{fig:fig8}. In contrast, RA performs nearest-neighbor retrieval for each image patch, ensuring that visually similar patches are correctly grouped, and thus becomes the primary choice of ASS. Table \ref{tab:table12} presents the anomaly detection performance of different assignment strategies on the three representative categories discussed above. The results demonstrate that ASS can effectively identify optimal assignment schemes for varying input data, improving detection performance while controlling computational costs, thereby underscoring the broad applicability of the HiAD framework. 

\begin{table}[t]
  \centering
  \renewcommand\arraystretch{1.15}
  \caption{Comparison of detector assignment strategies across different categories. $^\dagger$ denotes the strategy selected by ASS under the setting of $M=4$.}
   \resizebox{0.95\linewidth}{!}{
    \begin{tabular}{c|c|ccccc}
    \bottomrule
    \multicolumn{1}{c|}{\makecell{Anomaly Detection \\ Method}} & \makecell{Detector Assignment \\ Strategy} & \multicolumn{1}{c}{\textbf{I-AUC}} & \multicolumn{1}{c}{\textbf{P-AUC}} & \multicolumn{1}{c}{\textbf{P-AP}} & \multicolumn{1}{c}{\textbf{P-F1}} & \multicolumn{1}{c}{\textbf{PRO}} \\
    \hline
    \multicolumn{7}{c}{Wood (MVTec-2K)} \\
    \hline
    \multicolumn{1}{c|}{\multirow{5}[1]{*}{Dinomaly \cite{guo2025dinomaly}}} 
          & A2O${\mathsmallish{(1)}}$ & 93.89  & 99.40 & 82.51  & 75.57  & 96.09 \\
          & \cellcolor{gray!15}NA${\mathsmallish{(4)}}^\dagger$ & \cellcolor{gray!15}\textbf{94.27} & \cellcolor{gray!15}\textbf{99.47} & \cellcolor{gray!15}\textbf{83.17} & \cellcolor{gray!15}\textbf{76.13} & \cellcolor{gray!15}\textbf{97.07} \\
          & SCA${\mathsmallish{(4)}}$ & 94.16	 & 99.46 &  82.66 & 75.75  & 96.23 \\
          & RA${\mathsmallish{(4)}}$ & 94.18 & \textbf{99.47}&  82.67 & 75.86 & 96.00 \\
          & O2O${\mathsmallish{(16)}}$ & 93.08 & 99.34 &  82.55 &  74.90 & 95.87 \\
    \hline
    \multicolumn{7}{c}{PCB (VisA-2K)} \\
    \hline
    \multicolumn{1}{c|}{\multirow{5}[1]{*}{Dinomaly \cite{guo2025dinomaly}}} 
          & A2O${\mathsmallish{(1)}}$ & 96.53 & 99.53 & 83.49 & 81.57 & 96.74 \\
          & NA${\mathsmallish{(4)}}$ & 97.38 & 99.58 & 83.27 & 81.76 & 97.55 \\
          & \cellcolor{gray!15}SCA${\mathsmallish{(4)}}^\dagger$ & \cellcolor{gray!15}\textbf{98.98} & \cellcolor{gray!15}\textbf{99.68} & \cellcolor{gray!15}\textbf{86.00} & \cellcolor{gray!15}\textbf{83.44} & \cellcolor{gray!15}\textbf{98.25} \\
          & RA${\mathsmallish{(4)}}$ & 98.67 & 99.55 & 85.27 & 83.12	& 97.51 \\
          & O2O${\mathsmallish{(16)}}$ & 97.42 & \textbf{99.68} & 85.32 & 80.43 & 98.14 \\
    \hline
    \multicolumn{1}{c|}{\multirow{5}[1]{*}{DeSTSeg \cite{zhang2023destseg}}} 
          & A2O${\mathsmallish{(1)}}$ & 94.99 & 99.38 & 80.44 & 73.49 & 94.13 \\
          & NA${\mathsmallish{(4)}}$ & 96.16 & 99.51 & 84.20  & 73.57 & 94.11 \\
          & \cellcolor{gray!15}SCA${\mathsmallish{(4)}}^\dagger$ & \cellcolor{gray!15}\textbf{98.69} & \cellcolor{gray!15}99.66 & \cellcolor{gray!15}84.22 & \cellcolor{gray!15}75.61 & \cellcolor{gray!15}\textbf{95.42} \\
          & RA${\mathsmallish{(4)}}$ & 98.52 & \textbf{99.67} & \textbf{84.66} & 76.18 & 94.93 \\
          & O2O${\mathsmallish{(16)}}$ & 98.52 & 99.61 & 84.18 & \textbf{76.65} & 95.22 \\
    \hline
    \multicolumn{7}{c}{Screw (MVTec-2K)} \\
    \hline
    \multicolumn{1}{c|}{\multirow{3}[1]{*}{Dinomaly \cite{guo2025dinomaly}}} 
          & SCA${\mathsmallish{(4)}}$ & 97.35  & \textbf{99.94} & 73.69 & 70.81 & 99.71 \\
          & \cellcolor{gray!15}RA${\mathsmallish{(4)}}^\dagger$ & \cellcolor{gray!15}\textbf{98.13} & \cellcolor{gray!15}\textbf{99.94} & \cellcolor{gray!15}\textbf{73.80} & \cellcolor{gray!15}\textbf{70.95} & \cellcolor{gray!15}\textbf{99.79} \\
          & O2O${\mathsmallish{(16)}}$ &  95.93 & \textbf{99.94} & 70.84 & 69.41 & 99.63 \\
    \hline
    \multicolumn{1}{c|}{\multirow{3}[1]{*}{DeSTSeg \cite{zhang2023destseg}}} 
          & SCA${\mathsmallish{(4)}}$ & 95.60  & 99.35 & 49.76 & 51.32 & 97.78 \\
          & \cellcolor{gray!15}RA${\mathsmallish{(4)}}^\dagger$ & \cellcolor{gray!15}\textbf{98.48} & \cellcolor{gray!15}\textbf{99.61} & \cellcolor{gray!15}\textbf{61.29} & \cellcolor{gray!15}\textbf{61.52} & \cellcolor{gray!15}\textbf{98.57} \\
          & O2O${\mathsmallish{(16)}}$ & 97.62 & 99.44 & 58.96 & 60.22 & 98.56 \\
    \toprule
    \end{tabular}%
    }
  \label{tab:table12}%
  \vspace{-0.2cm}
\end{table}%

\begin{table}[t]
  \centering
  \caption{Ablation study results for large-scale anomalies.}
    \setlength{\tabcolsep}{2.8mm}
  \renewcommand{\arraystretch}{1.3}
  \scriptsize
    \begin{tabular}{c|ccccc}
    \bottomrule
    Method & \multicolumn{1}{c}{\textbf{I-AUC}} & \multicolumn{1}{c}{\textbf{P-AUC}} & \multicolumn{1}{c}{\textbf{P-AP}} & \multicolumn{1}{c}{\textbf{P-F1}} & \multicolumn{1}{c}{\textbf{PRO}} \\
    \hline
    DeSTSeg \cite{zhang2023destseg} & \multirow{1}[1]{*}{\textbf{99.26}} & \multirow{1}[1]{*}{96.80} & \multirow{1}[1]{*}{71.65} & \multirow{1}[1]{*}{69.06} & \multirow{1}[1]{*}{94.01} \\
    Tiling{\fontsize{6}{3}\selectfont (DeSTSeg)} \cite{rolih2024divide} & \multirow{1}[1]{*}{97.03} & \multirow{1}[1]{*}{94.32} & \multirow{1}[1]{*}{61.46} & \multirow{1}[1]{*}{59.76} & \multirow{1}[1]{*}{85.14} \\
    \cellcolor{gray!15}HiAD{\fontsize{6}{3}\selectfont (DeSTSeg)}  & \multicolumn{1}{c}{\multirow{1}[1]{*}{\cellcolor{gray!15}99.22}} & \multicolumn{1}{c}{\multirow{1}[1]{*}{\cellcolor{gray!15}\textbf{97.12}}} & \multicolumn{1}{c}{\multirow{1}[1]{*}{\cellcolor{gray!15}\textbf{72.92}}} & \multicolumn{1}{c}{\multirow{1}[1]{*}{\cellcolor{gray!15}\textbf{69.50}}} & \multicolumn{1}{c}{\multirow{1}[1]{*}{\cellcolor{gray!15}\textbf{95.42}}} \\
   \toprule
    \end{tabular}%
  \label{tab:table13}%
\end{table}%

\begin{table}[t]
  \centering
  \caption{Ablation study results under different super-resolution settings.}
  \renewcommand\arraystretch{1.15}
     \resizebox{\linewidth}{!}{
    \begin{tabular}{c|c|ccccc}
    \bottomrule
    \multicolumn{1}{c|}{Setting} & Method & \multicolumn{1}{c}{\textbf{I-AUC}} & \multicolumn{1}{c}{\textbf{P-AUC}} & \multicolumn{1}{c}{\textbf{P-AP}} & \multicolumn{1}{c}{\textbf{P-F1}} & \multicolumn{1}{c}{\textbf{PRO}} \\
    \hline
    \multicolumn{1}{c|}{\multirow{3}[1]{*}{Real-ESRGAN \cite{wang2021real}}} & PatchCore \cite{roth2022towards} & \textbf{99.49} & 99.52 & 69.53 & 68.39 & 98.20 \\
          & DeSTSeg \cite{zhang2023destseg} & 99.41 & 99.64 & \textbf{88.37} & \textbf{81.97} & \textbf{98.88} \\
          & Dinomaly \cite{guo2025dinomaly} & 92.12 & \textbf{99.77} & 78.45 & 77.18 & 98.50 \\
    \hline
    \multicolumn{1}{c|}{\multirow{3}[1]{*}{StableSR \cite{wang2024exploiting}}} & PatchCore \cite{roth2022towards} & \textbf{99.60} & 99.54 & 69.74 & 68.58 & 98.44 \\
          & DeSTSeg \cite{zhang2023destseg} & 99.56 & 99.72 & \textbf{88.40} & \textbf{82.10} & \textbf{98.82} \\
          & Dinomaly \cite{guo2025dinomaly} & 92.27 & \textbf{99.77} & 78.60  & 77.30  & 98.72 \\
    \hline
    \multicolumn{1}{c|}{\multirow{3}[1]{*}{\makecell{Train\,{\fontsize{6}{3}\selectfont(Real-ESRGAN)}\\Test\,{\fontsize{6}{3}\selectfont(StableSR)}}}} & PatchCore \cite{roth2022towards} & \textbf{99.25} & 99.49 & 69.03 & 68.23 & 97.93 \\
    \multicolumn{1}{c|}{} & DeSTSeg \cite{zhang2023destseg} & 99.10  & 99.57 & \textbf{88.07} & \textbf{81.66} & \textbf{98.55} \\
          & Dinomaly \cite{guo2025dinomaly} & 92.05 & \textbf{99.65} & 78.11 & 76.93 & 98.16 \\
    \hline
    \multicolumn{1}{c|}{\multirow{3}[1]{*}{\makecell{Train\,{\fontsize{6}{3}\selectfont(StableSR)}\\Test\,{\fontsize{6}{3}\selectfont(Real-ESRGAN)}}}} & PatchCore \cite{roth2022towards} & \textbf{99.24} & 99.50  & 69.46 & 68.32 & 98.19 \\
    \multicolumn{1}{c|}{} & DeSTSeg \cite{zhang2023destseg} & 99.13 & 99.63  & \textbf{88.05} & \textbf{81.79} & \textbf{98.42} \\
          & Dinomaly \cite{guo2025dinomaly} & 92.08 & \textbf{99.70} & 78.35 & 77.14 & 98.37 \\
    \toprule
    \end{tabular}%
    }
    \label{tab:table14}%
\end{table}%

\textit{3) Ablation Study on Large-Scale Anomalies.} Table \ref{tab:table13} reports the ablation study results of HiAD on large-scale anomalies in the MVTec-2K dataset. Detecting large-scale anomalies in high-resolution images poses an inherent challenge in modeling global context. We adopt DeSTSeg \cite{zhang2023destseg}, a fully convolutional model, as the anomaly detector. The low-resolution DeSTSeg detector operates at an input resolution of 512 $\times$ 512, whereas Tiling \cite{rolih2024divide} and HiAD operate at 2048 $\times$ 2048. The experimental results demonstrate that HiAD achieves performance comparable to its low-resolution counterpart on large-scale anomalies, while even yielding a slight improvement in localization accuracy. In contrast, Tiling exhibits a substantial performance degradation due to its limited ability to model global context. These results confirm that the low-resolution branch effectively compensates for the limited global modeling capability of patch-based detection, enabling HiAD to robustly detect large-scale anomalies while preserving high-resolution detection accuracy.

\textit{4) Ablation Study with Different Super-Resolution Settings.} Table \ref{tab:table14} validates the impact of using different super-resolution settings. Specifically, we use the Hazelnut category from the MVTec-AD dataset \cite{bergmann2019mvtec} as the subject of our investigation, which exhibits large intra-class variance and rich textural details. We adopt two widely studied super-resolution methods, Real-ESRGAN \cite{wang2021real} and StableSR \cite{wang2024exploiting}, to upscale the images from the original resolution of $ 1024 \times 1024$ to $4096 \times 4096$. In the anomaly synthesis stage, we use the same synthesis parameters to ensure that the synthetic anomalies contained in the two settings are completely consistent. In addition, we also report the experimental results under the setting of cross-evaluation. The experimental results show that different super-resolution settings have only a minor impact on anomaly detection performance, do not alter the performance ranking among the detectors, and do not affect our experimental conclusions. For industrial images, whose geometric and appearance variations are generally more constrained than those of natural images, different super-resolution methods yield broadly comparable results, confirming the reliability and robustness of our data construction and evaluation pipeline. More ablation results can be found in the \textit{Supplementary Material}.

\section{Limitations}

\textit{1) Scope and Limitations of the Synthetic Benchmarks.} Although the anomaly synthesis pipeline was carefully designed and subjected to rigorous manual quality control, the resulting high-resolution synthetic benchmarks inevitably retain certain synthesis-specific biases, as is common with other synthetic datasets \cite{zhou2023pad, li2024towards}. In particular, pasted anomalies cannot fully replicate the physical formation mechanisms or material interactions underlying real industrial defects. Moreover, synthetic benchmarks may not faithfully capture the real-world distribution of anomaly scales across different industrial products. Nevertheless, our extensive evaluations on both synthetic and natively captured real-world datasets yield consistent findings, suggesting that these potential biases do not materially affect our main conclusions. Therefore, MVTec-HD and VisA-HD can serve as reliable, controlled benchmarks for evaluating resolution scalability and sensitivity to subtle anomalies.

\textit{2) Discussion on Downsampled Evaluation.} Given the prohibitive computational cost of calculating pixel-level metrics at high resolutions, we follow common practice and perform pixel-level evaluation at an appropriately downsampled resolution. It should be emphasized, however, that the evaluation resolution for target datasets must be carefully selected according to the scale of anomalous regions, to ensure that subtle anomalies are preserved during downsampling and thereby avoid potential evaluation bias.

\section{Conclusion}

In this paper, we explore the task of high-resolution industrial image anomaly detection, which has significant practical value but has been insufficiently studied. We identify the core challenges in this domain and propose HiAD, a novel framework that employs a dual-branch architecture with multi-resolution feature fusion, achieving accurate detection of anomalous regions at varying scales while maintaining robustness to texture variations. To further enhance adaptability and efficiency, HiAD leverages a detector pool and a distribution-aware strategy selection mechanism to adaptively configure detector assignment schemes for different image distributions, facilitating practical deployment in diverse high-resolution scenarios under limited computational resources. Owing to its detector-agnostic design, HiAD provides a unified framework for extending a wide range of existing anomaly detectors to high-resolution industrial imagery while preserving their original supervision requirements and learning objectives. Furthermore, we introduce three high-resolution anomaly detection benchmarks, namely MVTec-HD, VisA-HD, and RealIAD-HD, aiming to emulate real-world high-resolution scenarios and enrich the benchmarks in this field. Extensive experiments demonstrate the effectiveness, scalability, and generalizability of HiAD across various benchmarks and task settings. In future work, we plan to develop specialized detectors for high-resolution inputs and further extend HiAD toward multimodal and language-guided zero-shot anomaly identification.

\bibliographystyle{IEEEtran}
\bibliography{main}

\begin{thebibliography}{10}
\providecommand{\url}[1]{#1}
\csname url@samestyle\endcsname
\providecommand{\newblock}{\relax}
\providecommand{\bibinfo}[2]{#2}
\providecommand{\BIBentrySTDinterwordspacing}{\spaceskip=0pt\relax}
\providecommand{\BIBentryALTinterwordstretchfactor}{4}
\providecommand{\BIBentryALTinterwordspacing}{\spaceskip=\fontdimen2\font plus
\BIBentryALTinterwordstretchfactor\fontdimen3\font minus
  \fontdimen4\font\relax}
\providecommand{\BIBforeignlanguage}[2]{{%
\expandafter\ifx\csname l@#1\endcsname\relax
\typeout{** WARNING: IEEEtran.bst: No hyphenation pattern has been}%
\typeout{** loaded for the language `#1'. Using the pattern for}%
\typeout{** the default language instead.}%
\else
\language=\csname l@#1\endcsname
\fi
#2}}
\providecommand{\BIBdecl}{\relax}
\BIBdecl

\bibitem{tao2022deep}
X.~Tao, X.~Gong, X.~Zhang, S.~Yan, and C.~Adak, ``Deep learning for
  unsupervised anomaly localization in industrial images: A survey,''
  \emph{IEEE Transactions on Instrumentation and Measurement}, pp. 1--21, 2022.

\bibitem{liu2024deep}
J.~Liu, G.~Xie, J.~Wang, S.~Li, C.~Wang, F.~Zheng, and Y.~Jin, ``Deep
  industrial image anomaly detection: A survey,'' \emph{Machine Intelligence
  Research}, vol.~21, no.~1, pp. 104--135, 2024.

\bibitem{cao2023high}
Y.~Cao, Y.~Zhang, and W.~Shen, ``High-resolution image anomaly detection via
  spatiotemporal consistency incorporated knowledge distillation,'' in
  \emph{2023 IEEE 19th International Conference on Automation Science and
  Engineering (CASE)}.\hskip 1em plus 0.5em minus 0.4em\relax IEEE, 2023, pp.
  1--6.

\bibitem{cao2025varad}
Y.~Cao, H.~Yao, W.~Luo, and W.~Shen, ``Varad: Lightweight high-resolution image
  anomaly detection via visual autoregressive modeling,'' \emph{IEEE
  Transactions on Industrial Informatics}, vol.~21, no.~4, pp. 3246--3255,
  2025.

\bibitem{rolih2024divide}
B.~Rolih, D.~Ameln, A.~Vaidya, and S.~Akcay, ``Divide and conquer:
  High-resolution industrial anomaly detection via memory efficient tiled
  ensemble,'' in \emph{Proceedings of the IEEE/CVF Conference on Computer
  Vision and Pattern Recognition Workshops}, 2024, pp. 3866--3875.

\bibitem{meininger2024unsupervised}
H.~Meininger and R.~Timofte, ``Unsupervised anomaly segmentation at high
  resolution with patch-divide-and-conquer and self-ensembling,'' in
  \emph{European Conference on Computer Vision Workshops}.\hskip 1em plus 0.5em
  minus 0.4em\relax Springer, 2024, pp. 57--73.

\bibitem{chen2019collaborative}
W.~Chen, Z.~Jiang, Z.~Wang, K.~Cui, and X.~Qian, ``Collaborative global-local
  networks for memory-efficient segmentation of ultra-high resolution images,''
  in \emph{Proceedings of the IEEE/CVF conference on computer vision and
  pattern recognition}, 2019, pp. 8924--8933.

\bibitem{guo2022isdnet}
S.~Guo, L.~Liu, Z.~Gan, Y.~Wang, W.~Zhang, C.~Wang, G.~Jiang, W.~Zhang, R.~Yi,
  L.~Ma \emph{et~al.}, ``Isdnet: Integrating shallow and deep networks for
  efficient ultra-high resolution segmentation,'' in \emph{Proceedings of the
  IEEE/CVF Conference on Computer Vision and Pattern Recognition}, 2022, pp.
  4361--4370.

\bibitem{qin2025boosting}
R.~Qin, X.~Liu, J.~Shi, L.~Lin, and J.~Yang, ``Boosting the dual-stream
  architecture in ultra-high resolution segmentation with resolution-biased
  uncertainty estimation,'' in \emph{Proceedings of the Computer Vision and
  Pattern Recognition Conference}, 2025, pp. 25\,960--25\,970.

\bibitem{heckler2026mvtec}
L.~Heckler-Kram, J.-H. Neudeck, U.~Scheler, R.~K{\"o}nig, and C.~Steger, ``The
  mvtec ad 2 dataset: Advanced scenarios for unsupervised anomaly detection,''
  \emph{International Journal of Computer Vision}, vol. 134, no.~4, p. 175,
  2026.

\bibitem{bergmann2019mvtec}
P.~Bergmann, M.~Fauser, D.~Sattlegger, and C.~Steger, ``Mvtec ad - a
  comprehensive real-world dataset for unsupervised anomaly detection,'' in
  \emph{Proceedings of the IEEE/CVF conference on computer vision and pattern
  recognition}, 2019, pp. 9592--9600.

\bibitem{zou2022spot}
Y.~Zou, J.~Jeong, L.~Pemula, D.~Zhang, and O.~Dabeer, ``Spot-the-difference
  self-supervised pre-training for anomaly detection and segmentation,'' in
  \emph{European Conference on Computer Vision}.\hskip 1em plus 0.5em minus
  0.4em\relax Springer, 2022, pp. 392--408.

\bibitem{wang2024real}
C.~Wang, W.~Zhu, B.-B. Gao, Z.~Gan, J.~Zhang, Z.~Gu, S.~Qian, M.~Chen, and
  L.~Ma, ``Real-iad: A real-world multi-view dataset for benchmarking versatile
  industrial anomaly detection,'' in \emph{Proceedings of the IEEE/CVF
  Conference on Computer Vision and Pattern Recognition}, 2024, pp.
  22\,883--22\,892.

\bibitem{roth2022towards}
K.~Roth, L.~Pemula, J.~Zepeda, B.~Sch{\"o}lkopf, T.~Brox, and P.~Gehler,
  ``Towards total recall in industrial anomaly detection,'' in
  \emph{Proceedings of the IEEE/CVF conference on computer vision and pattern
  recognition}, 2022, pp. 14\,318--14\,328.

\bibitem{defard2021padim}
T.~Defard, A.~Setkov, A.~Loesch, and R.~Audigier, ``Padim: a patch distribution
  modeling framework for anomaly detection and localization,'' in
  \emph{International Conference on Pattern Recognition}.\hskip 1em plus 0.5em
  minus 0.4em\relax Springer, 2021, pp. 475--489.

\bibitem{deng2022anomaly}
H.~Deng and X.~Li, ``Anomaly detection via reverse distillation from one-class
  embedding,'' in \emph{Proceedings of the IEEE/CVF conference on computer
  vision and pattern recognition}, 2022, pp. 9737--9746.

\bibitem{zavrtanik2021draem}
V.~Zavrtanik, M.~Kristan, and D.~Sko{\v{c}}aj, ``Dr{\ae}m--a discriminatively
  trained reconstruction embedding for surface anomaly detection,'' in
  \emph{2021 IEEE/CVF International Conference on Computer Vision
  (ICCV)}.\hskip 1em plus 0.5em minus 0.4em\relax IEEE, 2021, pp. 8310--8319.

\bibitem{liang2023omni}
Y.~Liang, J.~Zhang, S.~Zhao, R.~Wu, Y.~Liu, and S.~Pan, ``Omni-frequency
  channel-selection representations for unsupervised anomaly detection,''
  \emph{IEEE Transactions on Image Processing}, vol.~32, pp. 4327--4340, 2023.

\bibitem{zhang2023exploring}
J.~Zhang, X.~Chen, Y.~Wang, C.~Wang, Y.~Liu, X.~Li, M.-H. Yang, and D.~Tao,
  ``Exploring plain vit features for multi-class unsupervised visual anomaly
  detection,'' \emph{Computer Vision and Image Understanding}, vol. 253, p.
  104308, 2025.

\bibitem{zhang2025diffusionad}
H.~Zhang, Z.~Wang, D.~Zeng, Z.~Wu, and Y.-G. Jiang, ``Diffusionad: Norm-guided
  one-step denoising diffusion for anomaly detection,'' \emph{IEEE transactions
  on pattern analysis and machine intelligence}, vol.~47, no.~8, pp.
  7140--7152, 2025.

\bibitem{guo2025dinomaly}
J.~Guo, S.~Lu, W.~Zhang, F.~Chen, H.~Li, and H.~Liao, ``Dinomaly: The less is
  more philosophy in multi-class unsupervised anomaly detection,'' in
  \emph{Proceedings of the Computer Vision and Pattern Recognition Conference},
  2025, pp. 20\,405--20\,415.

\bibitem{yi2020patch}
J.~Yi and S.~Yoon, ``Patch svdd: Patch-level svdd for anomaly detection and
  segmentation,'' in \emph{Proceedings of the Asian conference on computer
  vision}, 2020, pp. 375--390.

\bibitem{gudovskiy2022cflow}
D.~Gudovskiy, S.~Ishizaka, and K.~Kozuka, ``Cflow-ad: Real-time unsupervised
  anomaly detection with localization via conditional normalizing flows,'' in
  \emph{Proceedings of the IEEE/CVF winter conference on applications of
  computer vision}, 2022, pp. 98--107.

\bibitem{bae2023pni}
J.~Bae, J.-H. Lee, and S.~Kim, ``Pni: Industrial anomaly detection using
  position and neighborhood information,'' in \emph{Proceedings of the IEEE/CVF
  International Conference on Computer Vision}, 2023, pp. 6373--6383.

\bibitem{yu2021fastflow}
J.~Yu, Y.~Zheng, X.~Wang, W.~Li, Y.~Wu, R.~Zhao, and L.~Wu, ``Fastflow:
  Unsupervised anomaly detection and localization via 2d normalizing flows,''
  \emph{arXiv preprint arXiv:2111.07677}, 2021.

\bibitem{liu2024cross}
B.~Liu, T.~Guo, B.~Luo, Z.~Cui, and J.~Yang, ``Cross-attention regression flow
  for defect detection,'' \emph{IEEE Transactions on Image Processing},
  vol.~33, pp. 5183--5193, 2024.

\bibitem{li2021cutpaste}
C.-L. Li, K.~Sohn, J.~Yoon, and T.~Pfister, ``Cutpaste: Self-supervised
  learning for anomaly detection and localization,'' in \emph{Proceedings of
  the IEEE/CVF conference on computer vision and pattern recognition}, 2021,
  pp. 9664--9674.

\bibitem{schluter2022natural}
H.~M. Schl{\"u}ter, J.~Tan, B.~Hou, and B.~Kainz, ``Natural synthetic anomalies
  for self-supervised anomaly detection and localization,'' in \emph{European
  Conference on Computer Vision}.\hskip 1em plus 0.5em minus 0.4em\relax
  Springer, 2022, pp. 474--489.

\bibitem{zhang2023prototypical}
H.~Zhang, Z.~Wu, Z.~Wang, Z.~Chen, and Y.-G. Jiang, ``Prototypical residual
  networks for anomaly detection and localization,'' in \emph{Proceedings of
  the IEEE/CVF conference on computer vision and pattern recognition}, 2023,
  pp. 16\,281--16\,291.

\bibitem{zhang2024realnet}
X.~Zhang, M.~Xu, and X.~Zhou, ``Realnet: A feature selection network with
  realistic synthetic anomaly for anomaly detection,'' in \emph{Proceedings of
  the IEEE/CVF Conference on Computer Vision and Pattern Recognition}, 2024,
  pp. 16\,699--16\,708.

\bibitem{wang2026stones}
S.~Wang, Y.~Hu, X.~Liu, S.~Wang, G.~Wang, C.~Xu, J.~Liu, and P.~Chen, ``Stones
  from other hills can polish jade: Zero-shot anomaly synthesis via
  cross-domain anomaly injection,'' \emph{IEEE Transactions on Image
  Processing}, vol.~35, pp. 3662--3674, 2026.

\bibitem{horwath2020understanding}
J.~P. Horwath, D.~N. Zakharov, R.~M{\'e}gret, and E.~A. Stach, ``Understanding
  important features of deep learning models for segmentation of
  high-resolution transmission electron microscopy images,'' \emph{npj
  Computational Materials}, vol.~6, no.~1, p. 108, 2020.

\bibitem{fu2024towards}
G.~Fu, Q.~Zhang, L.~Zhu, Q.~Lin, Y.~Wang, S.~Fan, and C.~Xiao, ``Towards
  high-resolution specular highlight detection,'' \emph{International Journal
  of Computer Vision}, vol. 132, no.~1, pp. 95--117, 2024.

\bibitem{cordts2016cityscapes}
M.~Cordts, M.~Omran, S.~Ramos, T.~Rehfeld, M.~Enzweiler, R.~Benenson,
  U.~Franke, S.~Roth, and B.~Schiele, ``The cityscapes dataset for semantic
  urban scene understanding,'' in \emph{Proceedings of the IEEE conference on
  computer vision and pattern recognition}, 2016, pp. 3213--3223.

\bibitem{zheng2021deep}
Z.~Zheng, X.~Li, Q.~Xu, and X.~Song, ``Deep inference networks for reliable
  vehicle lateral position estimation in congested urban environments,''
  \emph{IEEE transactions on image processing}, vol.~30, pp. 8368--8383, 2021.

\bibitem{ding2025hilm}
X.~Ding, J.~Han, H.~Xu, W.~Zhang, and X.~Li, ``Hilm-d: Enhancing mllms with
  multi-scale high-resolution details for autonomous driving,''
  \emph{International Journal of Computer Vision}, vol. 133, no.~8, pp.
  5379--5395, 2025.

\bibitem{jiang2022survey}
H.~Jiang, M.~Peng, Y.~Zhong, H.~Xie, Z.~Hao, J.~Lin, X.~Ma, and X.~Hu, ``A
  survey on deep learning-based change detection from high-resolution remote
  sensing images,'' \emph{Remote Sensing}, vol.~14, no.~7, p. 1552, 2022.

\bibitem{li2023anomaly}
J.~Li, X.~Wang, H.~Zhao, S.~Wang, and Y.~Zhong, ``Anomaly segmentation for
  high-resolution remote sensing images based on pixel descriptors,'' in
  \emph{Proceedings of the AAAI Conference on Artificial Intelligence},
  vol.~37, no.~4, 2023, pp. 4426--4434.

\bibitem{zheng2024single}
Z.~Zheng, Y.~Zhong, A.~Ma, and L.~Zhang, ``Single-temporal supervised learning
  for universal remote sensing change detection,'' \emph{International Journal
  of Computer Vision}, vol. 132, no.~12, pp. 5582--5602, 2024.

\bibitem{van2021deep}
J.~Van~der Laak, G.~Litjens, and F.~Ciompi, ``Deep learning in histopathology:
  the path to the clinic,'' \emph{Nature medicine}, vol.~27, no.~5, pp.
  775--784, 2021.

\bibitem{cheng2021robust}
S.~Cheng, S.~Liu, J.~Yu, G.~Rao, Y.~Xiao, W.~Han, W.~Zhu, X.~Lv, N.~Li, J.~Cai
  \emph{et~al.}, ``Robust whole slide image analysis for cervical cancer
  screening using deep learning,'' \emph{Nature communications}, vol.~12,
  no.~1, p. 5639, 2021.

\bibitem{li2023task}
H.~Li, C.~Zhu, Y.~Zhang, Y.~Sun, Z.~Shui, W.~Kuang, S.~Zheng, and L.~Yang,
  ``Task-specific fine-tuning via variational information bottleneck for
  weakly-supervised pathology whole slide image classification,'' in
  \emph{Proceedings of the IEEE/CVF conference on computer vision and pattern
  recognition}, 2023, pp. 7454--7463.

\bibitem{gu2024mamba}
A.~Gu and T.~Dao, ``Mamba: Linear-time sequence modeling with selective state
  spaces,'' in \emph{First Conference on Language Modeling}, 2024, pp. 1--32.

\bibitem{you2022unified}
Z.~You, L.~Cui, Y.~Shen, K.~Yang, X.~Lu, Y.~Zheng, and X.~Le, ``A unified model
  for multi-class anomaly detection,'' \emph{Advances in Neural Information
  Processing Systems}, vol.~35, pp. 4571--4584, 2022.

\bibitem{jacobs1991adaptive}
R.~A. Jacobs, M.~I. Jordan, S.~J. Nowlan, and G.~E. Hinton, ``Adaptive mixtures
  of local experts,'' \emph{Neural computation}, pp. 79--87, 1991.

\bibitem{shazeer2017outrageously}
N.~Shazeer, A.~Mirhoseini, K.~Maziarz, A.~Davis, Q.~Le, G.~Hinton, and J.~Dean,
  ``Outrageously large neural networks: The sparsely-gated mixture-of-experts
  layer,'' in \emph{International Conference on Learning Representations},
  2017, pp. 1--12.

\bibitem{wang2024exploiting}
J.~Wang, Z.~Yue, S.~Zhou, K.~C. Chan, and C.~C. Loy, ``Exploiting diffusion
  prior for real-world image super-resolution,'' \emph{International Journal of
  Computer Vision}, vol. 132, no.~12, pp. 5929--5949, 2024.

\bibitem{perez2003poisson}
P.~P{\'e}rez, M.~Gangnet, and A.~Blake, ``Poisson image editing,'' \emph{ACM
  Transactions on Graphics}, vol.~22, no.~3, pp. 313--318, 2003.

\bibitem{lehr2024ad3}
J.~Lehr, J.~Philipps, A.~Sargsyan, M.~Pape, and J.~Kr{\"u}ger, ``Ad3:
  Introducing a score for anomaly detection dataset difficulty assessment using
  viaduct dataset,'' in \emph{European Conference on Computer Vision}.\hskip
  1em plus 0.5em minus 0.4em\relax Springer, 2024, pp. 449--464.

\bibitem{tien2023revisiting}
T.~D. Tien, A.~T. Nguyen, N.~H. Tran, T.~D. Huy, S.~Duong, C.~D.~T. Nguyen, and
  S.~Q. Truong, ``Revisiting reverse distillation for anomaly detection,'' in
  \emph{Proceedings of the IEEE/CVF conference on computer vision and pattern
  recognition}, 2023, pp. 24\,511--24\,520.

\bibitem{zhang2023destseg}
X.~Zhang, S.~Li, X.~Li, P.~Huang, J.~Shan, and T.~Chen, ``Destseg: Segmentation
  guided denoising student-teacher for anomaly detection,'' in
  \emph{Proceedings of the IEEE/CVF Conference on Computer Vision and Pattern
  Recognition}, 2023, pp. 3914--3923.

\bibitem{macqueen1967some}
J.~MacQueen, ``Some methods for classification and analysis of multivariate
  observations,'' in \emph{Proceedings of the 5th Berkeley symposium on
  mathematical statistics and probability}, vol.~1, 1967, pp. 281--297.

\bibitem{batzner2024efficientad}
K.~Batzner, L.~Heckler, and R.~K{\"o}nig, ``Efficientad: Accurate visual
  anomaly detection at millisecond-level latencies,'' in \emph{Proceedings of
  the IEEE/CVF Winter Conference on Applications of Computer Vision}, 2024, pp.
  128--138.

\bibitem{simeoni2025dinov3}
O.~Sim{\'e}oni, H.~V. Vo, M.~Seitzer, F.~Baldassarre, M.~Oquab, C.~Jose,
  V.~Khalidov, M.~Szafraniec, S.~Yi, M.~Ramamonjisoa \emph{et~al.}, ``Dinov3,''
  \emph{arXiv preprint arXiv:2508.10104}, 2025.

\bibitem{radford2021learning}
A.~Radford, J.~W. Kim, C.~Hallacy, A.~Ramesh, G.~Goh, S.~Agarwal, G.~Sastry,
  A.~Askell, P.~Mishkin, J.~Clark \emph{et~al.}, ``Learning transferable visual
  models from natural language supervision,'' in \emph{International conference
  on machine learning}.\hskip 1em plus 0.5em minus 0.4em\relax PMLR, 2021, pp.
  8748--8763.

\bibitem{zhou2024anomalyclip}
Q.~Zhou, G.~Pang, Y.~Tian, S.~He, and J.~Chen, ``Anomalyclip: Object-agnostic
  prompt learning for zero-shot anomaly detection,'' in \emph{International
  Conference on Learning Representations}, 2024, pp. 49\,705--49\,737.

\bibitem{wang2021real}
X.~Wang, L.~Xie, C.~Dong, and Y.~Shan, ``Real-esrgan: Training real-world blind
  super-resolution with pure synthetic data,'' in \emph{Proceedings of the
  IEEE/CVF International Conference on Computer Vision Workshops}, 2021, pp.
  1905--1914.

\bibitem{zhou2023pad}
Q.~Zhou, W.~Li, L.~Jiang, G.~Wang, G.~Zhou, S.~Zhang, and H.~Zhao, ``Pad: A
  dataset and benchmark for pose-agnostic anomaly detection,'' \emph{Advances
  in Neural Information Processing Systems}, vol.~36, pp. 44\,558--44\,571,
  2023.

\bibitem{li2024towards}
W.~Li, X.~Xu, Y.~Gu, B.~Zheng, S.~Gao, and Y.~Wu, ``Towards scalable 3d anomaly
  detection and localization: A benchmark via 3d anomaly synthesis and a
  self-supervised learning network,'' in \emph{Proceedings of the IEEE/CVF
  conference on computer vision and pattern recognition}, 2024, pp.
  22\,207--22\,216.

\end{thebibliography}

\clearpage

\renewcommand*{\thefigure}{S\arabic{figure}}
\renewcommand*{\thetable}{S\arabic{table}}
\renewcommand*{\theequation}{S\arabic{equation}}
\renewcommand*{\thealgorithm}{S\arabic{algorithm}}
\renewcommand\thesection{\Alph{section}}
\setcounter{table}{0}
\setcounter{figure}{0}
\setcounter{equation}{0}
\setcounter{section}{0}
\setcounter{algorithm}{0}

\renewcommand*{\theHtable}{supp.\arabic{table}}
\renewcommand*{\theHfigure}{supp.\arabic{figure}}
\renewcommand*{\theHequation}{supp.\arabic{equation}}
\renewcommand*{\theHsection}{supp.\arabic{section}}

\providecommand*{\theHalgorithm}{}
\renewcommand*{\theHalgorithm}{supp.\arabic{algorithm}}

\renewcommand{\algorithmicrequire}{\textbf{Input:}}
\renewcommand{\algorithmicensure}{\textbf{Output:}}
\newcommand{\rev}[1]{\textcolor{blue}{#1}}

\markboth{Supplementary Material}%
{Shell \MakeLowercase{\textit{et al.}}: Towards High-Resolution Industrial Image Anomaly Detection}

\section{Overview}

We organize this supplementary material into the following sections: Appendix \ref{sec: supB} describes the detailed workflow of the Adaptive Strategy Selection. Appendix \ref{sec: supC} provides additional ablation studies investigating the effects of the downsampling rate in MRFF, the detector pool size, the patch size in the high-resolution branch, and the evaluation resolution for pixel-level metric computation. Appendix \ref{sec: supD} presents supplementary visualization results, and Appendix \ref{sec: supE} provides the metrics for each category.

\section{Workflow of Adaptive Strategy Selection}
\label{sec: supB} 
Algorithm~\ref{alg:ass} summarizes the workflow of Adaptive Strategy Selection. Specifically, we first estimate the detector-wise statistics of each candidate strategy on the training set, and then compute its ASS score using the held-out query features. The strategy achieving the lowest ASS score is selected for subsequent detector assignment. Notably, this selection procedure does not involve computationally expensive operations, and therefore its computational cost is negligible.

\section{Supplementary Ablation Studies}
\label{sec: supC}
\begin{figure}[t]
  \centering
   \includegraphics[width=\linewidth]{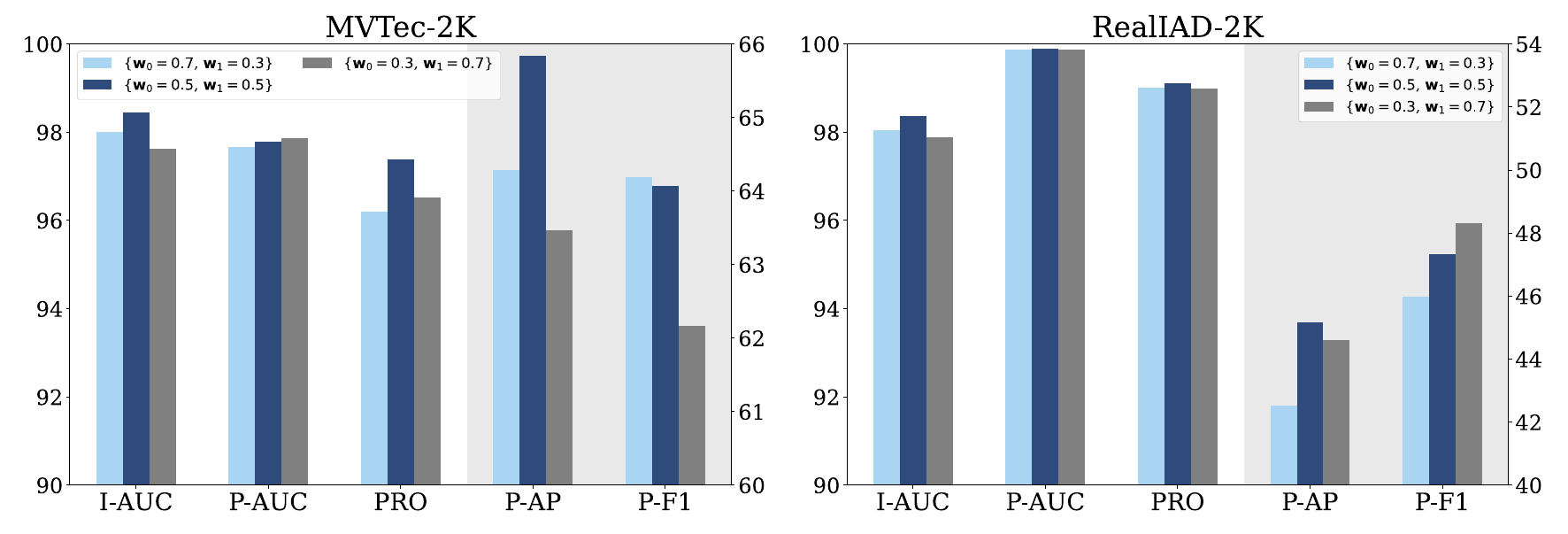}
   \caption{Ablation study on fusion weights in MRFF using downsampling rates of $\{0,1\}$, evaluated with PatchCore \cite{roth2022towards} at a resolution of $2048 \times 2048$.}
   \label{fig:figs1}
\end{figure}

\textit{1) Ablation Study on Downsampling Rate.} Table \ref{tab:tables1} investigates the impact of the downsampling rate on the MRFF, with experiments conducted at $2048 \times 2048$. The downsampling rate $\{0\}$ represents the use of the original pre-trained features for anomaly detection, while $\{0,1\}$ and $\{0,1,2\}$ respectively correspond to progressively downsampling the images to $1024 \times 1024$ and $512 \times 512$ for multi-resolution feature extraction. The experimental results demonstrate that fusing appropriately downsampled features generally improves anomaly detection performance; however, incorporating features from very low-resolution images may impair the detection of subtle anomalies, leading to performance degradation. To achieve a desirable trade-off between suppressing ultrafine texture noise and preserving fine-grained anomaly information, the downsampling rates are set to $\{0, 1\}$ for 2K resolution, and $\{0, 1, 2\}$ for 4K resolution. Furthermore, through fusion weights, we can further refine the control over the multi-resolution fusion features, as shown in Figure \ref{fig:figs1}. Theoretically, the multi-resolution fusion process introduces a risk of impairing highly localized anomaly signals. However, in our experiments, selecting appropriate downsampling rates and fusion weights does not cause noticeable degradation, and any minor impact is negligible compared to the substantial reduction in over-detection. 

\begin{algorithm}[t]
\caption{Adaptive Strategy Selection (ASS)} 
\label{alg:ass}
\begin{algorithmic}[1]
\REQUIRE Training feature set $\mathcal{F}_{\rm t}$; query feature set $\mathcal{F}_{\rm q}$; 
strategy pool $\mathcal{S}$; detector pool size $M$; trade-off parameter $\lambda$.
\ENSURE The selected assignment strategy $s^{*}$.
\FOR{each assignment strategy $s \in \mathcal{S}$}
    \STATE Assign training features in $\mathcal{F}_{\rm t}$ to $M$ detector-wise feature groups using $\Phi_{\mathcal{D}}^{s}$;
    \STATE Estimate detector-wise statistics $\{\boldsymbol{\mu}_m^s,\boldsymbol{\Sigma}_m^s\}_{m=1}^{M}$;
    \STATE Assign query features in $\mathcal{F}_{\rm q}$ using $\Phi_{\mathcal{D}}^{s}$;
    \STATE Compute the ASS score: 
    $\mathcal{L}_{\rm ASS}(s) = \mathcal{L}_{\rm FAF}(s) + \lambda \mathcal{L}_{\rm SPC}(s)$;
\ENDFOR
\STATE Select the optimal strategy: $s^{*} = \arg\min_{s\in\mathcal{S}} \mathcal{L}_{\rm ASS}(s)$;
\RETURN $s^{*}$
\end{algorithmic}
\end{algorithm}

\textit{2) Ablation Study on Detector Pool Size.} We further investigate the impact of the detector pool size in HiAD, with the experimental results presented in Table \ref{tab:tables2}. As the number of detectors increases, the number of image patches assigned to each detector decreases, which can enhance anomaly detection performance but also increases computational cost. For instance, increasing the detector pool size from 4 to 8 yields only marginal improvements in performance, while doubling both training duration and GPU memory usage. Therefore, we suggest setting an appropriate detector pool size based on the available computational resources to balance efficiency and anomaly detection performance.

\begin{table*}[t!] 
    \centering
    \renewcommand{\arraystretch}{1.2}
    \setlength{\tabcolsep}{3.0mm}
    \caption{Ablation study results on the downsampling rate in MRFF, where multi-resolution features are fused with equal weights at a resolution of $2048 \times 2048$.}
    \label{tab:tables1}
    \begin{tabular}{c|c|ccccc|ccccc}
    \bottomrule
    \multicolumn{1}{c|}{\multirow{2}[1]{*}{Method}} & \multirow{2}[1]{*}{Downsampling Rate} & \multicolumn{5}{c|}{MVTec-2K} & \multicolumn{5}{c}{RealIAD-2K} \\
\cline{3-12}          & \multicolumn{1}{c|}{} & \multicolumn{1}{c}{\textbf{I-AUC}} & \multicolumn{1}{c}{\textbf{P-AUC}} & \multicolumn{1}{c}{\textbf{P-AP}} & \multicolumn{1}{c}{\textbf{P-F1}} & \multicolumn{1}{c|}{\textbf{PRO}} & \multicolumn{1}{c}{\textbf{I-AUC}} & \multicolumn{1}{c}{\textbf{P-AUC}} & \multicolumn{1}{c}{\textbf{P-AP}} & \multicolumn{1}{c}{\textbf{P-F1}} & \multicolumn{1}{c}{\textbf{PRO}} \\
    \hline
    \multicolumn{1}{c|}{\multirow{3}[1]{*}{Dinomaly \cite{guo2025dinomaly}}} 
          & $\{0\}$        & 94.98 & 98.69 & 70.02 & 67.54 & 96.24 & 93.74 & 99.72 & 36.21 &  38.35 & 98.10 \\
          & \cellcolor{gray!15}$\{0, 1\}$     & \cellcolor{gray!15}\textbf{97.17} & \cellcolor{gray!15}99.07 & \cellcolor{gray!15}\textbf{72.98} & \cellcolor{gray!15}\textbf{69.10} & \cellcolor{gray!15}\textbf{97.58} & \cellcolor{gray!15}\textbf{94.26} & \cellcolor{gray!15}\textbf{99.86} & \cellcolor{gray!15}\textbf{39.62} & \cellcolor{gray!15}\textbf{40.88} &  \cellcolor{gray!15}\textbf{99.23}  \\
          & $\{0, 1, 2\}$  & 96.06 & \textbf{99.24} & 71.75 & 68.54 & 97.40 & 92.91 & 99.84 & 37.30 & 36.54 & 98.48 \\
    \hline
    \multicolumn{1}{c|}{\multirow{3}[1]{*}{ViTAD \cite{zhang2023exploring}}} 
          & $\{0\}$  & 95.45 & 98.74 & 63.91 & 62.50 & 97.17 & 93.59 & 99.74 & 34.64 & 41.77 & 97.96 \\ 
          & \cellcolor{gray!15}$\{0, 1\}$ & \cellcolor{gray!15}\textbf{96.45} & \cellcolor{gray!15}\textbf{99.06} & \cellcolor{gray!15}\textbf{64.78} & \cellcolor{gray!15}64.09 & \cellcolor{gray!15}\textbf{97.30} & \cellcolor{gray!15}\textbf{94.47} & \cellcolor{gray!15}\textbf{99.79} & \cellcolor{gray!15}\textbf{36.83} & \cellcolor{gray!15}\textbf{41.88} & \cellcolor{gray!15}\textbf{98.07} \\
          & $\{0, 1, 2\}$ & 95.51 & 98.96 & 63.94 & \textbf{64.14} & 95.62 & 93.40 & 99.71 & 28.82 & 35.40 & 97.68 \\
    \hline
    \multicolumn{1}{c|}{\multirow{3}[1]{*}{PatchCore \cite{roth2022towards}}} 
          & $\{0\}$   & 98.37 & 97.69 & 64.74 & 63.05 & 97.26 & 97.50  & 99.85 & 43.03 & 46.95 & 99.03 \\
          & \cellcolor{gray!15}$\{0, 1\}$ & \cellcolor{gray!15}\textbf{98.44} & \cellcolor{gray!15}97.79 & \cellcolor{gray!15}65.84 & \cellcolor{gray!15}64.07 & \cellcolor{gray!15}\textbf{97.39} & \cellcolor{gray!15}\textbf{98.37} & \cellcolor{gray!15}\textbf{99.89} & \cellcolor{gray!15}\textbf{45.16} & \cellcolor{gray!15}\textbf{47.34} & \cellcolor{gray!15}\textbf{99.10} \\
          & $\{0, 1, 2\}$ & 98.04 & \textbf{97.91} & \textbf{66.47} & \textbf{64.68} & 97.26 & 98.34 & 99.85 & 41.86 & 45.33 & 98.48 \\
    \hline
    \multicolumn{1}{c|}{\multirow{3}[1]{*}{PaDiM \cite{defard2021padim}}} 
          & $\{0\}$   & 81.03 & 96.74 & 37.82 & 41.79 & 94.53 & 86.66 & 99.58 & 18.99 & 27.64 & \textbf{98.06} \\
          & \cellcolor{gray!15}$\{0, 1\}$ & \cellcolor{gray!15}\textbf{86.22} & \cellcolor{gray!15}98.13 & \cellcolor{gray!15}44.43 & \cellcolor{gray!15}48.50  & \cellcolor{gray!15}\textbf{95.95} & \cellcolor{gray!15}\textbf{89.67} & \cellcolor{gray!15}\textbf{99.60} & \cellcolor{gray!15}\textbf{20.74} & \cellcolor{gray!15}\textbf{28.22} & \cellcolor{gray!15}97.98 \\
          & $\{0, 1, 2\}$ & 85.79 & \textbf{98.14} & \textbf{44.89} & \textbf{48.87} & 95.04 & 88.47 & 99.41 & 20.30  & 27.86 & 97.99 \\
    \toprule
    \end{tabular}%
    
    \vspace{5mm} 
    
    \renewcommand{\arraystretch}{1.2}
    \setlength{\tabcolsep}{3.1mm}
    \caption{Ablation study results on the size of the detector pool at a resolution of $2048 \times 2048$.}
    \label{tab:tables2}
    \begin{tabular}{c|c|ccccc|ccccc}
    \bottomrule
    \multicolumn{1}{c|}{\multirow{2}[1]{*}{Method}} & \multicolumn{1}{c|}{\multirow{2}[1]{*}{Detector Pool Size}} & \multicolumn{5}{c|}{MVTec-2K} & \multicolumn{5}{c}{RealIAD-2K} \\
    \cline{3-12}          &       & \multicolumn{1}{c}{\textbf{I-AUC}} & \multicolumn{1}{c}{\textbf{P-AUC}} & \multicolumn{1}{c}{\textbf{P-AP}} & \multicolumn{1}{c}{\textbf{P-F1}} & \multicolumn{1}{c|}{\textbf{PRO}} & \multicolumn{1}{c}{\textbf{I-AUC}} & \multicolumn{1}{c}{\textbf{P-AUC}} & \multicolumn{1}{c}{\textbf{P-AP}} & \multicolumn{1}{c}{\textbf{P-F1}} & \multicolumn{1}{c}{\textbf{PRO}} \\
    \hline
    \multicolumn{1}{c|}{\multirow{4}[1]{*}{RealNet \cite{zhang2024realnet}}} 
          & 1     & 93.13 & 95.91 & 57.65 & 57.37 & 88.42 & 91.03 & 99.58 & 45.46 & 48.13 & 91.84 \\
          & 2     & 95.53  & 96.74 & 59.45  & 60.12  & 90.51 & 92.29 & 99.60  & 45.84 & \textbf{49.16} & 92.37 \\
          & \cellcolor{gray!15}4     & \cellcolor{gray!15}\textbf{96.18} & \cellcolor{gray!15}96.85 & \cellcolor{gray!15}63.48 & \cellcolor{gray!15}\textbf{62.21} & \cellcolor{gray!15}\textbf{91.59}  & \cellcolor{gray!15}93.02 & \cellcolor{gray!15}\textbf{99.63} & \cellcolor{gray!15}\textbf{46.84} & \cellcolor{gray!15}48.86 & \cellcolor{gray!15}92.71 \\
          & 8     & 96.02 & \textbf{97.09} & \textbf{63.79} & 61.23 & 91.34 & \textbf{93.18} & 99.60  & 43.82 & 48.32 & \textbf{93.54} \\
    \hline
    \multicolumn{1}{c|}{\multirow{4}[1]{*}{ViTAD \cite{zhang2023exploring}}} 
          & 1     & 95.39 & 98.50 & 63.80 & 62.09 & 96.54 & 91.56 & 99.59 & 30.66 & 38.21 & 97.26 \\
          & 2     & 95.80 & 98.84 & 64.45  & 63.18  & 96.51 & 92.51 & 99.70  & 29.51 & 36.69 & 98.02 \\
          & \cellcolor{gray!15}4     & \cellcolor{gray!15}96.45 & \cellcolor{gray!15}\textbf{99.06} &  \cellcolor{gray!15}64.78 &  \cellcolor{gray!15}64.09  & \cellcolor{gray!15}\textbf{97.30}  & \cellcolor{gray!15}94.47 & \cellcolor{gray!15}\textbf{99.79} & \cellcolor{gray!15}\textbf{36.83} & \cellcolor{gray!15}\textbf{41.88} & \cellcolor{gray!15}\textbf{98.07} \\
          & 8     & \textbf{96.62} & 98.59 & \textbf{64.89} & \textbf{64.20}   & 96.64 & \textbf{94.64} & 99.77  & 35.82 & 41.37 & 98.01 \\
    \hline
    \multicolumn{1}{c|}{\multirow{4}[1]{*}{DeSTSeg \cite{zhang2023destseg}}} 
          & 1     & 95.10  & 97.40  & 70.22 & 66.34 & 96.49 & 90.11 & 99.37 & 52.78 & 53.31 & 94.61 \\
          & 2     & 96.15 & \textbf{97.81} & 70.32 & 66.32 & 96.30 & 92.37 & \textbf{99.50}  & 51.43 & 51.99 & 93.67 \\
          & \cellcolor{gray!15}4     & \cellcolor{gray!15}\textbf{97.93} & \cellcolor{gray!15}97.33 & \cellcolor{gray!15}70.70 & \cellcolor{gray!15}\textbf{68.33}  & \cellcolor{gray!15}96.51 & \cellcolor{gray!15}93.50  & \cellcolor{gray!15}99.46 & \cellcolor{gray!15}\textbf{54.67} & \cellcolor{gray!15}\textbf{53.96} & \cellcolor{gray!15}\textbf{95.55} \\
          & 8     & 97.84 & 97.40 & \textbf{70.83} & 68.14 & \textbf{96.70}  & \textbf{93.88} & 99.31 & 50.11 & 52.86 & 94.56 \\
    \toprule
    \end{tabular}%
    
   \vspace{5mm} 
   
  \renewcommand\arraystretch{1.3}
  \setlength{\tabcolsep}{3.45mm}
  \caption{Ablation study results on patch size in the high-resolution branch at a resolution of $2048 \times 2048$.}
    \label{tab:tables3}%
    \begin{tabular}{c|c|ccccc|ccccc}
    \bottomrule
    \multicolumn{1}{c|}{\multirow{2}[1]{*}{Method}} & \multirow{2}[1]{*}{Patch Size} & \multicolumn{5}{c|}{MVTec-2K} & \multicolumn{5}{c}{RealIAD-2K} \\
\cline{3-12}          & \multicolumn{1}{c|}{} & \multicolumn{1}{c}{\textbf{I-AUC}} & \multicolumn{1}{c}{\textbf{P-AUC}} & \multicolumn{1}{c}{\textbf{P-AP}} & \multicolumn{1}{c}{\textbf{P-F1}} & \multicolumn{1}{c|}{\textbf{PRO}} & \multicolumn{1}{c}{\textbf{I-AUC}} & \multicolumn{1}{c}{\textbf{P-AUC}} & \multicolumn{1}{c}{\textbf{P-AP}} & \multicolumn{1}{c}{\textbf{P-F1}} & \multicolumn{1}{c}{\textbf{PRO}} \\
\hline
    \multicolumn{1}{c|}{\multirow{3}[1]{*}{PatchCore \cite{roth2022towards}}} & \cellcolor{gray!15}{$512 \times 512$} & \cellcolor{gray!15}\textbf{98.44} & \cellcolor{gray!15}97.79 & \cellcolor{gray!15}\textbf{65.84} & \cellcolor{gray!15}\textbf{64.07} & \cellcolor{gray!15}97.39 & \cellcolor{gray!15}\textbf{98.37} & \cellcolor{gray!15}\textbf{99.89} & \cellcolor{gray!15}\textbf{45.16} & \cellcolor{gray!15}47.34 & \cellcolor{gray!15}99.10 \\
          & {$384 \times 384$} & 98.13 & 97.66 & 65.75 & 63.97 & \textbf{97.45} & 97.01 & 99.88 & 44.62 & 47.36 & 99.09 \\
          & {$256 \times 256$} & 97.61 & \textbf{98.02} & 64.71 & 63.07 & 97.40  & 97.60  & 99.87 & 44.74 & \textbf{47.89} & \textbf{99.14} \\
    \hline
    \multicolumn{1}{c|}{\multirow{3}[1]{*}{DeSTSeg \cite{zhang2023destseg}}} & \cellcolor{gray!15}{$512 \times 512$} & \cellcolor{gray!15}\textbf{97.93} & \cellcolor{gray!15}97.33 & \cellcolor{gray!15}\textbf{70.70} & \cellcolor{gray!15}\textbf{68.33} & \cellcolor{gray!15}96.51 & \cellcolor{gray!15}\textbf{93.50} & \cellcolor{gray!15}99.46 & \cellcolor{gray!15}\textbf{54.67} & \cellcolor{gray!15}\textbf{53.96} & \cellcolor{gray!15}95.55 \\
          & {$384 \times 384$} & 97.87  & \textbf{97.56} & 70.18 & 67.76 & \textbf{96.61} & 91.18 & \textbf{99.54} & 49.29 & 51.50 & \textbf{95.68}\\
          & {$256 \times 256$} & 97.36 & 97.21 & 69.92 & 67.12 & 96.38 & 89.65 & 99.31 & 48.01 & 50.73 & 94.14 \\
    \hline
    \multicolumn{1}{c|}{\multirow{3}[1]{*}{FastFlow \cite{yu2021fastflow}}} & \cellcolor{gray!15}{$512 \times 512$} & \cellcolor{gray!15}\textbf{91.71} & \cellcolor{gray!15}\textbf{98.52} & \cellcolor{gray!15}\textbf{54.22} & \cellcolor{gray!15}\textbf{55.44} & \cellcolor{gray!15}\textbf{94.48} & \cellcolor{gray!15}\textbf{93.66} & \cellcolor{gray!15}99.32 & \cellcolor{gray!15}\textbf{25.97} & \cellcolor{gray!15}\textbf{32.82} & \cellcolor{gray!15}96.79 \\
          & {$384 \times 384$} & 89.24 & 97.61 & 48.90 & 51.91 & 94.08 & 91.88 & \textbf{99.45} & 22.20  & 29.84 & \textbf{96.80} \\
          & {$256 \times 256$} & 89.90 & 97.90 & 50.91 & 52.85 & 93.40 & 92.98 & \textbf{99.45} & 21.89 & 29.70 & 96.57 \\          
    \toprule
    \end{tabular}
\end{table*}

\textit{3) Ablation Study on Patch Size.} Table \ref{tab:tables3} provides the experimental results of HiAD with different patch sizes. The results indicate that increasing the patch size to reduce the number of image patches typically improves anomaly detection performance. Moreover, we always set the stride size equal to the patch size, which means there is no overlap between patches. In our experiments, we do not observe a significant impact of the stride size on the results. This may be attributed to the fact that HiAD includes a low-resolution branch, enabling it to continuously detect large-scale anomalies spanning multiple patches without the need for repeated detection of patch edges. 

\begin{table}[t]
  \centering
  \caption{Ablation study results for different evaluation resolutions, using PatchCore as the detector.}
  \renewcommand\arraystretch{1.2}
   \setlength{\tabcolsep}{2.5mm}
    \begin{tabular}{c|cccc}
    \bottomrule
    \multicolumn{5}{c}{MVTec-2K} \\
    \hline
    Evaluation Resolution & \multicolumn{1}{c}{P-AUC} & \multicolumn{1}{c}{P-AP} & \multicolumn{1}{c}{P-F1} & \multicolumn{1}{c}{PRO} \\
    \hline
    2048 $\times$ 2048 & 97.7823 & 65.7886 & 64.0192 & 97.3480 \\
    1024 $\times$ 1024 & 97.7845 & 65.8274 & 64.0401 & 97.3616 \\
    \cellcolor{gray!15}512 $\times$ 512   & \cellcolor{gray!15}97.7912 & \cellcolor{gray!15}65.8409 & \cellcolor{gray!15}64.0747 & \cellcolor{gray!15}97.3924 \\
    256 $\times$ 256   & 97.8015 & 65.8736 & 64.1051 & 97.4519 \\
    \hline
    \multicolumn{5}{c}{RealIAD-2K} \\
    \hline
    Evaluation Resolution & \multicolumn{1}{c}{P-AUC} & \multicolumn{1}{c}{P-AP} & \multicolumn{1}{c}{P-F1} & \multicolumn{1}{c}{PRO} \\
    \hline
    2048 $\times$ 2048 & 99.8853 & 45.1112 & 47.2665 & 99.0517 \\
    1024 $\times$ 1024 & 99.8892 & 45.1596 & 47.3025 & 99.0796 \\
    \cellcolor{gray!15}512 $\times$ 512   & \cellcolor{gray!15}99.8930 & \cellcolor{gray!15}45.1590 & \cellcolor{gray!15}47.3367 & \cellcolor{gray!15}99.1049 \\
    256 $\times$ 256   & 99.8994 & 45.2173 & 47.4530 & 99.2172 \\
    \hline
    \multicolumn{5}{c}{MVTec-4K} \\
    \hline
    Evaluation Resolution & \multicolumn{1}{c}{P-AUC} & \multicolumn{1}{c}{P-AP} & \multicolumn{1}{c}{P-F1} & \multicolumn{1}{c}{PRO} \\
    \hline
    4096 $\times$ 4096 & 97.8225 & 63.1394 & 61.9607 & 97.1452 \\
    2048 $\times$ 2048 & 97.8245 & 63.1532 & 61.9777 & 97.2254 \\
    1024 $\times$ 1024 & 97.8361 & 63.1946 & 62.0538 & 97.2521 \\
    \cellcolor{gray!15}512 $\times$ 512   & \cellcolor{gray!15}97.8495 & \cellcolor{gray!15}63.2326 & \cellcolor{gray!15}62.0902 & \cellcolor{gray!15}97.2958 \\
    256 $\times$ 256   & 97.8808 & 63.3980 & 62.2730 & 97.4993 \\
    \toprule
    \end{tabular}%
  \label{tab:tables4}%
\end{table}%

\textit{4) Ablation Study on Evaluation Resolution.} Table \ref{tab:tables4} compares the pixel-level metrics computed at different evaluation resolutions. Throughout our experiments, both the predicted anomaly maps and the corresponding ground-truth masks are downsampled to 512 $\times$ 512 before pixel-level evaluation. This resolution is carefully selected with reference to the smallest anomalies in the benchmarks, whose bounding boxes are no smaller than 16 $\times$ 16 pixels and therefore remain represented after downsampling. As shown in Table \ref{tab:tables4}, the metrics computed at 512 $\times$ 512 remain highly consistent with those obtained at the original resolutions, with only minor differences arising primarily from boundary interpolation during rescaling. This evaluation protocol substantially reduces computational cost. For example, pixel-level evaluation of 4K images at their original resolution can require tens of GPU-hours and may even be more computationally expensive than detector training. In contrast, evaluation at 512 $\times$ 512 incurs negligible overhead, making extensive evaluation across multiple detectors and benchmarks computationally feasible without affecting the experimental conclusions. Nevertheless, 512 $\times$ 512 should not be regarded as a universally applicable evaluation resolution. Instead, the evaluation resolution must be carefully selected according to the anomaly-size distribution of the target dataset to ensure that small anomalous regions remain represented after downsampling.

\section{Visualization}
\label{sec: supD}

\begin{figure*}[h]
  \centering
   \includegraphics[width=\linewidth]{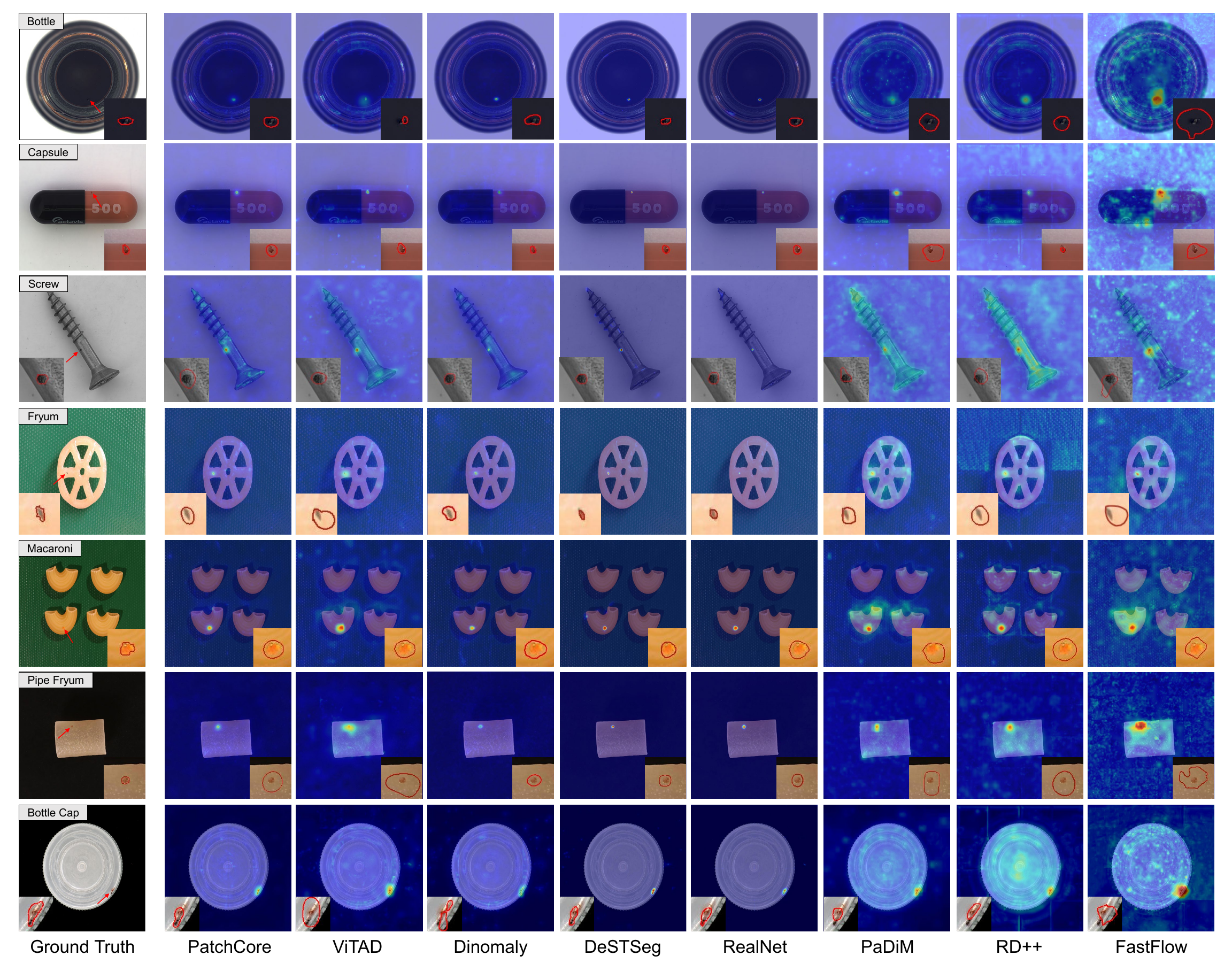}
   \caption{Qualitative results of HiAD using different anomaly detection methods for subtle anomaly detection at a resolution of $2048 \times 2048$.}
   \label{fig:figs2}
  \end{figure*}

\begin{figure*}[t]
  \centering
   \includegraphics[width=\linewidth]{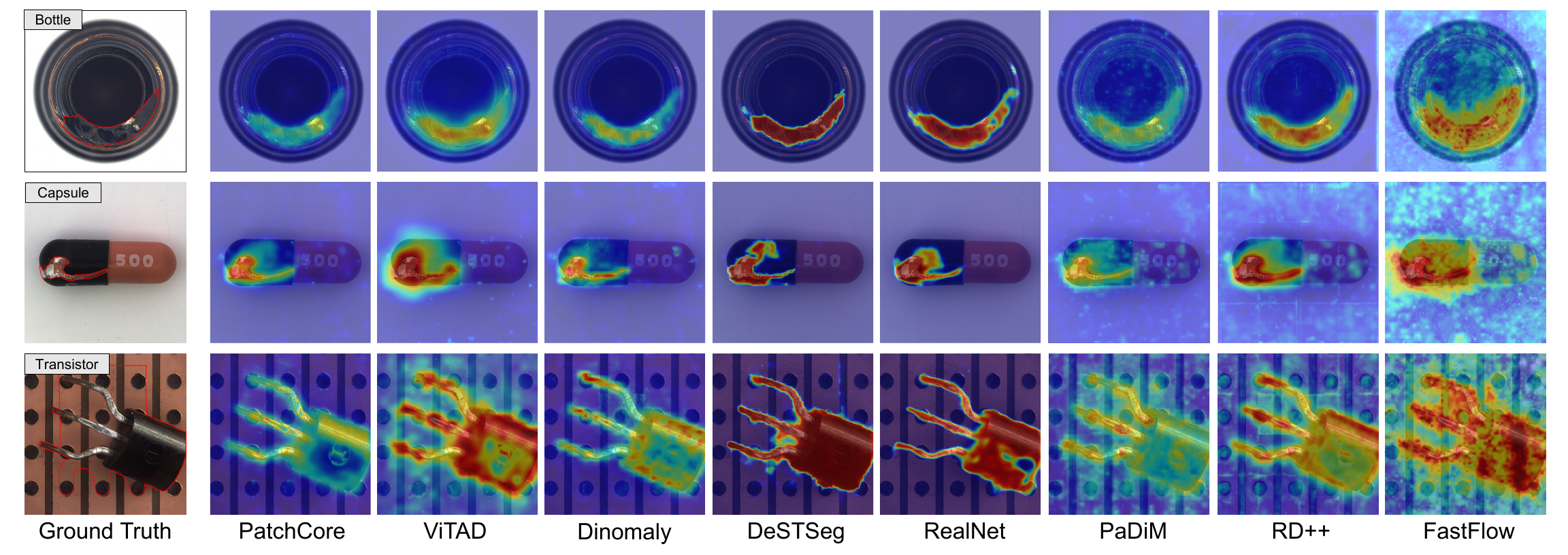}
   \caption{Qualitative results of HiAD using different anomaly detection methods for large-scale anomaly detection at a resolution of $2048 \times 2048$.}
   \label{fig:figs3}
\end{figure*}

\begin{figure*}[h]
  \centering
   \includegraphics[width=0.94\linewidth]{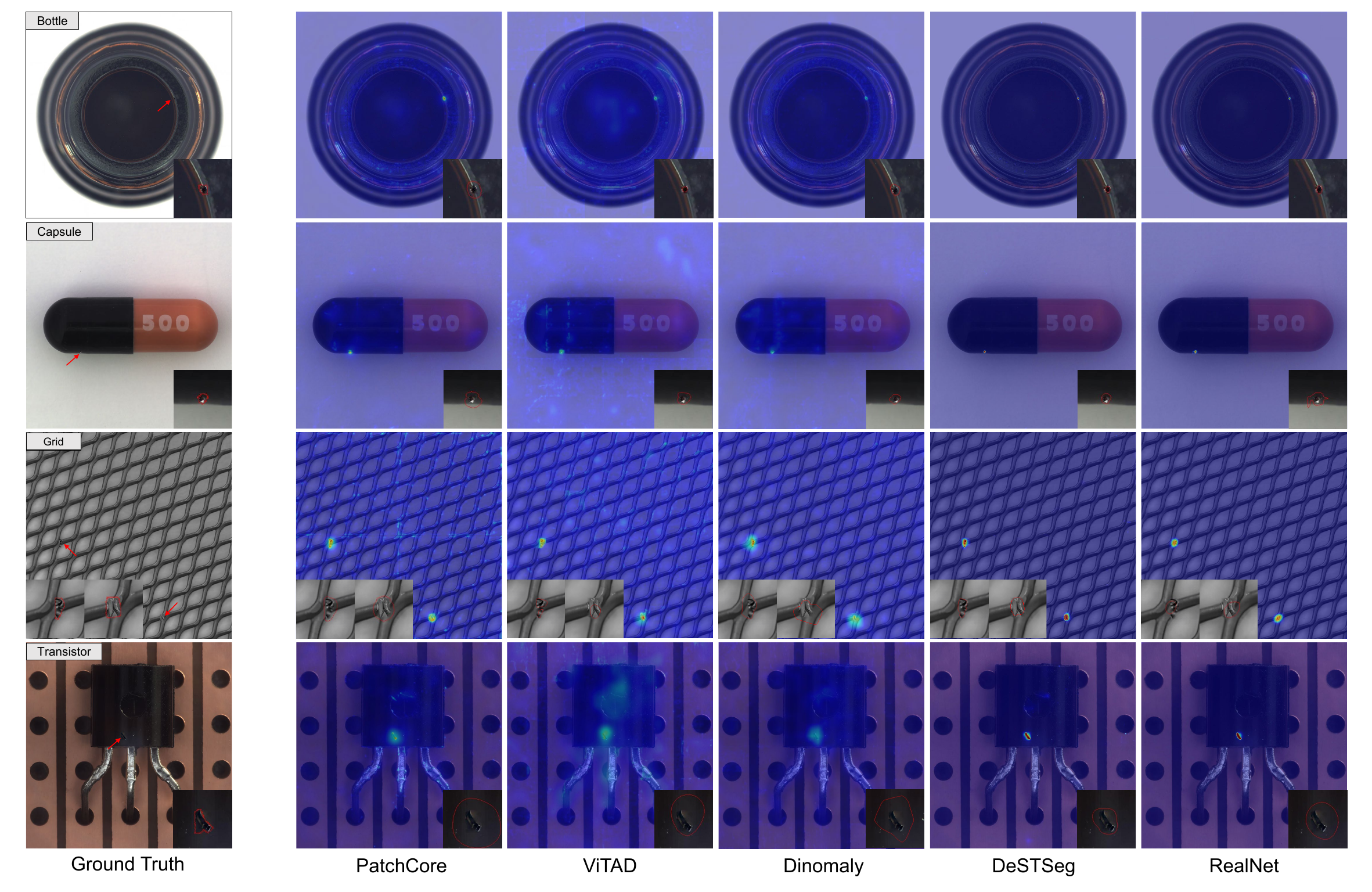}
   \vspace*{-0.1cm}
   \caption{Qualitative results of HiAD using different anomaly detection methods for subtle anomaly detection at a resolution of $4096 \times 4096$.}
   \label{fig:figs4}
   \vspace*{-1cm}
\end{figure*}

\begin{figure*}[t]
  \centering
   \includegraphics[width=0.94\linewidth]{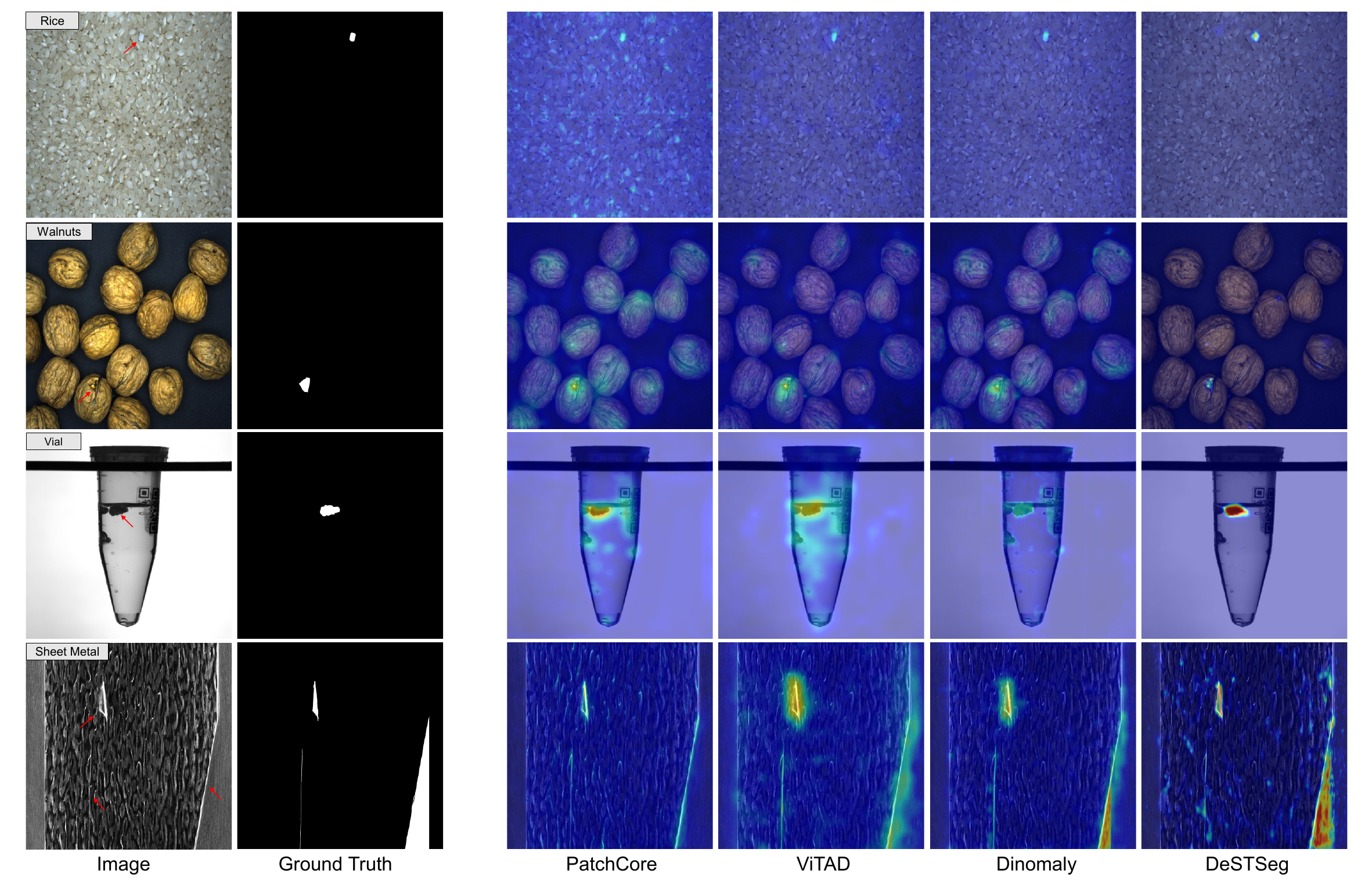}
   \caption{Qualitative results of HiAD using different anomaly detection methods on the MVTec 2 dataset.}
   \label{fig:figs5}
\end{figure*}

\begin{figure*}[t]
  \centering
   \includegraphics[width=\linewidth]{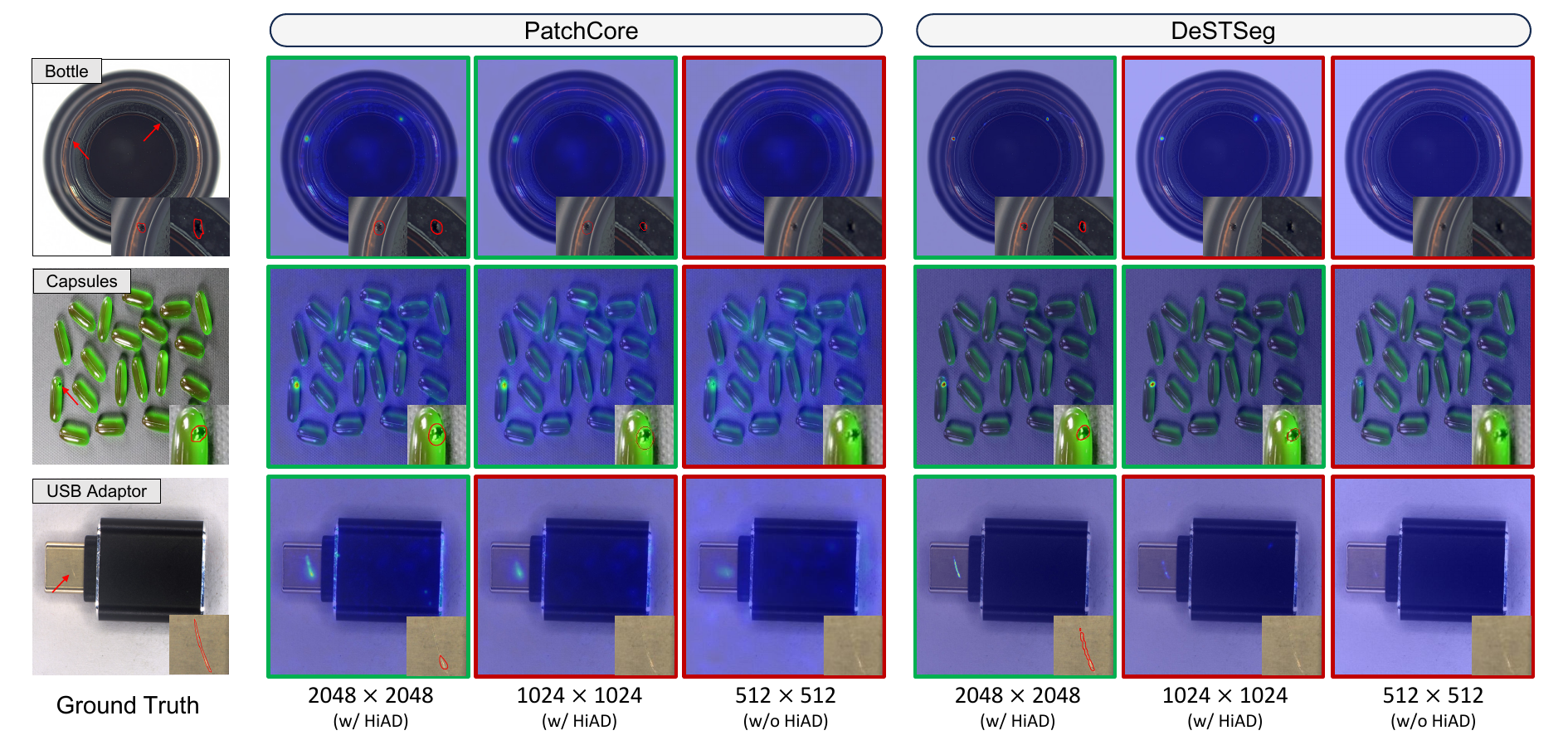}
   \caption{Qualitative comparison of anomaly detection results at different detection resolutions. Green borders indicate correctly identified anomalies, while red borders indicate missed detections.}
   \label{fig:figs6}
\end{figure*}

\begin{figure*}[t]
  \centering
   \includegraphics[width=\linewidth]{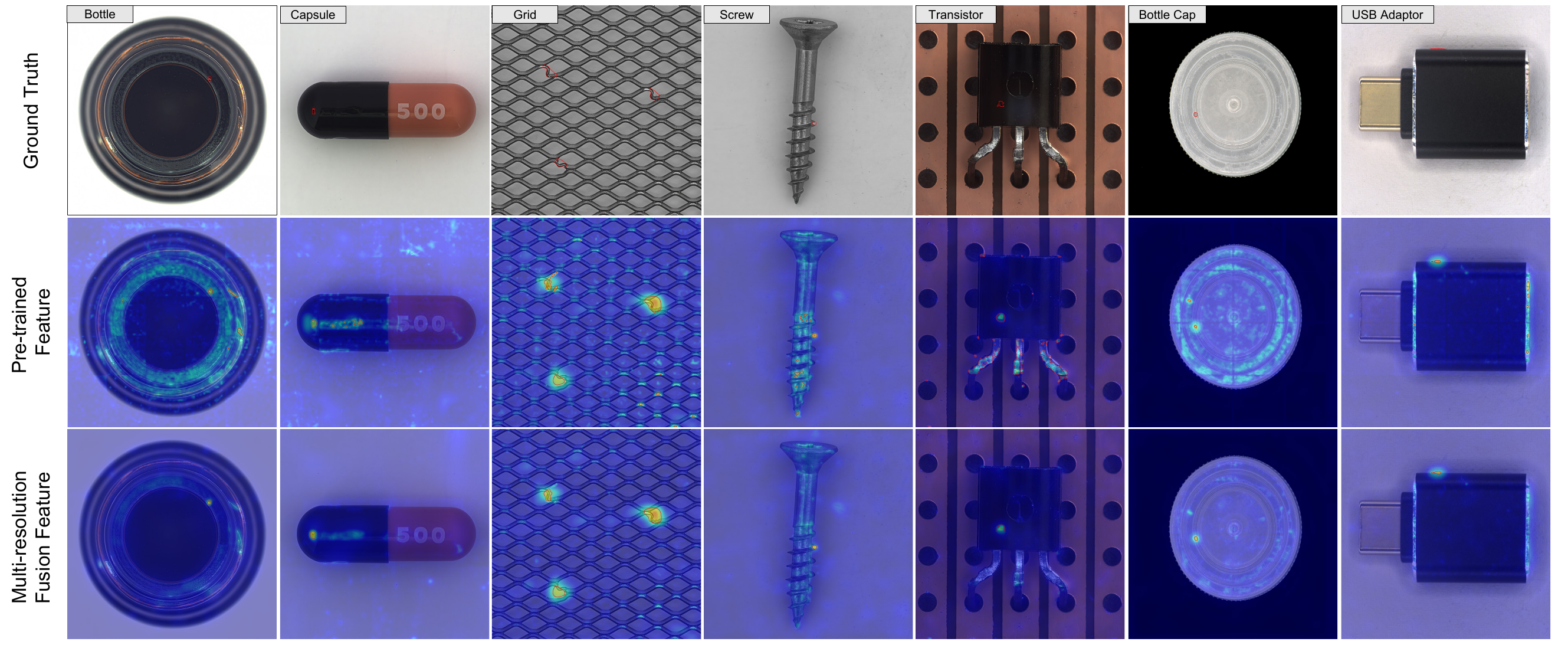}
   \caption{Qualitative comparison of pre-trained features and multi-resolution fusion features at a resolution of $2048 \times 2048$. The anomaly segmentation results are delineated by red boundaries.}
   \label{fig:figs7}
\end{figure*}

Figures \ref{fig:figs2} and \ref{fig:figs3} respectively illustrate the qualitative results of HiAD for detecting subtle and large-scale anomalies on the 2K datasets, while Figures \ref{fig:figs4} and \ref{fig:figs5} show the qualitative results for localizing subtle anomalies on the MVTec-4K and MVTec 2 datasets. These visualizations demonstrate that HiAD is widely applicable to various anomaly detection methods, enabling it to accurately localize anomalous regions of various scales in ultra-high-resolution images. Figure \ref{fig:figs6} shows the anomaly detection results at different detection resolutions, highlighting the necessity of high-resolution defect inspection. Figure \ref{fig:figs7} presents the qualitative results of original pre-trained features and multi-resolution fusion features, demonstrating the effectiveness of MRFF in mitigating ultrafine texture noise.

\section{Metrics for Each Category}
\label{sec: supE}
Tables \ref{tab:metric1}--\ref{tab:metric15} list category-specific anomaly detection metrics on the MVTec-2K, VisA-2K, and RealIAD-2K datasets, while Tables \ref{tab:metric16} and \ref{tab:metric17} provide the corresponding results on the MVTec-4K and MVTec 2 datasets, respectively. In these datasets, anomalies often appear as subtle defects, necessitating high-resolution inspection. Pairing HiAD with diverse anomaly detection methods yields promising performance in each category, underscoring HiAD's effectiveness and broad applicability.

\clearpage

\clearpage

\begin{table}[t]
  \centering
  \renewcommand\arraystretch{1.05}
  \caption{Anomaly detection results for the \textbf{Bottle} category in the MVTec-2K dataset.}
    \resizebox{\linewidth}{!}{
    {\small
\begin{tabular}{c|c|c|ccccc}
     \bottomrule
    Setting & Resolution & Method & I-AUC & P-AUC & P-AP & P-F1 & PRO \\
    \cline{1-8}  
        \multicolumn{1}{c|}{\multirow{14}[8]{*}{With HiAD}}  & \multicolumn{1}{c|}{\multirow{8}[1]{*}{$2048 \times 2048$}} & PatchCore  & 98.55 & 99.55 & 81.65 & 78.57 & \textbf{99.18} \\
          &       & Dinomaly  & \textbf{99.67} & 99.76 & 87.02 & 80.02 & 98.73 \\
          &       & ViTAD  & 98.68 & 99.64 & 79.70  & 76.03 & 97.52 \\
          &       & DeSTSeg  & 98.79 & 99.75 & 92.44 & 85.63 & 98.60 \\
          &       & RealNet & 95.88 & 98.89 & 80.88 & 75.04 & 95.43 \\
          &       & FastFlow & 94.20  & 99.35 & 69.40  & 67.32 & 94.67 \\
          &       & RD++ & 95.81 & 99.45 & 81.82 & 77.91 & 98.04 \\
\end{tabular}%
    }}
  \label{tab:metric1}%
  \vspace{-1.0cm}
\end{table}%

\begin{table}[t]
  \centering
  \renewcommand\arraystretch{1.05}
  \caption{Anomaly detection results for the \textbf{Capsule} category in the MVTec-2K dataset.}
    \resizebox{\linewidth}{!}{
    {\small
    \begin{tabular}{c|c|c|ccccc}
    \bottomrule
     Setting & Resolution & Method & I-AUC & P-AUC & P-AP & P-F1 & PRO \\
    \cline{1-8}   
    \multicolumn{1}{c|}{\multirow{14}[8]{*}{With HiAD}} & \multicolumn{1}{c|}{\multirow{8}[1]{*}{$2048 \times 2048$}} & PatchCore  & \textbf{99.65} & 99.24 & 55.28 & 53.53 & \textbf{99.12} \\
              &       & Dinomaly  & 96.32 & 99.31 & \textbf{57.51} & 55.69 &  98.88\\
          &       & ViTAD & 98.67 & 99.36 & 53.35 & 53.94 & 99.06 \\
          &       & DeSTSeg  & 97.56 & \textbf{99.47} & 50.77 & 57.81 & 95.41 \\
          &       & RealNet  & 95.88 & 97.52 & 54.26 & 57.55 & 88.24 \\
          &       & FastFlow & 92.29 & 99.31 & 50.10  & 51.12 & 96.15 \\
          &       & RD++ & 92.34 & 97.68 & 47.16 & 51.07 & 96.54 \\
\end{tabular}%
    }
    }
  \label{tab:metric2}%
\end{table}%

\begin{table}[t]
  \centering
  \renewcommand\arraystretch{1.05}
  \caption{Anomaly detection results for the \textbf{Grid} category in the MVTec-2K dataset.}
    \resizebox{\linewidth}{!}{
    {\small
      \begin{tabular}{c|c|c|ccccc}
     \bottomrule
    Setting & Resolution & Method & I-AUC & P-AUC & P-AP & P-F1 & PRO \\
    \cline{1-8}   
    \multicolumn{1}{c|}{\multirow{14}[8]{*}{With HiAD}} & \multicolumn{1}{c|}{\multirow{8}[1]{*}{$2048 \times 2048$}} & PatchCore  & 96.32 & 99.72 & 63.62 & 64.39 & \textbf{99.15} \\
          &       & Dinomaly  & \textbf{99.93} & 99.79 & 65.61 & 63.18 & 98.84 \\
          &       & ViTAD  & 97.16 & 99.51  & 51.12 & 50.11 & 98.82 \\
          &       & DeSTSeg  &  99.67 & 99.81 & 76.59 & 69.94 & 99.09 \\
          &       & RealNet  & 98.12 & 99.77 & 70.03 & 66.32 & 97.63 \\
          &       & FastFlow  & 91.96 & 99.29  & 47.15 & 47.70 & 95.36 \\
          &       & RD++   &  92.52	 & 99.74	 & 64.02 & 62.72 & 98.49 \\
\end{tabular}%
    }}
  \label{tab:metric3}%
  \vspace{-1.0cm}
\end{table}%

\begin{table}[t]
  \centering
  \renewcommand\arraystretch{1.05}
  \caption{Anomaly detection results for the \textbf{Hazelnut} category in the MVTec-2K dataset.}
    \resizebox{\linewidth}{!}{
    {\small
        \begin{tabular}{c|c|c|ccccc}
   \bottomrule
    Setting & Resolution & Method & I-AUC & P-AUC & P-AP & P-F1 & PRO \\
        \cline{1-8}   
    \multicolumn{1}{c|}{\multirow{14}[8]{*}{With HiAD}} & \multicolumn{1}{c|}{\multirow{8}[1]{*}{$2048 \times 2048$}} & PatchCore  & \textbf{98.87} & 99.57 & 71.49 & 69.86 & \textbf{98.85} \\
              &       & Dinomaly  & 94.20 & \textbf{99.80} & \textbf{82.62} & \textbf{78.75} &  97.71\\
          &       & ViTAD  & 96.43 & 99.71 & 75.08 & 73.68 & 98.13 \\
          &       & DeSTSeg  & 96.86 & 98.84 & 77.18 & 72.99 & 97.91 \\
          &       & RealNet  & 96.05 & 99.26 & 70.15 & 66.01 & 90.04 \\
          &       & FastFlow  & 90.01 & 98.54 & 54.11 & 54.82 & 93.33 \\
          &       & RD++   & 94.10  & 99.50  & 70.75 & 68.09 & 98.15 \\
\end{tabular}%
    }
    }
  \label{tab:metric4}%
\end{table}%

\clearpage

\begin{table}[t]
  \centering
  \renewcommand\arraystretch{1.05}
  \caption{Anomaly detection results for the \textbf{Screw} category in the MVTec-2K dataset.}
    \resizebox{\linewidth}{!}{
   {\small      
   \begin{tabular}{c|c|c|ccccc}
        \bottomrule
        Setting & Resolution & Method & I-AUC & P-AUC & P-AP & P-F1 & PRO \\
        \cline{1-8}   
        \multicolumn{1}{c|}{\multirow{14}[8]{*}{With HiAD}} & \multicolumn{1}{c|}{\multirow{8}[1]{*}{$2048 \times 2048$}} & PatchCore  & \textbf{98.74} & 99.84 & 72.34 & 68.78 & 99.75 \\
                  &       & Dinomaly  & 98.13 & \textbf{99.94} & \textbf{73.80} & \textbf{70.95} & \textbf{99.79} \\
              &       & ViTAD  & 91.04 & 99.85 & 55.24 & 60.74 & 99.46 \\
              &       & DeSTSeg  & 98.48  & 99.61 & 61.29 & 61.52 & 98.57 \\
              &       & RealNet  & 90.83 & 99.41 & 44.68 & 50.06 & 94.88 \\
              &       & FastFlow  & 93.41 & 99.69 & 45.37 & 50.74 & 99.04 \\
              &       & RD++  & 87.07 & 99.39 & 45.24 & 50.18 & 98.70 \\
\end{tabular}%
        }
    }
  \label{tab:metric5}%
  \vspace*{-1cm}
\end{table}%

\begin{table}[t]
  \centering
  \renewcommand\arraystretch{1.05}
  \caption{Anomaly detection results for the \textbf{Transistor} category in the MVTec-2K dataset.}
    \resizebox{\linewidth}{!}{
    {\small
       \begin{tabular}{c|c|c|ccccc}
        \bottomrule
         Setting & Resolution & Method & I-AUC & P-AUC & P-AP & P-F1 & PRO \\
        \cline{1-8}   
        \multicolumn{1}{c|}{\multirow{14}[8]{*}{With HiAD}} & \multicolumn{1}{c|}{\multirow{8}[1]{*}{$2048 \times 2048$}} & PatchCore  & 97.27 & 87.64 & 43.19 & 44.75 & 88.43 \\
              &       & Dinomaly  & \textbf{97.69} & 95.45 & 61.11 & 59.00 & 92.05 \\
              &       & ViTAD  & 96.22 & \textbf{96.01} & \textbf{61.71} & 60.95 & 91.58 \\
              &       & DeSTSeg  & 95.81 & 84.98 & 52.60  & 53.37 & 90.72 \\
              &       & RealNet  & 97.04 & 84.05 & 41.06 & 44.74 & 79.09 \\
              &       & FastFlow  & 89.24 & 95.63 & 52.34 & 53.27 & 92.69 \\
              &       & RD++   & 87.10  & 83.10  & 40.94 & 44.25 & 86.74 \\
\end{tabular}%
    }
    }
  \label{tab:metric6}%
\end{table}%

\begin{table}[t]
  \centering
  \renewcommand\arraystretch{1.05}
  \caption{Anomaly detection results for the \textbf{Wood} category in the MVTec-2K dataset.}
    \resizebox{\linewidth}{!}{
   {\small
       \begin{tabular}{c|c|c|ccccc}
    \bottomrule
    Setting & Resolution & Method & I-AUC & P-AUC & P-AP & P-F1 & PRO \\
    \cline{1-8}   
    \multicolumn{1}{c|}{\multirow{14}[8]{*}{With HiAD}} & \multicolumn{1}{c|}{\multirow{8}[1]{*}{$2048 \times 2048$}} & PatchCore  & \textbf{99.68} & 98.98 & 73.30  & 68.62 & \textbf{97.23} \\
              &       & Dinomaly  & 94.27 & \textbf{99.47} & 83.17 & 76.13 & 97.07 \\
          &       & ViTAD  & 96.98 & 99.33 & 77.27 & 73.17 & 96.53 \\
          &       & DeSTSeg  & 98.36 & 	98.88 & 84.05 & 77.03 & 95.28 \\
          &       & RealNet  & 99.43 & 99.05 & 83.32  & 75.76 & 95.79 \\
          &       & FastFlow  & 90.87 & 97.80 & 61.06 & 63.13 & 90.09 \\
          &       & RD++  & 94.00 & 98.32 & 65.93  & 64.88 & 94.35 \\
\end{tabular}%
    }
    }
  \label{tab:metric7}%
  \vspace*{-1cm}
\end{table}%

\begin{table}[t]
  \centering
  \renewcommand\arraystretch{1.05}
  \caption{Anomaly detection results for the \textbf{Capsules} category in the VisA-2K dataset.}
    \resizebox{\linewidth}{!}{
   {\small
           \begin{tabular}{c|c|c|ccccc}
   \bottomrule
    Setting & Resolution & Method & I-AUC & P-AUC & P-AP & P-F1 & PRO \\
    \cline{1-8}   
    \multicolumn{1}{c|}{\multirow{14}[8]{*}{With HiAD}} & \multicolumn{1}{c|}{\multirow{8}[1]{*}{$2048 \times 2048$}} & PatchCore  & 97.84 & 99.63 & 72.45 & 71.23 & \textbf{99.26} \\
              &       & Dinomaly  & \textbf{98.98} & \textbf{99.78} & 65.75 & 65.27 & \textbf{99.26} \\
          &       & ViTAD  & 92.78 & 99.31 & 45.89 & 51.32 & 96.75 \\
          &       & DeSTSeg  & 92.44 & 99.52& 73.53 & 73.69 & 97.49 \\
          &       & RealNet  & 90.13 & 99.13 & 60.23 & 64.68  & 95.18 \\
          &       & FastFlow  & 91.39 & 99.39 & 44.73 & 54.49 & 95.64 \\
          &       & RD++  & 89.99 & 98.69  & 61.86 & 59.02 & 95.88 \\
\end{tabular}%
    }
    }
  \label{tab:metric8}%
\end{table}%

\clearpage

\begin{table}[t]
  \centering
  \renewcommand\arraystretch{1.05}
  \caption{Anomaly detection results for the \textbf{Fryum} category in the VisA-2K dataset.}
    \resizebox{\linewidth}{!}{
    {\small
        \begin{tabular}{c|c|c|ccccc}
    \bottomrule
    Setting & Resolution & Method & I-AUC & P-AUC & P-AP & P-F1 & PRO \\
    \cline{1-8}   
    \multicolumn{1}{c|}{\multirow{14}[8]{*}{With HiAD}} & \multicolumn{1}{c|}{\multirow{8}[1]{*}{$2048 \times 2048$}} & PatchCore  & 99.19 & 99.73 & 52.97 & \textbf{54.53} & 98.50 \\
              &       & Dinomaly  & \textbf{99.96}	& 99.69 & 43.65 & 45.49 & \textbf{98.58} \\
          &       & ViTAD  & 96.89 & 99.73 & 41.13 & 45.47 & 98.32 \\
          &       & DeSTSeg  & 95.70  & 98.26 & \textbf{54.39} & 53.41 & 91.63 \\
          &       & RealNet  & 91.89 & 97.50  & 29.90  & 35.98 & 80.12 \\
          &       & FastFlow  & 97.50  & 99.44 & 44.28 & 47.82 & 94.15 \\
          &       & RD++  & 95.68 & 99.54 & 40.13 & 48.99 & 95.24 \\
\end{tabular}%
    }
    }
  \label{tab:metric9}%
  \vspace*{-1cm}
\end{table}%

\begin{table}[t]
  \centering
  \renewcommand\arraystretch{1.05}
  \caption{Anomaly detection results for the \textbf{Macaroni} category in the VisA-2K dataset.}
    \resizebox{\linewidth}{!}{
     {\small
        \begin{tabular}{c|c|c|ccccc}
        \bottomrule
        Setting & Resolution & Method & I-AUC & P-AUC & P-AP & P-F1 & PRO \\
        \cline{1-8}   
        \multicolumn{1}{c|}{\multirow{14}[8]{*}{With HiAD}} & \multicolumn{1}{c|}{\multirow{8}[1]{*}{$2048 \times 2048$}} & PatchCore  & \textbf{98.29} & \textbf{99.95} & \textbf{51.77} & \textbf{52.57} & 99.17 \\
                  &       & Dinomaly  & 97.98  & 99.89 & 30.21 & 34.59 & \textbf{99.21} \\
              &       & ViTAD  & 93.70  & 99.51 & 35.32 & 40.62 & 94.22 \\
              &       & DeSTSeg  & 91.23 & 99.40  & 40.50  & 45.20  & 90.68 \\
              &       & RealNet  & 91.74 & 99.72 & 39.84 & 43.82 & 88.80 \\
              &       & FastFlow  & 96.38 & 99.77 & 26.22 & 30.71 & 98.07 \\
              &       & RD++  & 94.06 & 99.66 & 31.95 & 39.52 & 96.90 \\
\end{tabular}%
        }
    }
  \label{tab:metric10}%
\end{table}%

\begin{table}[t]
  \centering
  \renewcommand\arraystretch{1.05}
  \caption{Anomaly detection results for the \textbf{PCB} category in the VisA-2K dataset.}
    \resizebox{\linewidth}{!}{
    {\small
        \begin{tabular}{c|c|c|ccccc}
    \bottomrule
     Setting & Resolution & Method & I-AUC & P-AUC & P-AP & P-F1 & PRO \\
    \cline{1-8}   
    \multicolumn{1}{c|}{\multirow{14}[8]{*}{With HiAD}} & \multicolumn{1}{c|}{\multirow{8}[1]{*}{$2048 \times 2048$}} & PatchCore  & \textbf{99.15} & 99.60  & \textbf{87.08} & 81.28 & 97.80 \\
              &       & Dinomaly  & 98.98 & 99.68 & 86.00 & \textbf{83.44} & \textbf{98.25} \\
          &       & ViTAD  & 93.89 & 99.56 & 80.20  & 74.77 & 96.98 \\
          &       & DeSTSeg  & 98.69 & 99.66 & 84.22 & 75.61 & 95.42 \\
          &       & RealNet  & 93.32 & 99.19 & 79.16 & 74.07 & 91.79 \\
          &       & FastFlow  & 92.11 & 99.50  & 70.01 & 67.47 & 93.67 \\
          &       & RD++  & 91.92 & 99.72 & 77.42 & 76.19 & 95.52 \\
\end{tabular}%
    }
    }
  \label{tab:metric11}%
  \vspace*{-1cm}
\end{table}%

\begin{table}[t]
  \centering
  \renewcommand\arraystretch{1.05}
  \caption{Anomaly detection results for the \textbf{Pipe Fryum} category in the VisA-2K dataset.}
    \resizebox{\linewidth}{!}{
    {\small
         \begin{tabular}{c|c|c|ccccc}
        \bottomrule
        Setting & Resolution & Method & I-AUC & P-AUC & P-AP & P-F1 & PRO \\
        \cline{1-8}   
        \multicolumn{1}{c|}{\multirow{14}[8]{*}{With HiAD}} & \multicolumn{1}{c|}{\multirow{8}[1]{*}{$2048 \times 2048$}} & PatchCore  & \textbf{99.61} & 99.39 & 56.07 & 60.29 & 98.96 \\
              &       & Dinomaly  & 96.94 & 99.68 & 58.00 & 60.65 & \textbf{98.98} \\
              &       & ViTAD  & 98.68 & 99.68 & 59.83 & 61.37 & 98.69 \\
              &       & DeSTSeg  & 93.06 & 96.68 & 66.33 & 64.70  & 82.73 \\
              &       & RealNet  & 95.45 & 98.81 & 62.98 & 63.63 & 88.44 \\
              &       & FastFlow  & 96.88 & \textbf{99.79} & \textbf{70.32} & \textbf{71.50} & 97.53 \\
              &       & RD++   & 98.23 & 99.10  & 47.06 & 55.21 & 98.18 \\
\end{tabular}%
        }
    }
  \label{tab:metric12}%
\end{table}%

\clearpage

\begin{table}[t]
  \centering
  \renewcommand\arraystretch{1.05}
  \caption{Anomaly detection results for the \textbf{Bottle Cap} category in the RealIAD-2K dataset.}
    \resizebox{\linewidth}{!}{
       {\small
            \begin{tabular}{c|c|c|ccccc}
    \bottomrule
        Setting & Resolution & Method & I-AUC & P-AUC & P-AP & P-F1 & PRO \\
        \cline{1-8}   
        \multicolumn{1}{c|}{\multirow{14}[8]{*}{With HiAD}} & \multicolumn{1}{c|}{\multirow{8}[1]{*}{$2048 \times 2048$}} & PatchCore  & 96.89 & 99.95 & 53.74 & 55.83 & 99.44 \\
                  &       & Dinomaly  & 96.75 & 99.95 & 43.92 & 42.44  & 99.42 \\
              &       & ViTAD  & 96.53 & 99.96 & 38.87 & 44.52 & 99.69 \\
              &       & DeSTSeg  & 96.39 & 99.40  & \textbf{71.31} & \textbf{66.29} & 98.54 \\
              &       & RealNet  & 95.63 & \textbf{99.98} & 59.05 & 59.13 & 99.05 \\
              &       & FastFlow  & \textbf{97.59} & \textbf{99.98} & 32.64 & 38.56 & \textbf{99.88} \\
              &       & RD++  & 88.84 & 98.84 & 38.33 & 45.43 & 92.04 \\
\end{tabular}%
        }
    }
  \label{tab:metric13}%
  \vspace*{-1cm}
\end{table}%

\begin{table}[t]
  \centering
  \renewcommand\arraystretch{1.05}
  \caption{Anomaly detection results for the \textbf{Mint} category in the RealIAD-2K dataset.}
    \resizebox{\linewidth}{!}{
   {\small     
   \begin{tabular}{c|c|c|ccccc}
        \bottomrule
         Setting & Resolution & Method & I-AUC & P-AUC & P-AP & P-F1 & PRO \\
        \cline{1-8}   
        \multicolumn{1}{c|}{\multirow{14}[8]{*}{With HiAD}} & \multicolumn{1}{c|}{\multirow{8}[1]{*}{$2048 \times 2048$}} & PatchCore  & \textbf{98.48} & \textbf{99.87} & 38.50  & 40.69 & \textbf{99.55} \\
                  &       & Dinomaly  & 89.27 & 99.71 & 33.95 & 39.54 & 98.79 \\
              &       & ViTAD  & 91.33 & 99.76 & 33.67 & 41.50 & 98.93 \\
              &       & DeSTSeg  & 89.71 & 99.57 & \textbf{46.98} & \textbf{47.15} & 96.94 \\
              &       & RealNet  & 86.98 & 99.09 & 37.25 & 41.06 & 89.52 \\
              &       & FastFlow  & 86.23 & 98.21 & 22.42 & 30.19 & 92.60 \\
              &       & RD++  & 84.90  & 98.49 & 24.10  & 32.48 & 95.13 \\
\end{tabular}%
        }
    }
  \label{tab:metric14}%
\end{table}%

\begin{table}[t]
  \centering
  \renewcommand\arraystretch{1.05}
  \caption{Anomaly detection results for the \textbf{USB Adaptor} category in the RealIAD-2K dataset.}
    \resizebox{\linewidth}{!}{
    {\small
        \begin{tabular}{c|c|c|ccccc}
    \bottomrule
    Setting & Resolution & Method & I-AUC & P-AUC & P-AP & P-F1 & PRO \\
    \cline{1-8}   
    \multicolumn{1}{c|}{\multirow{14}[8]{*}{With HiAD}} & \multicolumn{1}{c|}{\multirow{8}[1]{*}{$2048 \times 2048$}} & PatchCore  & \textbf{99.73} & 99.84 & 43.23 & 45.50  & 98.31 \\
              &       & Dinomaly  & 96.75 & \textbf{99.92} & 40.99 & 40.66 & \textbf{99.49} \\
          &       & ViTAD  & 95.55 & 99.64 & 37.94 & 39.63 & 95.58 \\
          &       & DeSTSeg  & 94.41 & 99.42 & \textbf{45.72} & \textbf{48.44} & 91.18 \\
          &       & RealNet  & 96.45 & 99.81 & 44.22 & 46.39 & 89.57 \\
          &       & FastFlow & 97.16 & 99.77 & 22.84 & 29.71 & 97.89 \\
          &       & RD++  & 90.36 & 99.28 & 24.65 & 30.91 & 94.88 \\
\end{tabular}%
    }
    }
    \vspace{11.9cm}
  \label{tab:metric15}%
\end{table}%

\clearpage

\begin{table*}[t]
  \centering
  \renewcommand\arraystretch{1.2}
  \caption{Anomaly detection results of HiAD on the MVTec-4K dataset at a resolution of $4096 \times 4096$.}
  \resizebox{0.9\linewidth}{!}{
    \begin{tabular}{c|c|ccccc}
    
    \bottomrule
    Category & Method & I-AUC & P-AUC & P-AP & P-F1 & PRO \\
    \hline
    \multirow{4}{*}{Bottle} 
          & Dinomaly & \textbf{99.88} & 99.71 & 85.69 & 79.73 & 98.49 \\
          & DeSTSeg  & \textbf{99.88} & 99.67 & \textbf{92.58} & \textbf{85.99} & 95.99 \\
          & PatchCore & 99.69 & 99.52 & 81.03 & 78.35 & \textbf{98.66} \\
          & ViTAD    & 97.68 & \textbf{99.74} & 84.00 & 80.47 & 97.56 \\
    \hline
    \multirow{4}{*}{Capsule} 
          & Dinomaly & 95.34 & \textbf{99.37} & \textbf{60.23} & 57.12 & \textbf{99.05} \\
          & DeSTSeg  & \textbf{98.56} & 99.35 & 53.30 & \textbf{59.89} & 95.07 \\
          & PatchCore & 97.22 & 99.34 & 56.50 & 54.87 & 98.88 \\
          & ViTAD    & 94.90 & 99.15 & 52.25 & 54.96 & 98.53 \\
    \hline
    \multirow{4}{*}{Grid} 
          & Dinomaly & \textbf{98.90} & 99.77& 58.17 & 59.00 & 98.81 \\
          & DeSTSeg  & 98.12 & \textbf{99.85} & \textbf{77.68} & \textbf{72.15} & \textbf{98.93} \\
          & PatchCore & 86.08 & 99.47 & 50.90 & 54.02 & 98.47 \\
          & ViTAD    & 98.43 & 99.57 & 43.20 & 49.11 & 98.87 \\
    \hline
    \multirow{4}{*}{Hazelnut} 
          & Dinomaly & 92.15 & \textbf{99.79} & 78.69 & 77.29 & 98.78 \\
          & DeSTSeg  & 99.45 & 99.78 & \textbf{88.71} & \textbf{82.32} & 99.06 \\
          & PatchCore & \textbf{99.71} & 99.56 & 69.25 & 68.30 & \textbf{99.22} \\
          & ViTAD    & 97.75 & 99.70 & 74.08 & 72.80 & 98.81 \\
    \toprule
    \end{tabular}

    \hskip 10pt 

    \begin{tabular}{c|c|ccccc}
    \bottomrule
    Category & Method & I-AUC & P-AUC & P-AP & P-F1 & PRO \\
    \hline
    \multirow{4}{*}{Screw} 
          & Dinomaly & 93.17 & \textbf{99.91} & 64.52 & 65.24 & \textbf{99.75} \\
          & DeSTSeg  & \textbf{97.51} & 99.59 & \textbf{74.58} & \textbf{68.71} & 97.07\\
          & PatchCore & 94.06 & 99.85 & 70.86 & 67.05 & 99.68 \\
          & ViTAD    & 91.80 & 99.81 & 54.21 & 55.33 & 99.45 \\
    \hline
    \multirow{4}{*}{Transistor} 
          & Dinomaly & 97.44 & 95.00 & 58.15 & 57.28 & 90.43 \\
          & DeSTSeg  & \textbf{98.97} & 89.78 & 57.00 & 54.09 & \textbf{91.47} \\
          & PatchCore & 93.99 & 88.45 & 44.37 & 45.85 & 88.06 \\
          & ViTAD    & 89.27 & \textbf{95.33} & \textbf{60.52} & \textbf{59.83} & 90.09 \\
    \hline
    \multirow{4}{*}{Wood} 
          & Dinomaly & 97.40 & \textbf{99.40} & 83.43 & 75.29 & \textbf{98.66} \\
          & DeSTSeg  & \textbf{99.19} & 98.96 & \textbf{84.88} & \textbf{78.02} & 96.81 \\
          & PatchCore & 96.49 & 98.79 & 69.68 & 66.17 & 98.13 \\
          & ViTAD    & 98.27 & 99.25 & 75.64 & 71.61 & 96.49 \\		
    \hline
    \multirow{4}{*}{\textbf{AVG}} 
          & Dinomaly & 96.33 & \textbf{98.99} & 69.84 & 67.28 & \textbf{97.71} \\
          & DeSTSeg  & \textbf{98.81} & 98.14 & \textbf{75.53} & \textbf{71.60} & 96.34 \\
          & PatchCore & 95.32 & 97.85 & 63.23 & 62.09 & 97.30 \\
          & ViTAD    & 95.44 & 98.94 & 63.41 & 63.44 & 97.11 \\
    \toprule
    \end{tabular}
  }
    \label{tab:metric16}
\end{table*}

\begin{table*}[t]
  \centering
  \renewcommand\arraystretch{1.2}
  \caption{Anomaly detection results on the MVTec 2 dataset. $\mathcal{y}$ denotes the use of HiAD.}
  \resizebox{0.9\linewidth}{!}{
  \large
    \begin{tabular}{c|c|c|ccccc|cccccccc}
    \bottomrule
    \multicolumn{1}{c|}{Category} & Resolution & Method & \multicolumn{1}{c}{I-AUC} & \multicolumn{1}{c}{P-AUC} & \multicolumn{1}{c}{P-AP} & \multicolumn{1}{c}{P-F1} & \multicolumn{1}{c|}{PRO} & \multicolumn{1}{c|}{Category} & \multicolumn{1}{c|}{Resolution} & \multicolumn{1}{c|}{Method} & \multicolumn{1}{c}{I-AUC} & \multicolumn{1}{c}{P-AUC} & \multicolumn{1}{c}{P-AP} & \multicolumn{1}{c}{P-F1} & \multicolumn{1}{c}{PRO} \\
    \hline
    \multicolumn{1}{c|}{\multirow{8}[1]{*}{Can}} & {\multirow{3}[1]{*}{Original$\mathcal{y}$}} & Dinomaly & 62.24 & \textbf{91.86} & 4.31  & 8.72  & \textbf{71.86} & \multicolumn{1}{c|}{\multirow{8}[1]{*}{Vial}} & \multicolumn{1}{c|}{{\multirow{3}[1]{*}{Original$\mathcal{y}$}}} & \multicolumn{1}{c|}{Dinomaly} & \multicolumn{1}{c}{85.17} & \multicolumn{1}{c}{93.26} & \multicolumn{1}{c}{33.38} & \multicolumn{1}{c}{35.42} & \multicolumn{1}{c}{92.75} \\
          & \multirow{3}[1]{*}{(2048 $\times$ 1024)} & DeSTSeg & \textbf{64.92} & 82.98 & \textbf{5.75} & 6.34  & 64.46 & \multicolumn{1}{c|}{} & \multicolumn{1}{c|}{\multirow{3}[1]{*}{(2048 $\times$ 1536)}} & \multicolumn{1}{c|}{DeSTSeg} & \multicolumn{1}{c}{88.14} & \multicolumn{1}{c}{88.35} & \multicolumn{1}{c}{35.15} & \multicolumn{1}{c}{36.44} & \multicolumn{1}{c}{84.27} \\
          & \multicolumn{1}{r|}{} & PatchCore & 57.30  & 87.95 & 2.19  & 4.55  & 68.24 & \multicolumn{1}{c|}{} & \multicolumn{1}{r|}{} & \multicolumn{1}{c|}{PatchCore} & \multicolumn{1}{c}{85.61} & \multicolumn{1}{c}{93.22} & \multicolumn{1}{c}{31.13} & \multicolumn{1}{c}{\textbf{42.31}} & \multicolumn{1}{c}{93.23} \\
          & \multicolumn{1}{r|}{} & ViTAD & 59.71 & 74.94 & 3.24  & \textbf{8.79} & 65.54 & \multicolumn{1}{c|}{} & \multicolumn{1}{r|}{} & \multicolumn{1}{c|}{ViTAD} & \multicolumn{1}{c}{75.17} & \multicolumn{1}{c}{93.35} & \multicolumn{1}{c}{28.28} & \multicolumn{1}{c}{30.71} & \multicolumn{1}{c}{90.49} \\
\end{tabular}%
}
\label{tab:metric17}
\end{table*}%

\end{document}